\documentclass{article}

\usepackage{iclr2027_conference,times}

\usepackage{amsmath,amsfonts,bm}

\def\eqref#1{equation~\ref{#1}}

\def\1{\bm{1}}

\DeclareMathAlphabet{\mathsfit}{\encodingdefault}{\sfdefault}{m}{sl}
\SetMathAlphabet{\mathsfit}{bold}{\encodingdefault}{\sfdefault}{bx}{n}

\usepackage{amsmath}
\usepackage{amssymb}
\usepackage{booktabs}
\usepackage{array}
\usepackage{longtable}
\usepackage{graphicx}
\usepackage{wrapfig}
\usepackage{microtype}
\usepackage{xcolor}
\usepackage{colortbl}
\usepackage{tcolorbox}
\usepackage{hyperref}
\usepackage{url}

\definecolor{citationpurple}{HTML}{71558F}
\definecolor{referenceblue}{HTML}{1F4E79}
\hypersetup{
    colorlinks=true,
    citecolor=citationpurple,
    linkcolor=referenceblue,
    urlcolor=black,
    pdfborder={0 0 0}
}

\DeclareRobustCommand{\crossref}[1]{\textcolor{referenceblue}{#1}}

\definecolor{takeawayback}{HTML}{F4EFF9}
\definecolor{takeawayframe}{HTML}{AC98C2}
\definecolor{takeawaylabel}{HTML}{71558F}
\newsavebox{\takeawaycontent}
\newenvironment{takeaway}[1]{%
    \par\addvspace{10pt}%
    \def\takeawaynumber{#1}%
    \begin{lrbox}{\takeawaycontent}%
    \begin{minipage}{\dimexpr\linewidth-16.6pt\relax}%
    \normalsize\setlength{\parindent}{0pt}\setlength{\parskip}{0pt}%
    \strut\ignorespaces
}{%
    \unskip\strut\par\end{minipage}\end{lrbox}%
    \noindent\begin{tikzpicture}
        \node[draw=takeawayframe,fill=takeawayback,line width=0.6pt,
            rounded corners=2.5pt,inner xsep=8pt,inner ysep=7pt,
            outer sep=0pt] (takeawaybody) {\usebox{\takeawaycontent}};
        \node[anchor=west,fill=takeawaylabel,text=white,
            rounded corners=2pt,
            font=\fontfamily{phv}\fontseries{b}\fontshape{n}\footnotesize,
            inner xsep=8pt,inner ysep=2.2pt,outer sep=0pt]
            at ([xshift=8pt]takeawaybody.north west)
            {T{\fontsize{7.2}{9}\selectfont\textls[40]{AKEAWAY}}\,\takeawaynumber};
    \end{tikzpicture}\par\addvspace{8pt}%
}

\newcolumntype{L}[1]{>{\raggedright\arraybackslash}p{#1}}
\newcommand{\modelname}[1]{{\def\UrlFont{\normalfont}\def\UrlBigBreaks{\do\-}\nolinkurl{#1}}}
\title{Beyond Prompt Count: How Data Shapes\\Transfer in On-Policy Distillation}

\usetikzlibrary{svg.path}
\newcommand{\githubmark}{%
    \raisebox{-1.5pt}{\tikz[xscale=0.65,yscale=-0.65]{%
        \path[fill=black] svg {M6.766 11.328c-2.063-.25-3.516-1.734-3.516-3.656 0-.781.281-1.625.75-2.188-.203-.515-.172-1.609.063-2.062.625-.078 1.468.25 1.968.703.594-.187 1.219-.281 1.985-.281.765 0 1.39.094 1.953.265.484-.437 1.344-.765 1.969-.687.218.422.25 1.515.046 2.047.5.593.766 1.39.766 2.203 0 1.922-1.453 3.375-3.547 3.64.531.344.89 1.094.89 1.954v1.625c0 .468.391.734.86.547C13.781 14.359 16 11.53 16 8.03 16 3.61 12.406 0 7.984 0 3.563 0 0 3.61 0 8.031a7.88 7.88 0 0 0 5.172 7.422c.422.156.828-.125.828-.547v-1.25c-.219.094-.5.156-.75.156-1.031 0-1.64-.562-2.078-1.609-.172-.422-.36-.672-.719-.719-.187-.015-.25-.093-.25-.187 0-.188.313-.328.625-.328.453 0 .844.281 1.25.86.313.452.64.655 1.031.655s.641-.14 1-.5c.266-.265.47-.5.657-.656};%
    }}%
}
\definecolor{abstractlabel}{HTML}{0052D9}
\colorlet{abstractback}{abstractlabel!5!white}
\colorlet{abstractframe}{abstractlabel!35!white}
\author{Jiaxuan Wang, Jiafei Lyu, Yuchen Cai, Siye Wu, Pengyuan Wang,
Jiashun Liu, Xiang Cheng, Kai Yang, Yangkun Chen, Saiyong Yang, Lan-Zhe Guo}
\hypersetup{
    pdftitle={Beyond Prompt Count: How Data Shapes Transfer in On-Policy Distillation},
    pdfauthor={Jiaxuan Wang, Jiafei Lyu, Yuchen Cai, Siye Wu, Pengyuan Wang, Jiashun Liu, Xiang Cheng, Kai Yang, Yangkun Chen, Saiyong Yang, Lan-Zhe Guo}
}

\AtBeginDocument{\ClearShipoutPicture\pagestyle{plain}}
\fancypagestyle{arxivfirst}{%
    \fancyhf{}
    \fancyhead[L]{\raisebox{0pt}[8pt][0pt]{%
        \includegraphics[height=16pt]{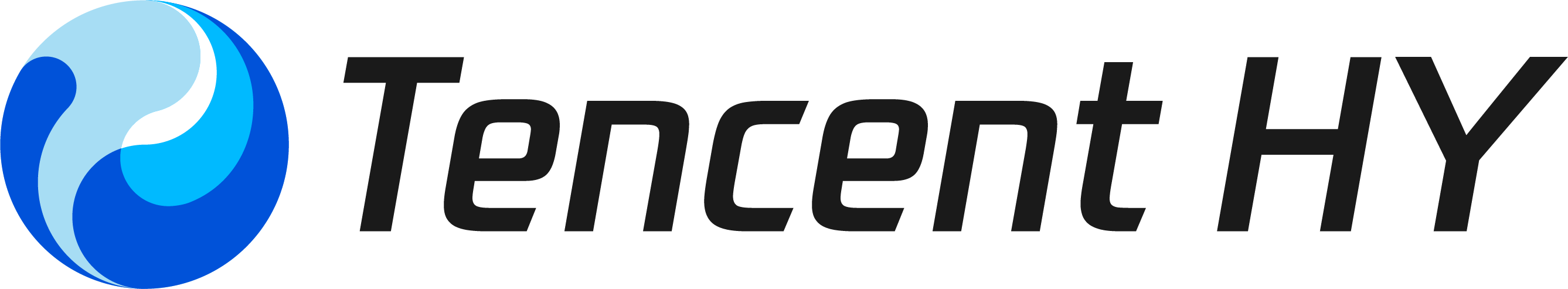}}}
    \fancyfoot[C]{\thepage}
    \renewcommand{\headrulewidth}{0.4pt}
    \renewcommand{\footrulewidth}{0pt}
}

\makeatletter
\renewcommand{\maketitle}{%
    \thispagestyle{arxivfirst}%
    \begingroup
    \centering
    \setlength{\parskip}{0pt}%
    \vspace*{-36pt}
    {\LARGE\scshape\@title\par}
    \vspace{10pt}
    {\fontsize{10}{15}\selectfont\bfseries
        Jiaxuan Wang\textsuperscript{1,2,3,*},\hspace{6pt}%
        Jiafei Lyu\textsuperscript{3},\hspace{6pt}%
        Yuchen Cai\textsuperscript{3},\hspace{6pt}%
        Siye Wu\textsuperscript{3},\hspace{6pt}%
        Pengyuan Wang\textsuperscript{3},\hspace{6pt}%
        Jiashun Liu\textsuperscript{3},\\
        Xiang Cheng\textsuperscript{4},\hspace{6pt}%
        Kai Yang\textsuperscript{3},\hspace{6pt}%
        Yangkun Chen\textsuperscript{3},\hspace{6pt}%
        Saiyong Yang\textsuperscript{3},\hspace{6pt}%
        Lan-Zhe Guo\textsuperscript{1,2,$\dagger$}\par}
    \vspace{6pt}
    {\fontsize{9}{11}\selectfont
    \textsuperscript{1}State Key Laboratory of Novel Software Technology, Nanjing University\\
    \textsuperscript{2}School of Intelligence Science and Technology, Nanjing University\\
    \textsuperscript{3}LLM Department, Tencent\qquad
    \textsuperscript{4}RUC\par}
    \vspace{6pt}
    \arxivresources
    \endgroup
    \begingroup
    \renewcommand{\thefootnote}{\fnsymbol{footnote}}%
    \let\@footnotetext\H@@footnotetext
    \footnotetext[1]{Work done while Jiaxuan Wang (%
        \href{mailto:wangjx@lamda.nju.edu.cn}{%
            \textcolor{referenceblue}{wangjx@lamda.nju.edu.cn}})
        was an intern at the Hunyuan Team, Tencent.}%
    \footnotetext[2]{Corresponding author:
        \href{mailto:guolz@nju.edu.cn}{%
            \textcolor{referenceblue}{guolz@nju.edu.cn}}.}%
    \endgroup
    \vspace{7pt}
}
\makeatother

\newsavebox{\arxivabstractcontent}
\renewenvironment{abstract}{%
    \par\addvspace{8pt}%
    \begin{lrbox}{\arxivabstractcontent}%
    \begin{minipage}{\dimexpr\linewidth-23pt\relax}%
    \setlength{\parindent}{0pt}\ignorespaces
}{%
    \par\end{minipage}\end{lrbox}%
    \noindent\begin{tikzpicture}
        \node[draw=abstractframe,fill=abstractback,line width=0.5pt,
            rounded corners=3pt,inner xsep=11pt,inner ysep=10pt,
            outer sep=0pt] (abstractbody) {\usebox{\arxivabstractcontent}};
        \node[fill=white,text=abstractlabel,font=\large\scshape,
            inner xsep=7pt,inner ysep=0pt,outer sep=0pt]
            at (abstractbody.north) {Abstract};
    \end{tikzpicture}\par
}

\newcommand{\arxivresources}{%
    \par\begingroup\centering\fontsize{10}{12}\selectfont
    \githubmark\hspace{3pt}\textbf{Code:}
    \href{https://github.com/wyy-1112/dissecting-opd}{%
        \textcolor{referenceblue}{https://github.com/wyy-1112/dissecting-opd}}%
    \par\endgroup
}

\begin{document}

\maketitle

\begin{abstract}
On-policy distillation (OPD) trains students using teacher feedback on
their own sampled responses, yet how prompt choice shapes transfer across
teacher--student pairs remains poorly understood. We systematically study
prompt quantity, source, and selection across RL- and SFT-continuation pairs
and cross-model settings. We find that OPD can be highly prompt-efficient:
a few prompts can approach large-pool performance, with four DAPO prompts
matching the observed mathematics score of 3,840 DeepMath prompts. However,
prompt utility is relational rather than intrinsic: changing only the
teacher can reverse the relative effectiveness of mathematics and code
prompts. To characterize these transfer differences, we analyze parameter
and functional changes across prompt supports and model pairs. Functional
alignment with the teacher varies across supports and target tasks; in
continuation pairs, teacher-aligned prediction changes can coexist with
weak parameter alignment. Continued OPD on effective supports can restore
performance after unfavorable transfer. Finally, targeted selection does not consistently outperform
uniform random sampling, and filtering out a source that performs poorly
alone yields no consistent gain across three paired support draws.
Overall, our results distinguish prompt efficiency from prompt
interchangeability and show that effective data choice depends on the
teacher--student pair and target capability, with random sampling providing
a competitive baseline in the studied settings.
\end{abstract}

\begingroup
\setlength{\intextsep}{15pt}
\begin{figure}[!ht]
    \centering
    \includegraphics[width=\linewidth]{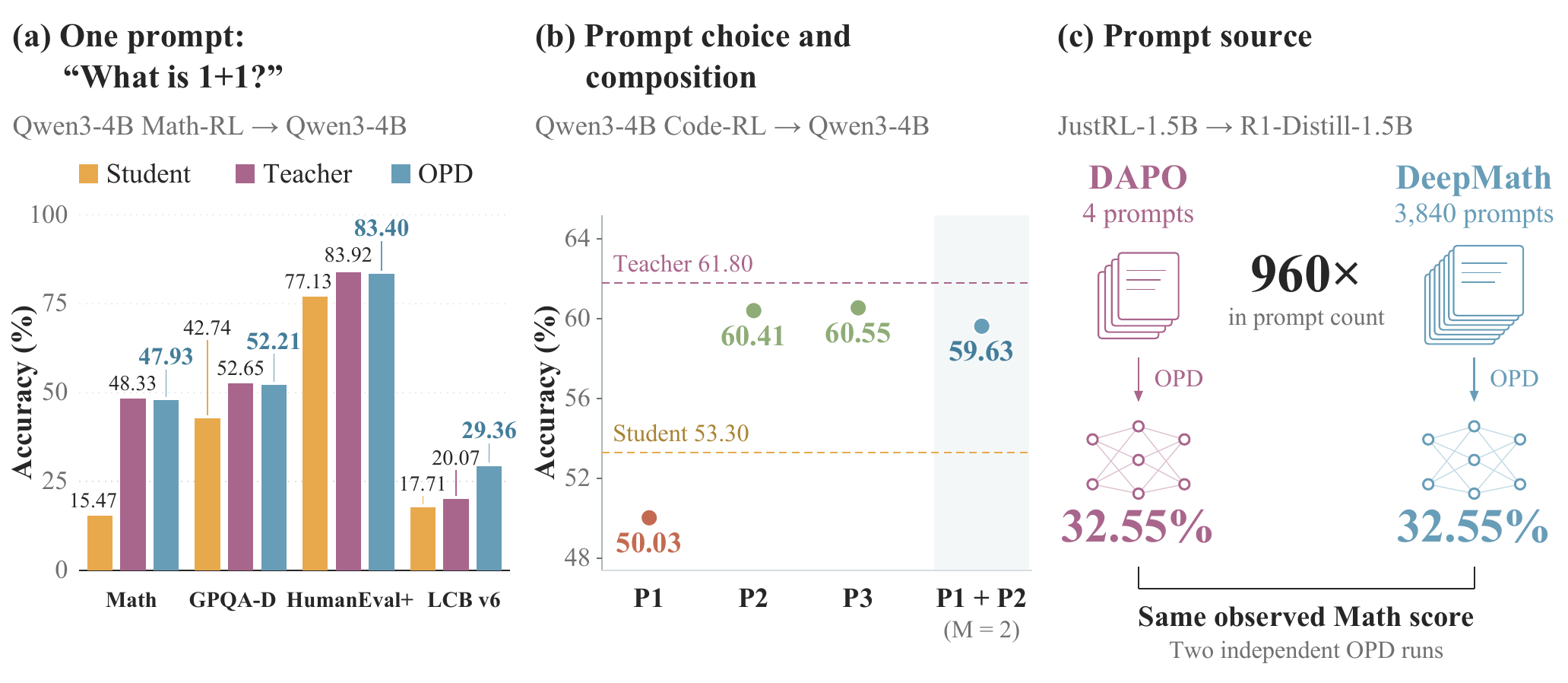}
    \caption{\textbf{Prompt efficiency does not imply prompt interchangeability.}
    A single prompt can induce substantial transfer, but outcomes vary with
    prompt identity and composition. Moreover, a few prompts from one source
    can match the observed mathematics score of thousands from another.
    These examples motivate our study of prompt count, source,
    teacher--student pairing, and selection.
    In (b), $P_1$--$P_3$ denote the individual code prompts described in
    \crossref{Appendix~\ref{app:figure1-code-prompts}}, while $P_1+P_2$ denotes joint
    OPD using the first two prompts.}
    \label{fig:overview}
\end{figure}
\endgroup

\clearpage
\section{Introduction}
\label{sec:introduction}

On-policy distillation (OPD) is increasingly used in large language model
(LLM) post-training to transfer capabilities to smaller students
\citep{yang2025qwen3,zheng2025hymt15}, consolidate domain-specialized teachers
\citep{xiao2026mimo,li2026katcoder,deepseek2026v4}, and recover capabilities
acquired in earlier training stages \citep{zeng2026glm5}.

In OPD, the student samples its own responses, and the teacher provides
token-level feedback at the prefixes visited by the student
\citep{lin2020autoregressive,agarwal2024onpolicy,lu2025onpolicydistillation}.
A fixed prompt can elicit different trajectories as the student evolves.
Prompt choice therefore shapes the contexts in which teacher supervision
is applied, motivating the study of both how many prompts are needed and
which prompts are useful.

Recent work improves transfer by aligning prompt content and templates
with the teacher's post-training data \citep{li2026rethinking}.
Concurrent work demonstrates substantial transfer from a single query
\citep{fu2026onetraining} and shows that OPD can propagate strengths and
weaknesses beyond the training domain, with generalization differing
between same-origin and cross-origin pairs \citep{li2026everycoin}.
\crossref{Appendix~\ref{app:related-work}} discusses related work in further detail.
These findings motivate a distinction between prompt sufficiency and
prompt interchangeability. Our central question is how prompt utility
depends on the teacher--student pair and target capability. A related
practical question is whether explicit prompt selection or source
filtering yields consistent gains over uniform random sampling.

We study prompt quantity, source, teacher--student pairing, and selection
across RL- and SFT-continuation pairs and cross-model settings.
\crossref{Figure~\ref{fig:overview}} illustrates both low prompt requirements and
sensitivity to prompt choice: four DAPO prompts match the observed
mathematics score of 3,840 DeepMath prompts, while individual code prompts
can produce positive or negative transfer.
\crossref{Figure~\ref{fig:data-preferences}(b)} further demonstrates teacher-dependent
source preferences: switching between Math-RL and Code-RL teachers reverses
the relative effectiveness of mathematics and code supports across all
four evaluations, with the initial student and both supports held fixed.
For multi-domain teachers, source preferences vary across target
capabilities, while broader mixtures at a fixed prompt count can improve
some capabilities at the expense of others.

To understand how different prompt supports shape transfer, we analyze
parameter updates, predictions on common prefixes, hidden-state
representations, and generated behavior. Functional alignment with the
teacher varies across supports and target tasks. When the teacher is
obtained by further training the student's initial checkpoint,
teacher-aligned functional changes can coexist with weak parameter
alignment. Hidden-state interventions further probe the role of
representations in unfavorable transfer, while continued OPD on effective
supports can restore performance.

This data sensitivity does not imply consistent gains from targeted
selection: uniform random sampling leads the JustRL--DeepMath study in
four-task average and remains competitive in the code and 30B-to-4B
comparisons. \crossref{Table~\ref{tab:mixed-pool-selection}} shows no consistent gain
from excluding GSM8K from a DeepMath/DAPO/GSM8K pool across three paired
support draws, despite its poor standalone performance.

Our contributions are threefold:
\begingroup
\setlength{\topsep}{4pt}
\setlength{\partopsep}{0pt}
\setlength{\itemsep}{2pt}
\setlength{\parsep}{0pt}
\begin{itemize}
\item \textbf{Data requirements and preferences.}
We systematically compare prompt counts and sources across teacher--student
pairs, including controlled teacher swaps with the student initialization
and prompt supports held fixed.

\item \textbf{Functional transfer, failure, and recovery.}
We combine parameter and functional diagnostics with hidden-state
interventions and continued distillation to characterize unfavorable
transfer and performance recovery.

\item \textbf{Practical data selection.}
We evaluate within-pool selection and mixed-pool source filtering against
uniform random sampling, examining both transfer performance and
selection cost.
\end{itemize}
\endgroup
\par\vfill

\section{Experimental Setup}
\label{sec:experimental-setup}

\subsection{On-policy distillation}

Let $\pi_T$ denote the frozen teacher and $\pi_\theta$ the student.
We sample prompts from a fixed support $\mathcal S_M$ of $M$ distinct
prompts and generate responses with the rollout policy $\pi_{\bar\theta}$.
OPD aligns student and teacher distributions at the resulting prefixes
$s_t=(x,y_{<t})$ using reverse KL \citep{agarwal2024onpolicy}:
\begin{equation}
    \mathcal L_{\mathrm{RKL}}(\theta;\bar\theta)
    =
    \mathbb E_{s\sim d_{\bar\theta}}
    \left[
        D_{\mathrm{KL}}\!\left(
            \pi_\theta(\cdot\mid s)\,\Vert\,\pi_T(\cdot\mid s)
        \right)
    \right],
    \label{eq:opd-rkl}
\end{equation}
where $d_{\bar\theta}$ is the induced distribution over valid response
prefixes. Collected prefixes and $\bar\theta$ remain fixed during
optimization. We use a sampled-token policy-gradient surrogate
\citep{lu2025onpolicydistillation}; implementation and training details
appear in \crossref{Appendix~\ref{app:experimental-details}}.

\subsection{Model pairs and data}

In a continuation pair, the teacher is obtained by RL or SFT from the
student's exact initial checkpoint; cross-model teachers are not
continuations of that checkpoint.
We use the Qwen3 \citep{yang2025qwen3} and DeepSeek-R1-Distill
\citep{guo2025deepseekr1} model families.
\crossref{Table~\ref{tab:study-pairs}} lists the pairs and data sources;
construction details appear in \crossref{Appendix~\ref{app:implementation}}.

\begingroup
\setlength{\intextsep}{6pt}
\begin{table}[!t]
    \setlength{\abovecaptionskip}{0pt}
    \setlength{\belowcaptionskip}{2pt}
    \caption{\textbf{Teacher--student pairs and OPD sources.}
    The first three teachers are trained by us; the others are publicly
    released models. We list constituent data sources; support construction,
    including mixtures and source filtering, is detailed in
    \crossref{Appendix~\ref{app:prompt-support-construction}}.}
    \label{tab:study-pairs}
    \centering
    \includegraphics[width=\linewidth,pagebox=cropbox,
        trim=18bp 13.3bp 18bp 52.3bp,clip]{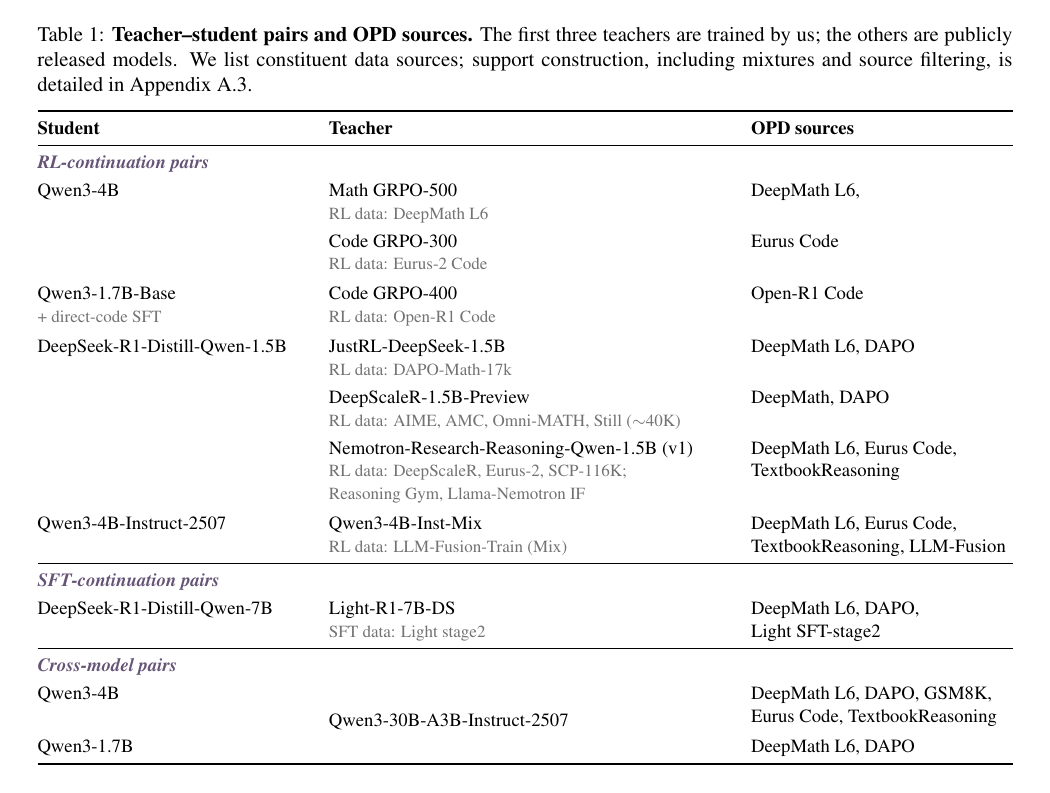}
\end{table}
\endgroup

The support $\mathcal S_M$ is selected from a candidate pool $\mathcal C$
of size $C$. Fresh responses are generated throughout training, so fixing
$M$ does not fix token or compute budgets. Support construction and
prompt templates are detailed in
\crossref{Appendices~\ref{app:prompt-support-construction} and~\ref{app:prompt-templates}}.

\subsection{Evaluation}

Math reports mean@16 averaged equally over AIME 2024/2025 and HMMT
February/November. GPQA-Diamond \citep{rein2024gpqa},
HumanEval+ \citep{liu2023evalplus}, and LiveCodeBench v6 \citep{jain2025livecodebench}
use mean@8, with MBPP+ added for code comparisons; direct-code 1.7B
comparisons use mean@16. Multi-domain evaluations additionally include
IFEval \citep{zhou2023ifeval}, IFBench \citep{pyatkin2025ifbench},
and BFCL-v3 \citep{patil2025bfcl}.
Mean@$k$ averages per-question correctness over $k$ responses;
macro averages are defined with the results. Decoding and scoring
details appear in \crossref{Appendix~\ref{app:evaluation-protocols}}.

For random selection, we report mean $\pm$ sample SD across
independent support draws, varying selection seeds while holding
the training seed fixed.

\section{Data Requirements}
\label{sec:data-requirements}
\label{sec:prompt-count}

\begin{figure}[!t]
    \centering
    \includegraphics[width=\linewidth]{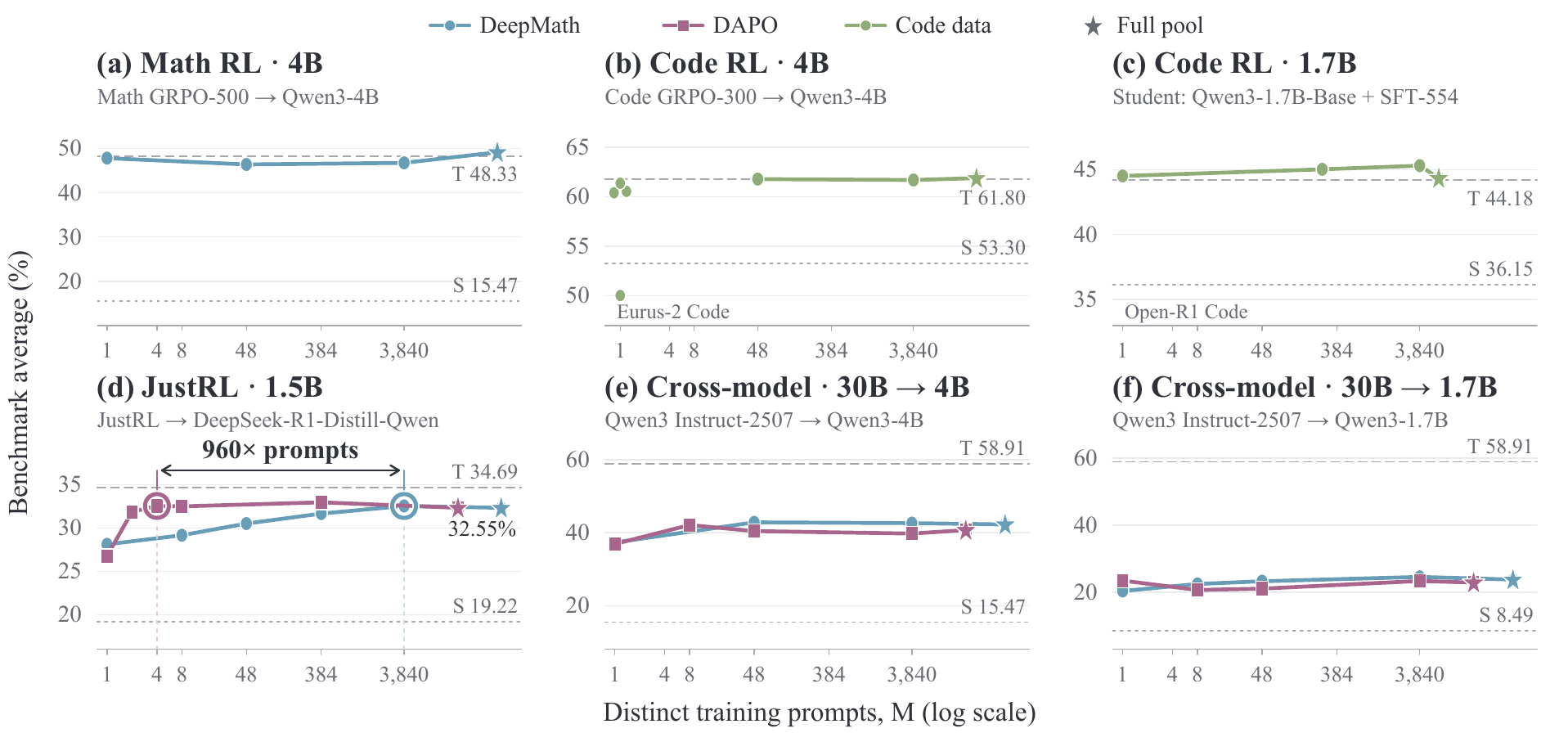}
    \caption{\textbf{Prompt-count scaling across six teacher--student pairs.}
    Stars mark full-pool OPD; S/T denote the initial student and teacher.
    Circled points in (d) attain the same observed Math score with
    4 DAPO versus 3,840 DeepMath prompts.
    Training budgets are comparison-specific; full results appear in
    \crossref{Appendix~\ref{app:prompt-count-results}}.}
    \label{fig:prompt-scaling}
\end{figure}

\crossref{Figure~\ref{fig:prompt-scaling}} examines how OPD performance varies
with the number of distinct training prompts.

\subsection{Few prompts can approach large-pool performance}
\label{sec:pair-scaling}

In the Qwen3-4B Math-RL pair, one prompt reaches $44.74$ on Math,
versus $46.77$ with 3,840 prompts, at a matched total of 3,840 rollouts.
With Code-RL, 48 prompts come within $0.10$ points of full-pool OPD
with 22,618 prompts at step 100.
Training curves are provided in
\crossref{Appendix~\ref{app:support-size-training-curves}}.

This pattern also appears beyond RL continuation.
In 30B-Instruct-to-4B distillation, DeepMath48 performs comparably
to DeepMath3840 at step 15.
For the Light-R1 SFT-continuation pair, eight DAPO or DeepMath prompts
achieve math scores of $40.83$ and $40.63$, respectively, compared
with $40.99$ using the original 3,533-prompt source.

However, selected single-prompt Code-RL runs range from $50.03$
to $61.37$, spanning the initial student's $53.30$.
We examine these unfavorable outcomes and their recovery in
\crossref{Section~\ref{sec:transfer-failure-recovery}}.

\subsection{Prompt-count scaling depends on source and pair}
\label{sec:source-scaling}

Full-pool performance need not predict sensitivity to prompt count.
With JustRL, DAPO and DeepMath achieve similar full-pool scores
but diverge at $M=8$.
\crossref{Appendix~\ref{app:prompt-count-results}} reports the same observed Math
score of $32.55$ for four DAPO and 3,840 DeepMath prompts,
a $960\times$ difference in distinct prompt count.

Scaling also varies across students with the same teacher and source.
With the 30B Instruct teacher and DAPO, the 4B student's score rises
from $36.9$ at $M=1$ to $42.1$ at $M=8$, whereas the 1.7B student's
score remains nearly unchanged from $M=1$ to $M=3{,}840$.

Since repeated prompts still generate fresh trajectories,
\crossref{Appendix~\ref{app:trajectory-reuse}} separately examines the quality--cost
trade-offs of reusing those trajectories.

\begin{takeaway}{1}
Few prompts can support effective OPD, but prompt sufficiency does not imply
prompt interchangeability.
\end{takeaway}

\begingroup
\setlength{\parskip}{4pt}
\section{Data Preferences across Teacher--Student Pairs}
\label{sec:data-preferences}

At a fixed support size of $M=48$, we compare prompt sources
and mixtures within each teacher--student pair.

\begin{figure}[!t]
    \centering
    \setlength{\abovecaptionskip}{4pt}
    \includegraphics[width=\linewidth]{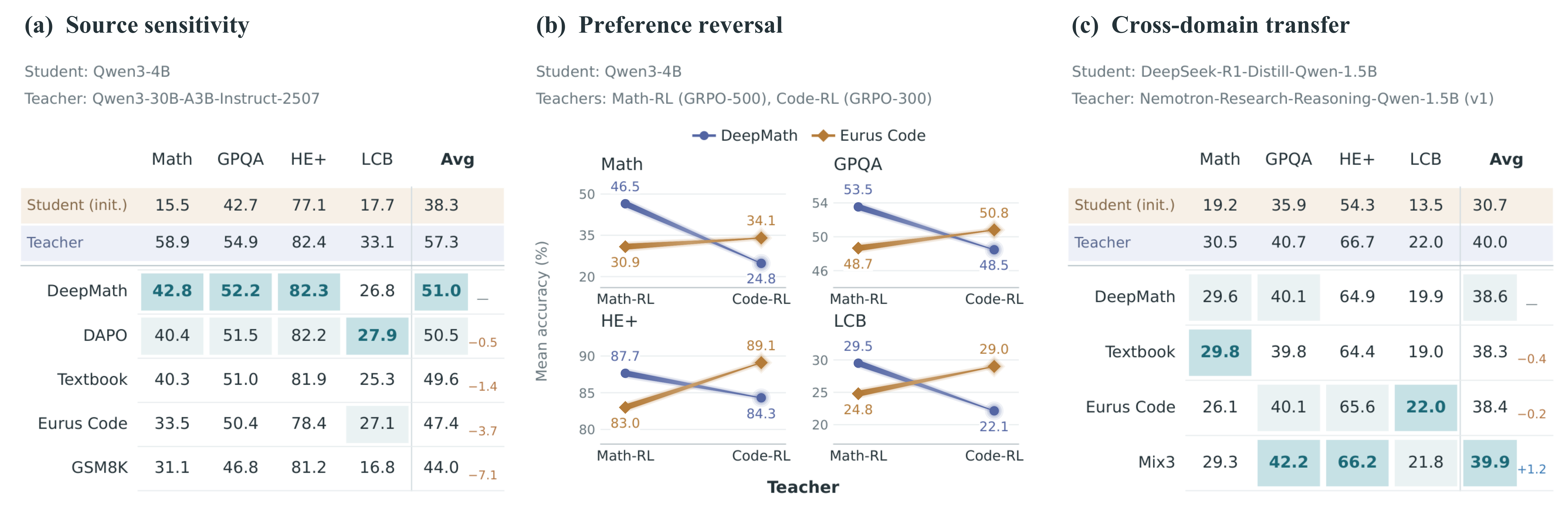}
    \caption{\textbf{Data-source preferences across teacher--student pairs.}
    \textbf{(a)} Source sensitivity with a 30B Instruct teacher.
    \textbf{(b)} Preference reversals between Math-RL and Code-RL teachers
    with the same initial student.
    \textbf{(c)} Cross-domain transfer from the Nemotron teacher.
    Avg is the unweighted mean of the four displayed benchmark groups;
    side annotations report differences from DeepMath. Darker and lighter
    shading mark the highest and second-highest OPD scores within each column.}
    \label{fig:data-preferences}
    \label{fig:fixed-support}
\end{figure}

\subsection{Source sensitivity at fixed support size}
\label{sec:source-preferences}

\begingroup
\setlength{\intextsep}{5pt}
\setlength{\columnsep}{12pt}
\begin{wraptable}{r}{0.40\linewidth}
    \centering
    \small
    \setlength{\tabcolsep}{5pt}
    \renewcommand{\arraystretch}{1.08}
    \setlength{\abovecaptionskip}{4pt}
    \setlength{\belowcaptionskip}{4pt}
    \caption{\textbf{Token control.} $M=48$; Avg averages Math, GPQA, HE+, and LCB.}
    \label{tab:source-token-match}
    \begin{tabular}{lrr}
        \toprule
        Data & Tokens (M) & Avg \\
        \midrule
        GSM8K & 1.19 & 44.01 \\
        \rowcolor[HTML]{F1F2F4}
        \hspace{0.5em}+ token match & 16.14 & 44.84 \\
        DeepMath & 16.63 & 51.02 \\
        \bottomrule
    \end{tabular}
\end{wraptable}

With equal prompt counts, DeepMath and GSM8K supports score $42.8$
and $31.1$ on Math, respectively, for a Qwen3-4B student and 30B Instruct
teacher in \crossref{Figure~\ref{fig:data-preferences}(a)}.

\crossref{Table~\ref{tab:source-token-match}} and \crossref{Appendix~\ref{app:token-match-breakdown}}
show GSM8K trailing DeepMath by $11.25$ points on Math and $9.36$ on LCB
despite approximately matched cumulative supervision-token budgets.
Supervision volume alone therefore does not explain the source gap.
\par
\endgroup

\subsection{Teacher-dependent source preferences}
\label{sec:teacher-dependent-preferences}

Switching from the Math-RL to Code-RL teacher reverses source
preferences on all four evaluations in \crossref{Figure~\ref{fig:data-preferences}(b)},
with the Qwen3-4B initialization and both 48-prompt supports fixed.
On Math, DeepMath outperforms Eurus Code under the Math-RL teacher
($46.51$ versus $30.89$), whereas Eurus Code outperforms DeepMath
under the Code-RL teacher ($34.06$ versus $24.84$).
The preferred source thus depends on the teacher even for the same
student and target task.
In Avg$_4$, the source-preference reversal in
\crossref{Figure~\ref{fig:data-preferences}(b)} persists across all five tested
support--training seed configurations in
\crossref{Appendix~\ref{app:teacher-source-training-curves}}, with a teacher--source
difference-in-differences of $13.1$--$14.1$ percentage points.

\subsection{Task-specific transfer and mixture trade-offs}
\label{sec:multidomain-preferences}
\label{sec:mixture-preferences}

Data composition also changes which capabilities benefit.
With the Mix-RL teacher, Code48's GPQA scores span $61.24$--$61.81$
across three support draws, versus DeepMath48's $47.35$ in
\crossref{Table~\ref{tab:mixrl-data-preferences}}.
Even within instruction following, IF48 improves IFBench over
the initial student while lowering IFEval.

\begin{table}[!t]
    \centering
    \begingroup
    \definecolor{prefStudentBg}{HTML}{F6EFE5}
    \definecolor{prefTeacherBg}{HTML}{EAEFF8}
    \definecolor{prefBestBg}{HTML}{C3E0E5}
    \definecolor{prefSecondBg}{HTML}{E9F2F3}
    \definecolor{prefBestText}{HTML}{1D7282}
    \definecolor{prefStudentText}{HTML}{8B6C48}
    \definecolor{prefTeacherText}{HTML}{5E6F9E}
    \newcommand{\prefBest}[1]{\cellcolor{prefBestBg}\textcolor{prefBestText}{\textbf{#1}}}
    \newcommand{\prefSecond}[1]{\cellcolor{prefSecondBg}#1}
    \small
    \setlength{\tabcolsep}{3.5pt}
    \renewcommand{\arraystretch}{1.05}
    \setlength{\abovecaptionskip}{0pt}
    \setlength{\belowcaptionskip}{4pt}
    \caption{\textbf{Data-source preferences with a multi-domain RL teacher.}
    Qwen3-4B-Inst-Mix $\rightarrow$ Qwen3-4B-Instruct-2507; $M=48$.
    Teal marks the best and second-best OPD scores;
    mixture construction is detailed in \crossref{Appendix~\ref{app:prompt-support-construction}}.}
    \label{tab:mixrl-data-preferences}
    \begin{tabular}{@{}l*{8}{r}@{}}
        \toprule
        \textbf{Model / OPD support} & \textbf{Math} & \textbf{GPQA-D}
        & \textbf{HumanEval+} & \textbf{LCB v6} & \textbf{IFEval}
        & \textbf{IFBench} & \textbf{BFCL-v3} & \textbf{Avg}$_6$ \\
        \midrule
        \rowcolor{prefStudentBg}
        \textcolor{prefStudentText}{Student (init.)}
        & 45.83 & 45.52 & 81.48 & 27.86 & 83.71 & 30.42 & 63.91 & 53.61 \\
        \rowcolor{prefTeacherBg}
        \textcolor{prefTeacherText}{Teacher}
        & 48.18 & 45.27 & 79.57 & 30.21 & 86.02 & 39.96 & 70.81 & 56.17 \\
        \midrule
        DeepMath & 48.28 & 47.35 & 81.86 & 29.71 & \prefSecond{83.90}
        & 29.04 & 64.25 & 54.65 \\
        Textbook & 46.46 & 51.14 & \prefBest{83.23} & 28.93 & 83.50
        & 27.88 & 62.66 & 54.69 \\
        Eurus Code & \prefSecond{48.49} & \prefBest{61.81} & 81.17
        & \prefSecond{31.93} & 83.13 & 28.67 & \prefSecond{66.50} & \prefBest{57.63} \\
        IF & 41.51 & 46.46 & 68.29 & 22.07 & 78.35 & \prefBest{42.56} & 53.25 & 48.67 \\
        Agent & 43.38 & 50.88 & 78.35 & 28.21 & 82.99 & 26.73 & 62.28 & 52.99 \\
        \midrule
        Mix3 & \prefBest{48.59} & \prefSecond{59.34} & \prefSecond{82.01}
        & \prefBest{32.36} & \prefBest{83.93} & 28.79 & 64.28 & 57.16 \\
        Mix5 & 47.55 & 55.18 & 80.18 & 31.14 & 82.05 & \prefSecond{36.79}
        & \prefBest{69.56} & \prefSecond{57.17} \\
        \bottomrule
    \end{tabular}
    \endgroup
\end{table}

Mixtures likewise improve some tasks at the expense of others.
Under Nemotron, Mix3 improves GPQA, HumanEval+, and LCB over DeepMath
in \crossref{Figure~\ref{fig:data-preferences}(c)}, with a small Math decrease.
For Mix-RL, Mix5 reallocates the same 48-prompt budget across five
domains, reselecting prompts within each pool as in
\crossref{Appendix~\ref{app:prompt-support-construction}}.
Relative to Mix3, it improves IFBench by $8.00$ points and BFCL
by $5.28$ points, while GPQA decreases by $4.16$ points.

\vspace{-6pt}
\begin{takeaway}{2}
Source preferences are relational: the same source can be effective or
ineffective depending on the teacher--student pair and target capability,
while broader mixtures redistribute transfer gains.
\end{takeaway}
\vspace{-6pt}
\endgroup

\begingroup
\raggedbottom
\section{Transfer Analysis}
\label{sec:transfer-failure-recovery}
\label{sec:mechanism}

We examine how prompt supports shape the student's functional and
behavioral changes, whether unfavorable transfer can be repaired, and
what these results imply for prompt selection.

\subsection{Functional alignment}
\label{sec:functional-transfer}
\label{sec:data-shapes-transfer}

For continuation pairs, we compare OPD and teacher changes from the
shared student initialization. Parameter alignment measures weight
displacements; functional alignment measures token log-probability
changes on common frozen prefixes, defined in
\crossref{Appendix~\ref{app:functional-measurement}}.
Functional cosine measures directional alignment; $\alpha$ is the
teacher-direction projection coefficient. Diagnostic KL closure measures
the reduction in the conditional KL gap to the teacher.

\crossref{Figure~\ref{fig:functional-changes}(a)} reports median parameter and
functional cosines of $0.069$ and $0.859$ across 43 checkpoints from
seven teacher families, including intermediate states.
OPD can therefore reproduce teacher-aligned prediction changes without
following the teacher's parameter-update direction.

\begin{figure}[!htb]
    \centering
    \setlength{\abovecaptionskip}{4pt}
    \includegraphics[width=\linewidth]{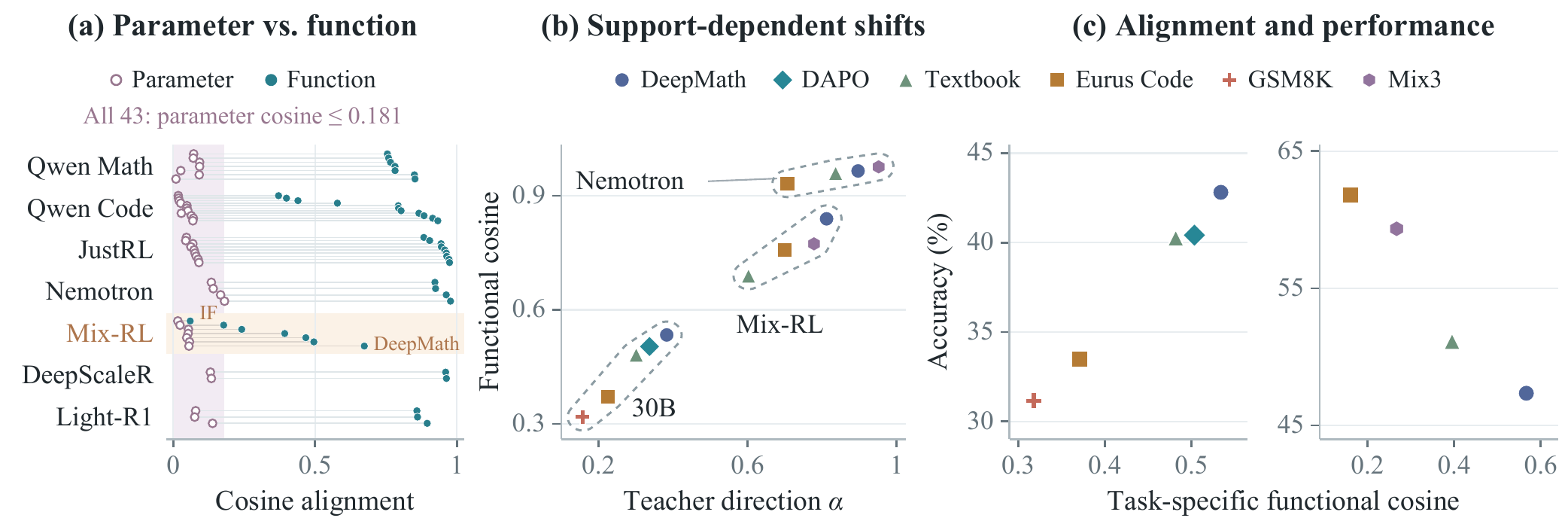}
    \caption{\textbf{Parameter and functional alignment.}
    \textbf{(a)} Parameter and all-domain functional cosines for 43 checkpoints;
    each line joins the same endpoint.
    \textbf{(b)} Teacher-direction projection $\alpha$ versus functional cosine
    on Math probes, grouped by model pair.
    \textbf{(c)} Functional cosine versus accuracy: 30B-to-4B Math (left)
    and Mix-RL-to-4B GPQA (right).
    Definitions: \crossref{Appendix~\ref{app:functional-measurement}}.}
    \label{fig:functional-changes}
\end{figure}

We next examine support-dependent prediction changes across tasks in
\crossref{Figure~\ref{fig:functional-changes}(b,c)}.
In 30B-to-4B distillation, DeepMath reduces the diagnostic teacher--student
KL gap on frozen Math prefixes by $11.20\%$, whereas GSM8K widens it
by $13.64\%$. GSM8K nevertheless has positive functional cosine
($0.318$), showing that directional alignment alone does not ensure
closer predictions. Eurus Code closes only $1.13\%$ of the Math gap but
$42.30\%$ of the LCB gap.
\crossref{Appendix~\ref{app:source-functional-diagnostics}} details these diagnostics,
which measure task-specific prediction changes, not benchmark gains.

\begin{takeaway}{3}
OPD can produce teacher-aligned functional changes through substantially
different parameter updates; the extent of alignment depends on the support
and task.
\end{takeaway}

\subsection{Generation behavior}
\label{sec:rl-behavior}

\begingroup
\setlength{\intextsep}{5pt}
\setlength{\columnsep}{12pt}
\begin{wrapfigure}{r}{\dimexpr.5\linewidth-.5\columnsep\relax}
    \centering
    \setlength{\parskip}{0pt}
    \includegraphics[width=\linewidth]{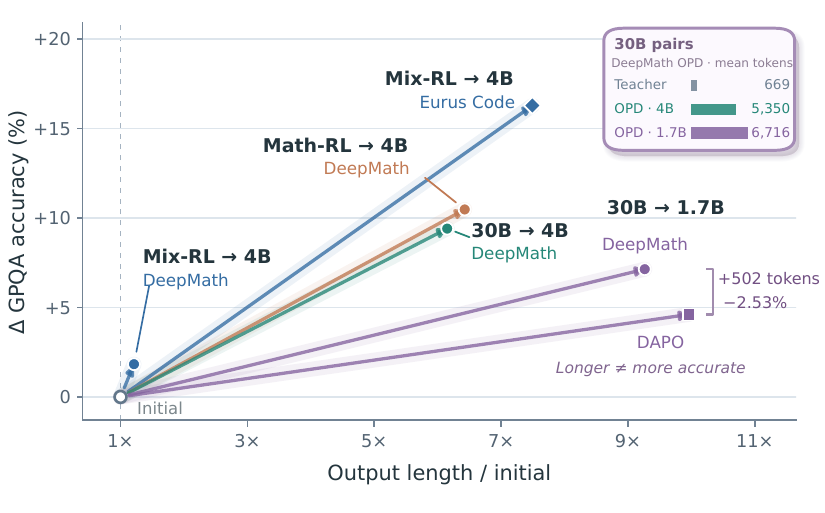}\par
    \nointerlineskip
    \setlength{\abovecaptionskip}{2pt}
    \caption{\textbf{GPQA accuracy and output-length changes.}
    All supports contain $M=48$ prompts; changes are relative
    to each pair's initial student. Inset: mean output tokens.}
    \label{fig:accuracy-length}
\end{wrapfigure}

We next move from common-prefix predictions to freely generated responses
in \crossref{Figure~\ref{fig:accuracy-length}}. With Mix-RL, Eurus Code yields longer
GPQA responses and higher accuracy than DeepMath; two additional Code
support draws reproduce this pattern. DeepMath OPD also produces longer
responses than the teacher in both 30B pairs, so successful transfer need
not imitate the teacher's response length. Yet in 30B-to-1.7B distillation,
DAPO produces longer responses than DeepMath with lower accuracy.
Longer generation thus accompanies some gains but does not consistently
indicate better transfer.

For Mix-RL Code48 and Mix3, most additional text precedes the final answer.
Among the $15$--$16\%$ of outputs with an identifiable early answer,
correct-to-wrong revisions outnumber wrong-to-correct revisions.
The results link supports to generation behavior
without establishing that extra tokens cause accuracy gains.
Complete accuracy--length measurements and output analyses are in
\crossref{Table~\ref{tab:accuracy-length}} and \crossref{Appendix~\ref{app:generation-behavior}}.
\par
\endgroup

\subsection{Failure and recovery}
\label{sec:pool-switch-recovery}

Can interventions redirect these support-dependent changes?
In the Code pair, we compare a degraded endpoint $A'$, an effective
single-prompt endpoint $B$, and native M48 trained directly on 48 prompts.
\crossref{Figure~\ref{fig:transfer-failure-recovery}(a)} shows similar layerwise CKA
\citep{kornblith2019similarity} profiles for $A'$ and native M48 relative to full-pool OPD, despite their
performance gap.
We therefore test behavior directly through hidden-state replacement
and continued distillation.

\begin{figure}[!htb]
    \centering
    \setlength{\abovecaptionskip}{4pt}
    \includegraphics[width=\linewidth]{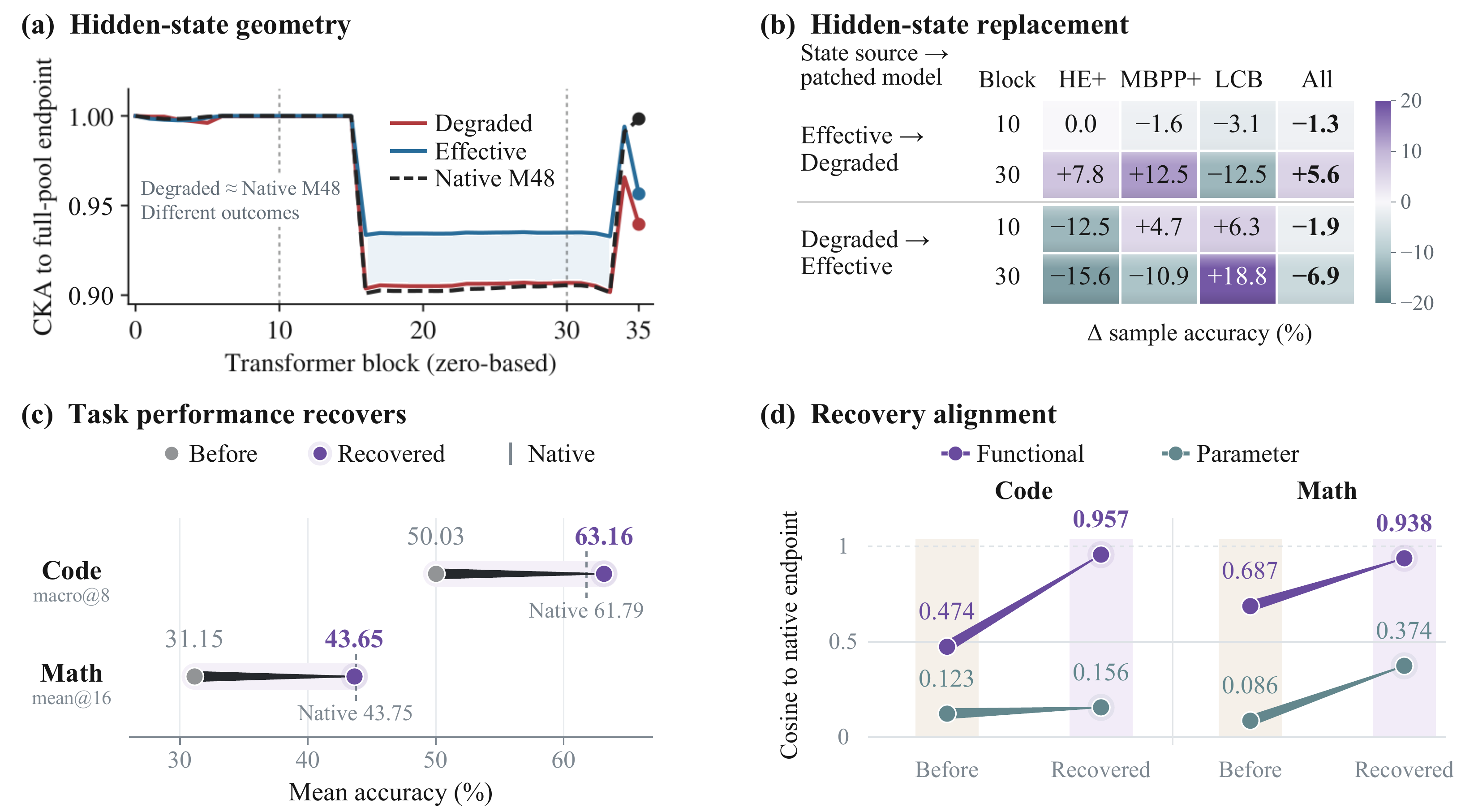}
    \caption{\textbf{Redirecting unfavorable transfer.}
    \textbf{(a)} CKA relative to full-pool OPD\@.
    \textbf{(b)} Bidirectional hidden-state replacement.
    \textbf{(c)} Performance recovery through continued distillation;
    vertical ticks mark native references.
    \textbf{(d)} Functional versus parameter alignment to native endpoints.
    Reference checkpoints and unequal Code update budgets are specified in
    \crossref{Appendix~\ref{app:recovery-protocols}}.}
    \label{fig:transfer-failure-recovery}
\end{figure}

At each decoding step, we replace the recipient's current-position hidden
vector after block 30 (zero-based) with the donor's.
On the 20-question probe in \crossref{Figure~\ref{fig:transfer-failure-recovery}(b)},
with eight samples per question, $B\to A'$ raises
accuracy from $18.75\%$ to $24.38\%$, while $A'\to B$ lowers it from
$28.13\%$ to $21.25\%$.
Self-replacement leaves accuracy unchanged; block 10 gives no aggregate
improvement. The tested representation interface can thus alter the
degraded model's behavior during evaluation.

\crossref{Figure~\ref{fig:transfer-failure-recovery}(c)} shows recovery through
continued OPD.
Switching the degraded Code model to an effective 48-prompt support
raises Code Macro from $50.03$ to $63.16$, above the initial student's
$53.30$. Switching from GSM8K to DeepMath also recovers performance
in the 30B-to-4B Math setting. Code recovery adds training and resets
the optimizer; it does not match the native reference's update budget.

\paragraph{Continuation data at matched update counts.}
Does recovery reflect additional training alone, or does the continuation
support matter? \crossref{Figure~\ref{fig:support-switch-main}} compares four
paths in 30B-to-4B distillation. Each first-stage checkpoint, trained for
15 updates on GSM8K or DeepMath, branches into a further 15 updates on
either source. All four continuations reset Adam and retain the same
learning rate, batch size, and training seed. Thus, additional updates
and optimizer resets are shared across the paired arms, although token
budgets remain unmatched.

Continuing on DeepMath rather than GSM8K improves formal Avg$_4$ by
$7.98$ points from the GSM8K-trained checkpoint and $5.60$ points from
the DeepMath-trained checkpoint. The two DeepMath continuations reach
close endpoint scores despite different histories. The validation curves
show how the paths separate during continuation; their AIME24/25
mean@2 metric is distinct from the formal four-task endpoint evaluation.
The continuation support therefore remains consequential at matched
update counts, rather than recovery being explained by extra updates alone.
\crossref{Appendix~\ref{app:support-switch-controls}} provides the full protocol
and supplementary parameter comparisons.

\begin{figure}[t]
    \centering
    \includegraphics[width=0.90\linewidth]{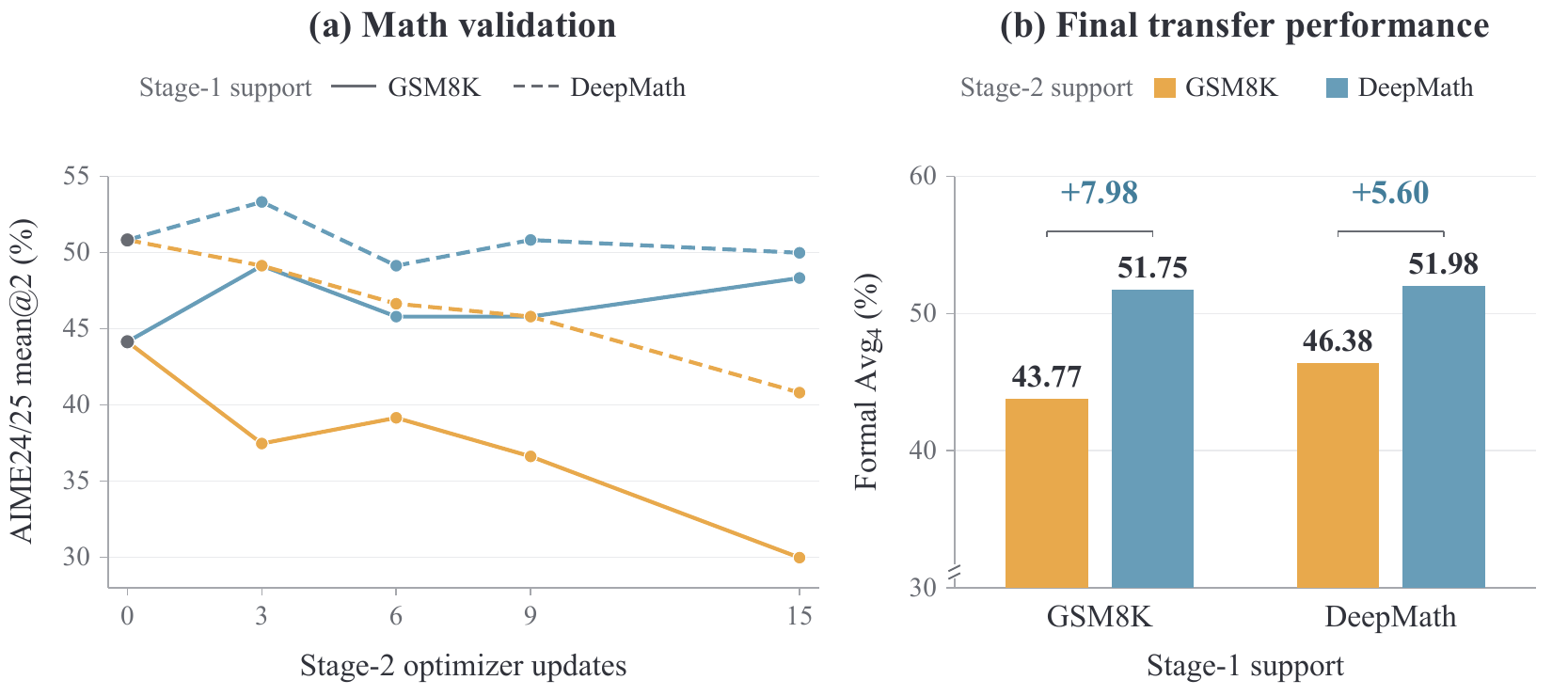}
    \caption{\textbf{Continuation data shapes recovery at matched update counts.}
    30B Instruct $\to$ Qwen3-4B, with 15 updates per stage and a fresh Adam
    optimizer in every second-stage run; token budgets are not matched.
    Orange and blue indicate stage-2 GSM8K and DeepMath, respectively.
    \textbf{(a)} AIME24/25 validation mean@2; solid/dashed lines start from
    GSM8K/DeepMath checkpoints. Gray step-0 markers show the shared starting
    scores of $44.17$/$50.83$.
    \textbf{(b)} Formal Avg$_4$ (\%), with the y-axis truncated at $30\%$,
    grouped by stage-1 source; annotations
    show the gains from using DeepMath rather than GSM8K in stage 2.}
    \label{fig:support-switch-main}
\end{figure}

\crossref{Figure~\ref{fig:transfer-failure-recovery}(d)} shows strong functional but
weak parameter alignment between recovered models and native references.
These results show that unfavorable training histories need not prevent
strong subsequent transfer, even when parameter updates remain different.
\crossref{Appendix~\ref{app:recovery-protocols}} further shows that Math recovery
persists under alternative answer scoring.

\begin{takeaway}{4}
Unfavorable OPD outcomes are recoverable: representation
interventions can redirect behavior, while continued OPD on effective
supports can restore performance without retracing the parameter path of
successful training.
\end{takeaway}

\subsection{Prompt selection}
\label{sec:prompt-selection}

These analyses characterize transfer after training. We now test
whether selecting prompts beforehand improves transfer over
uniform random sampling.
\crossref{Table~\ref{tab:prompt-selection}} compares eight JustRL--DeepMath selectors.
Uniform random sampling achieves the highest reported four-task average
at both $M=8$ and $M=48$ ($37.65$ and $37.63$), without additional GPU screening.
Other selectors lead on individual tasks; method definitions are in
\crossref{Appendix~\ref{app:prompt-selection-methods}}.

\begin{table}[t]
    \centering
    \setlength{\abovecaptionskip}{0pt}
    \caption{\textbf{Prompt selection within DeepMath.}
    JustRL-1.5B $\rightarrow$ DS-1.5B. Random selectors report mean $\pm$ sample SD
    over three sampled supports. Bold marks the column best within each $M$.
    Costs estimate selection-only GPU hours; $0$ denotes no GPU screening.}
    \label{tab:prompt-selection}
    \label{tab:prompt_selection}
    \begingroup
    \small
    \setlength{\tabcolsep}{2.8pt}
    \renewcommand{\arraystretch}{1.00}
    \newcommand{\selcell}[2]{%
        \makebox[22.5pt][r]{#1}%
        \makebox[25pt][l]{%
            \if\relax\detokenize{#2}\relax\else
                {\fontsize{8}{9}\selectfont\textcolor[HTML]{596579}{\,$\pm$\,#2}}%
            \fi}%
    }
    \begin{tabular*}{\linewidth}{@{\extracolsep{\fill}}lcccccr@{}}
        \toprule
        Method & Math & GPQA-D & HE+ & LCB & Avg$_4$ & GPU$\cdot$h \\
        \midrule
        \addlinespace[1pt]
        \multicolumn{7}{@{}l}{$M=8,\quad C=384$} \\
        \addlinespace[3pt]
        Uniform random
            & \selcell{\textbf{31.09}}{0.50}
            & \selcell{39.44}{1.24}
            & \selcell{62.40}{0.32}
            & \selcell{17.67}{0.86}
            & \selcell{\textbf{37.65}}{0.38} & 0 \\
        Stratified random
            & \selcell{30.36}{0.39}
            & \selcell{38.19}{0.58}
            & \selcell{61.74}{0.46}
            & \selcell{17.93}{0.41}
            & \selcell{37.06}{0.25} & 1.1 \\
        Semantic diversity & \selcell{30.21}{} & \selcell{39.46}{} & \selcell{62.88}{} & \selcell{17.36}{} & \selcell{37.48}{} & $<0.1$ \\
        Hard selection & \selcell{28.96}{} & \selcell{38.07}{} & \selcell{\textbf{63.57}}{} & \selcell{\textbf{18.21}}{} & \selcell{37.20}{} & 1.1 \\
        Shortest & \selcell{28.75}{} & \selcell{39.08}{} & \selcell{61.28}{} & \selcell{16.86}{} & \selcell{36.49}{} & 2.2 \\
        Longest & \selcell{29.82}{} & \selcell{\textbf{40.12}}{} & \selcell{62.48}{} & \selcell{17.14}{} & \selcell{37.39}{} & 2.2 \\
        T--S disagreement & \selcell{28.53}{} & \selcell{38.64}{} & \selcell{62.91}{} & \selcell{17.68}{} & \selcell{36.94}{} & 3.3 \\
        Cost-D-opt & \selcell{28.91}{} & \selcell{39.90}{} & \selcell{62.27}{} & \selcell{17.43}{} & \selcell{37.13}{} & 7.5 \\
        \midrule
        \addlinespace[1pt]
        \multicolumn{7}{@{}l}{$M=48,\quad C=2{,}304$} \\
        \addlinespace[3pt]
        Uniform random
            & \selcell{29.84}{0.31}
            & \selcell{\textbf{41.02}}{0.48}
            & \selcell{62.40}{0.37}
            & \selcell{17.26}{0.28}
            & \selcell{\textbf{37.63}}{0.21} & 0 \\
        Stratified random
            & \selcell{\textbf{30.72}}{0.28}
            & \selcell{37.95}{0.42}
            & \selcell{61.58}{0.35}
            & \selcell{17.83}{0.29}
            & \selcell{37.02}{0.19} & 6.6 \\
        Semantic diversity & \selcell{29.16}{} & \selcell{38.83}{} & \selcell{61.72}{} & \selcell{17.25}{} & \selcell{36.74}{} & $<0.1$ \\
        Hard selection & \selcell{27.91}{} & \selcell{39.42}{} & \selcell{62.15}{} & \selcell{\textbf{18.04}}{} & \selcell{36.88}{} & 6.6 \\
        Shortest & \selcell{29.38}{} & \selcell{40.56}{} & \selcell{61.94}{} & \selcell{16.88}{} & \selcell{37.19}{} & 15.2 \\
        Longest & \selcell{30.25}{} & \selcell{40.64}{} & \selcell{61.85}{} & \selcell{17.22}{} & \selcell{37.49}{} & 15.2 \\
        T--S disagreement & \selcell{28.84}{} & \selcell{39.15}{} & \selcell{62.18}{} & \selcell{17.63}{} & \selcell{36.95}{} & 16.3 \\
        Cost-D-opt & \selcell{30.45}{} & \selcell{38.74}{} & \selcell{\textbf{63.12}}{} & \selcell{16.93}{} & \selcell{37.31}{} & 47.7 \\
        \bottomrule
    \end{tabular*}
    \endgroup
\end{table}

Additional comparisons cover Code-RL-to-4B and 30B-to-4B distillation.
Semantic diversity leads on
Code Macro and stratified random on 30B-to-4B Avg$_4$, while uniform random
remains competitive without GPU screening. Higher screening cost does
not consistently improve transfer. Complete benchmark scores appear in
\crossref{Appendices~\ref{app:code-prompt-selection}
and~\ref{app:crossmodel-prompt-selection}}.

Initial-gradient diagnostics in \crossref{Appendix~\ref{app:initial-gradient-representativeness}}
may help explain random sampling's competitiveness.
In JustRL--DeepMath ($C=384$, $M=8$), random
supports preserve the candidate-pool mean direction (mean cosine $0.971$;
fifth percentile $0.952$). However, \crossref{Figure~\ref{fig:selection-gradient}}
shows that lower gradient approximation error does not consistently
predict higher final accuracy.

\paragraph{Mixed pools and source composition.}
Does the weak standalone performance of a source justify filtering it
before sampling from a mixed pool?
\crossref{Table~\ref{tab:mixed-pool-selection}} shows no consistent gain from removing
GSM8K across three paired draws, despite its weak standalone performance.
Both arms use $M=48$ prompts from a DeepMath/DAPO/GSM8K candidate pool,
with filtering removing GSM8K before selection. The paired Math and OOD
differences change direction across draws.

\begin{figure}[t]
    \centering
    \begin{minipage}[c]{0.54\linewidth}
            \makeatletter\def\@captype{table}\makeatother
    \centering
    \setlength{\abovecaptionskip}{0pt}
    \setlength{\belowcaptionskip}{6pt}
    \hyphenpenalty=10000
    \caption{\raggedright\textbf{Source filtering in a mixed pool.}
    \mbox{30B Instruct} $\to$ 4B, $M=48$.
    Scores (\%) are mean $\pm$ SD across three support draws.}
    \label{tab:mixed-pool-selection}
    \begingroup
    \small
    \setlength{\tabcolsep}{2pt}
    \renewcommand{\arraystretch}{1.30}
    \newcommand{\mixsd}[2]{%
        \makebox[20pt][r]{#1}%
        \makebox[23pt][l]{{\fontsize{8}{9}\selectfont\textcolor[HTML]{596579}{\,$\pm$\,#2}}}%
    }
    \begin{tabular*}{\linewidth}{@{}l@{\extracolsep{\fill}}cc@{}}
        \toprule
        Sampling strategy & Math mean@16 & OOD mean@8 \\
        \midrule
        Mixed random & \mixsd{40.7}{2.3} & \mixsd{54.7}{0.7} \\
        Source-filtered random & \mixsd{40.5}{1.0} & \mixsd{54.5}{0.8} \\
        \midrule[0.3pt]
        Filtered $-$ Mixed & \mixsd{$-$0.3}{2.0} & \mixsd{$-$0.2}{0.9} \\
        \bottomrule
    \end{tabular*}
    \endgroup

    \end{minipage}\hfill
    \begin{minipage}[c]{0.43\linewidth}
        \setlength{\abovecaptionskip}{3pt}
        \input{sections/figure_selection_gradient}
    \end{minipage}
\end{figure}

\par\addvspace{8pt}\noindent
\begin{minipage}{\linewidth}
    \makeatletter\def\@captype{figure}\makeatother
    \centering
    \includegraphics[width=\linewidth,trim=0bp 18bp 0bp 0bp,clip]{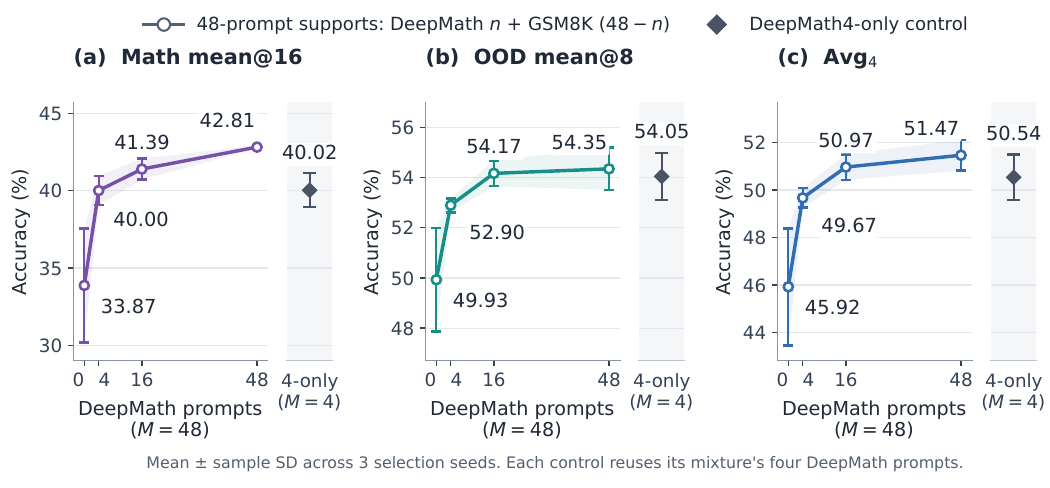}
    \caption{\textbf{Source composition and a four-prompt control.}
    30B Instruct $\to$ Qwen3-4B, 25 updates.
    DeepMath/GSM8K composition sweep (circles, $M=48$) and DeepMath4-only
    control (diamonds, $M=4$). Within each seed, the control reuses exactly
    the four DeepMath prompts in the corresponding 4/44 mixture.
    Error bars and bands show sample SD across selection seeds 42/43/44.}
    \label{fig:mixture-composition}
\end{minipage}\par\addvspace{8pt}

\crossref{Figure~\ref{fig:mixture-composition}} probes this result through a
DeepMath/GSM8K composition sweep and a paired four-prompt control.
Replacing four of 48 GSM8K prompts with DeepMath raises mean Math from
$33.87$ to $40.00$. Yet using only those same four DeepMath prompts
already reaches $40.02$, with higher mean OOD and Avg$_4$ scores than
the 4/44 mixture. The improvement over GSM8K-only therefore does not
establish complementary benefits from combining the sources. The
four-prompt control changes support size and may also change prompt
repetition; all runs retain the same 25-update training configuration.
Together, these controls show why a source's standalone performance
is insufficient to determine either a filtering rule or its contribution
within a mixture. Construction and full filtering results are in
\crossref{Appendix~\ref{app:mixed-pool-selection}}.

\begin{takeaway}{5}
Strong transfer does not require a carefully curated prompt pool: uniform
random sampling remains competitive, while targeted selection or source
filtering did not consistently improve transfer in our experiments.
\end{takeaway}

\section{Discussion}
\label{sec:discussion}
\label{sec:limitations}
\label{sec:conclusion}

Prompt efficiency does not imply interchangeability: prompt utility
depends on the teacher--student pair and target capability. One possible
interpretation is that simple prompts can support broadly elicitable
behavioral changes, whereas more context-dependent reasoning behaviors
benefit from supports that expose relevant trajectories. The contrasting
prompt requirements of Qwen3-4B Math-RL and JustRL are consistent with this
interpretation. Continued OPD on effective supports can restore performance
after unfavorable transfer despite different parameter updates. Random
sampling remains competitive, while filtering a source that performs poorly
alone yields no consistent gain in our mixed-pool experiments. Selection
strategies should therefore demonstrate gains beyond random sampling while
accounting for screening cost.

\par
\endgroup
\label{page:main-text-end}

\bibliography{references}
\bibliographystyle{iclr2027_conference}

\appendix
\clearpage
\section{Related Work}
\label{app:related-work}
\label{sec:related-work}

\paragraph{On-policy distillation.}
Autoregressive distillation uses teacher supervision on student-generated
sequences to address the mismatch between training and inference
\citep{lin2020autoregressive,agarwal2024onpolicy}.
Subsequent work develops divergence objectives and optimization strategies
for language-model distillation
\citep{gu2024minillm,ko2025distillm2,ko2026reopold}, including selective
supervision on critical spans in privileged-context self-distillation
\citep{wang2026trace}.
These developments establish the algorithmic basis for our study,
which examines how prompt quantity and composition affect transfer
under a fixed OPD training procedure.

\paragraph{Understanding OPD transfer.}
\citet{li2026rethinking} identify compatible thinking patterns and
additional transferable teacher capabilities as factors associated
with successful OPD, and improve transfer through prompt-content
and template alignment.
Concurrent work demonstrates substantial transfer from a single query,
relating its effectiveness to training-state coverage and studying
the effects of query diversity and alignment dynamics
\citep{fu2026onetraining}.
Other concurrent work shows that OPD can propagate both strengths
and weaknesses across domains, with generalization depending on
model origin and interactions among teachers
\citep{li2026everycoin}.
Our independently conducted study also finds substantial few-prompt
and cross-domain transfer, and examines how prompt-count scaling
and source preferences vary across teacher--student pairs.
Controlled teacher comparisons and support-switch interventions
connect these preferences to unfavorable transfer and recovery.
Parameter and functional diagnostics further examine how recovery
can emerge through different parameter updates.

\paragraph{Data requirements and selection.}
Small-data post-training has also been studied in supervised
instruction tuning and reinforcement learning.
LIMA demonstrates strong instruction-following performance from
a small curated dataset \citep{zhou2023lima}, while one-shot RLVR
obtains substantial reasoning gains from a single training example
\citep{wang2025oneshotrlvr}.
Gradient-based methods such as LESS select instruction data
according to its relevance to target capabilities
\citep{xia2024less}.
In OPD, a fixed prompt support supplies evolving student trajectories
with teacher feedback, so distinct prompt count and supervision
volume are separate quantities.
We evaluate whether difficulty, length, semantic diversity, and
gradient-based selection improve transfer over random sampling
at fixed support sizes.
We also test whether a source's poor standalone performance
justifies removing it from a mixed candidate pool.

\section{Implementation and Evaluation Details}
\label{app:implementation}
\label{app:data-audits}

\subsection{Model and data sources}

\crossref{Table~\ref{tab:study-pairs}} lists the student and teacher models and OPD
data. The external \modelname{JustRL-DeepSeek-1.5B} teacher is an RL
descendant of \modelname{DeepSeek-R1-Distill-Qwen-1.5B}, with reported RL
data DAPO-Math-17k \citep{he2025justrl}. \modelname{DeepScaleR-1.5B-Preview} also continues RL
from \modelname{DeepSeek-R1-Distill-Qwen-1.5B}, using approximately 40K
problems from AIME, AMC, Omni-MATH, and Still \citep{deepscaler2025}.
Other released continuation teachers are
\modelname{Nemotron-Research-Reasoning-Qwen-1.5B} (v1) \citep{liu2025prorl},
\modelname{Qwen3-4B-Inst-Mix} \citep{wu2026consolidating}, and
\modelname{Light-R1-7B-DS} \citep{wen2025lightr1}.
The \modelname{Qwen3-30B-A3B-Instruct-2507} teacher is an official release.
The Math-RL and Code-RL teachers differ in RL data and training history;
the teacher-switch comparison does not isolate the effect of RL data domain.

\noindent\textbf{Model construction.}
Math RL and code RL start from \modelname{Qwen3-4B} using
DeepMath L6 \citep{he2026deepmath} and Eurus-2 Code \citep{cui2025process},
respectively. The 1.7B branch starts from
\modelname{Qwen3-1.7B-Base}, followed by our direct-code SFT checkpoint at
step 554 and then code RL on cleaned Open-R1 Code \citep{openr12025}. Checkpoint steps below
identify the selected models, not necessarily the total training duration.
The direct-code OPD student is initialized at this SFT-554 checkpoint,
and its teacher continues from the same checkpoint with code RL.
This initialization is distinct from the official \modelname{Qwen3-1.7B}
used in the cross-model comparisons. DeepMath L6 retains examples with
level $\geq 6$.

\noindent\textbf{Direct-code SFT data.}
Our curated mixture combines NVIDIA OpenCodeInstruct,
bigcode self-oss-instruct-sc2, Magicoder-OSS-Instruct-75K, and the
synthetic and seed SFT subsets of rStar-Coder.
We standardize the examples to direct Python code without
\texttt{<think>} traces and decontaminate the data against HumanEval+,
MBPP+, and LiveCodeBench v6.

\subsection{Training hyperparameters}
\label{app:experimental-details}

\crossref{Tables~\ref{tab:math-rl}--\ref{tab:opd-hyperparameters}} report the key
settings for teacher construction and OPD.

\paragraph{Sampled-token OPD implementation.}
We implement the distribution-matching objective in
\crossref{Equation~\ref{eq:opd-rkl}} through a sampled-token policy-gradient
surrogate \citep{lu2025onpolicydistillation}.
For a token $y_t$ sampled from the rollout policy $\pi_{\bar\theta}$
at prefix $s_t=(x,y_{<t})$, the K1 log-ratio and detached feedback are
\begin{equation}
    \begin{aligned}
        \widehat{k}_t
        &= \log\pi_{\bar\theta}(y_t\mid s_t)
           -\log\pi_T(y_t\mid s_t),\\
        \widehat{A}_t
        &= -\operatorname{sg}\!\left[
            \operatorname{clip}(\widehat{k}_t,-10,10)
        \right],
    \end{aligned}
    \label{eq:opd-feedback}
\end{equation}
where $\operatorname{sg}$ denotes stop-gradient.
The update averages over valid response tokens without accumulating
future-token feedback. Task rewards are used for monitoring and
reward-based selection baselines, but do not enter the OPD loss.

\noindent\textbf{OPD training duration.}
Nemotron runs use 75 optimization steps; Mix-RL runs use 25.
The \texttt{s100} label specifies the data-presentation schedule, not the
number of optimization steps actually executed.
The prompt-count comparisons in \crossref{Section~\ref{sec:data-requirements}}
report step-100 endpoints for Qwen3-4B Code-RL and JustRL, and step 15
for 30B Instruct to 4B with DeepMath. The Qwen3-4B Math-RL comparison
matches the total number of rollouts at 3,840.

\paragraph{Comparison protocols.}
Prompt-count sweeps vary $M$ within a pair and source under the specified
rollout and update protocol. Source comparisons fix the pair and $M=48$
while changing the source or mixture. The teacher-switch comparison
fixes the initial Qwen3-4B student and both supports while swapping
Math-RL and Code-RL teachers. Within-pool selection comparisons fix the
candidate pool, support size, and training configuration while changing
the selector. The mixed-pool comparison samples before and after
excluding GSM8K from DeepMath--DAPO--GSM8K candidates.
We record distinct prompts, fresh rollouts, optimization updates, and
token budgets separately; fixing $M$ does not match computational cost.

All lengths in the following tables are in tokens. PPO mini-batch sizes
are configuration values; per-device forward and backward passes use
dynamic token packing.

\begingroup
\newsavebox{\mathRLTableBox}
\newsavebox{\codeRLTableBox}
\newsavebox{\sftTableBox}
\newsavebox{\sftRLTableBox}
\newsavebox{\opdTableBox}
\setlength{\intextsep}{6pt}
\begin{table}[!ht]
    \small
    \setlength{\tabcolsep}{4pt}
    \renewcommand{\arraystretch}{1.02}
    \setlength{\abovecaptionskip}{0pt}
    \setlength{\belowcaptionskip}{4pt}
    \setlength{\aboverulesep}{1.5pt}
    \setlength{\belowrulesep}{2pt}
    \sbox{\mathRLTableBox}{\begin{minipage}[t]{0.48\linewidth}
        \vspace{0pt}
        \caption{\textbf{Math RL (GRPO).}}
        \label{tab:math-rl}
        \centering
\begin{tabular}{@{}L{\dimexpr.68\linewidth-4pt\relax}
                    L{\dimexpr.32\linewidth-4pt\relax}@{}}
    \toprule
    Hyperparameter & Value \\
    \midrule
    Train batch size & 128 \\
    PPO mini-batch size & 128 \\
    Rollout $n$ & 8 \\
    Max.\ prompt length & 2,048 \\
    Max.\ response length & 16,384 \\
    Temperature & 1.0 \\
    Top-$p$ & 1.0 \\
    Learning rate & $1\times10^{-6}$ \\
    KL coefficient & 0 \\
    Checkpoint step & 500 \\
    \bottomrule
\end{tabular}

    \end{minipage}}
    \sbox{\codeRLTableBox}{\begin{minipage}[t]{0.48\linewidth}
        \vspace{0pt}
        \caption{\textbf{Code RL (GRPO).}}
        \label{tab:code-rl}
        \centering
\begin{tabular}{@{}L{\dimexpr.68\linewidth-4pt\relax}
                    L{\dimexpr.32\linewidth-4pt\relax}@{}}
    \toprule
    Hyperparameter & Value \\
    \midrule
    Train batch size & 128 \\
    PPO mini-batch size & 128 \\
    Rollout $n$ & 8 \\
    Max.\ prompt length & 2,048 \\
    Max.\ response length & 8,192 \\
    Temperature & 1.0 \\
    Top-$p$ & 1.0 \\
    Learning rate & $1\times10^{-6}$ \\
    KL coefficient & 0 \\
    Checkpoint step & 300 \\
    \bottomrule
\end{tabular}

    \end{minipage}}
    \sbox{\sftTableBox}{\begin{minipage}[t]{0.48\linewidth}
        \vspace{0pt}
        \caption{\textbf{Direct-code SFT.}}
        \label{tab:direct-code-sft}
        \centering
\begin{tabular}{@{}L{\dimexpr.68\linewidth-4pt\relax}
                    L{\dimexpr.32\linewidth-4pt\relax}@{}}
    \toprule
    Hyperparameter & Value \\
    \midrule
    Train batch size & 128 \\
    Max.\ sequence length & 4,096 \\
    Warm-up ratio & 0.03 \\
    Learning rate & $3\times10^{-6}$ \\
    Training epochs & 1 \\
    Checkpoint step & 554 \\
    \bottomrule
\end{tabular}

    \end{minipage}}
    \sbox{\sftRLTableBox}{\begin{minipage}[t]{0.48\linewidth}
        \vspace{0pt}
        \caption{\textbf{Code RL after SFT (GRPO).}}
        \label{tab:sft-code-rl}
        \centering
\begin{tabular}{@{}L{\dimexpr.68\linewidth-4pt\relax}
                    L{\dimexpr.32\linewidth-4pt\relax}@{}}
    \toprule
    Hyperparameter & Value \\
    \midrule
    Train batch size & 128 \\
    PPO mini-batch size & 128 \\
    Rollout $n$ & 8 \\
    Max.\ prompt length & 4,096 \\
    Max.\ response length & 4,096 \\
    Temperature & 0.8 \\
    Top-$p$ & 0.95 \\
    Learning rate & $5\times10^{-7}$ \\
    KL coefficient & 0.001 \\
    Checkpoint step & 400 \\
    \bottomrule
\end{tabular}

    \end{minipage}}
    \sbox{\opdTableBox}{\begin{minipage}[t]{0.48\linewidth}
        \vspace{0pt}
        \caption{\textbf{OPD hyperparameters.}}
        \label{tab:opd-hyperparameters}
        \centering
\begin{tabular}{@{}L{\dimexpr.54\linewidth-4pt\relax}
                    L{\dimexpr.46\linewidth-4pt\relax}@{}}
    \toprule
    Hyperparameter & Value \\
    \midrule
    Train batch size & 256 \\
    PPO mini-batch size & 256 \\
    Rollout $n$ & 1 \\
    Max.\ prompt length & 2,048 \\
    Max.\ response length & 16,384 (Math)\newline 8,192 (Code) \\
    Temperature & 1.0 \\
    Top-$p$ & 1.0 \\
    Learning rate & $1\times10^{-5}$ \\
    Optimization steps & Pair-specific \\
    Distillation objective & K1 sampled\newline reverse-KL +\newline policy gradient \\
    Log-ratio clamp & $[-10,10]$ \\
    \bottomrule
\end{tabular}

    \end{minipage}}

    \begin{minipage}[t]{0.48\linewidth}
        \vspace{0pt}
        \usebox{\mathRLTableBox}
        \par\vspace{9pt}
        \usebox{\sftTableBox}
        \par\vspace{9pt}
        \usebox{\opdTableBox}
    \end{minipage}\hfill
    \begin{minipage}[t]{0.48\linewidth}
        \vspace{0pt}
        \usebox{\codeRLTableBox}
        \par\vspace{9pt}
        \usebox{\sftRLTableBox}
    \end{minipage}
\end{table}
\endgroup

\subsection{Prompt support construction}
\label{app:prompt-support-construction}

For the nested sweeps described below, each fixed candidate population
and selection seed defines one ordered sequence. The support $\mathcal S_M$ contains its
first $M$ entries, so $\mathcal S_{M_1}\subset\mathcal S_{M_2}$ for
$M_1<M_2$; the construction scripts assert this nesting.
The selection seed is 42 except for the Code48 redraws.
In the data-presentation pipeline, the training seed is also 42: it
controls presentation order without changing support membership.
Each support is frozen before its run. Nesting applies within a pool
and selection seed, not across different sources or redraw seeds.

\paragraph{Stratified nested prefixes.}
At each prefix extension, the construction chooses the stratum with the
largest relative quota deficit. Within a stratum, a SHA-256 ordering
keyed by the selection seed and example ID randomizes the examples.
This produces covariate-balanced nested prefixes, not a uniform
permutation of the entire pool.

For the main Qwen3-4B Math sweep, the population is the first 3,840
examples consumed by the corresponding full-data OPD run
(batch 256, 15 updates), drawn from DeepMath-103K's
\texttt{train\_filtered\_level6} split. Strata cross five baseline-step
bins, four prompt-length bins, and binary answer correctness.
Supports are nested at $M\in\{1,48,384,3072,3840\}$.
The earlier batch-1,024, 25-update sweep instead uses a 25,600-example
population and $M\in\{1,48,384,3072,25600\}$.
The main DeepMath48 comes from the 3,840-example sequence, not a
separate draw from the earlier 25,600-example population.
These construction populations are distinct from source-level full-pool
references, including the 57,046-prompt JustRL DeepMath reference.

The Qwen3-4B Code sweep uses all 22,618 Eurus Code GRPO examples,
stratified by \texttt{data\_source} and six prompt-length quantiles,
with $M\in\{1,48,3840,22618\}$.
Code48 redraws use the same population and algorithm with selection
seeds 43--47. They are not nested with the seed-42 support.
These redraws change support membership while keeping the training
presentation seed at 42.

\paragraph{Uniform nested prefixes.}
The following controls instead take prefixes of one NumPy PCG64
permutation with selection seed 42, without difficulty or gradient
screening. For the 30B teacher, cleaned DAPO-Math \citep{yu2025dapo} has
17,170 candidates and $M\in\{1,8,48,3840\}$;
GSM8K train \citep{cobbe2021gsm8k} has 7,473 candidates and
$M\in\{1,48\}$. The JustRL DAPO $M=4$, $8$, and $384$ supports also
use uniform nested prefixes of the cleaned 17,170-example pool.
Science48 is a frozen 48-example prefix of 126,397 physics, chemistry,
and biology examples from TextbookReasoning \citep{fan2025megascience}.

\paragraph{Shared supports across pairs.}
Qwen3-4B Math, 30B-to-4B DeepMath, Nemotron, and Mix-RL use the same
48 DeepMath IDs. JustRL's DeepMath scan directly reuses the frozen
Qwen3-4B Math support files rather than drawing new supports.
Light-R1's DeepMath $M=8$ support uses the same schedule as
JustRL's DeepMath random-8 support, drawn from
\texttt{train\_filtered\_level6} (level $\geq 6$).
The Code sweep, Nemotron, and Mix-RL likewise share the seed-42 Code48;
30B, Nemotron, and Mix-RL share the frozen Science48 indices.
Shared membership does not imply identical presentation schedules or
training budgets.
Nemotron and Mix-RL differ in model pair, training endpoint, and
DeepMath presentation schedule; source effects are compared within each pair.

\paragraph{Instruction-following and agent supports.}
IF48 and Agent48 come from the LLM-Fusion training parquet files in
the \texttt{Siye01/llm-fusion} collection \citep{wu2026consolidating}.
The IF file, \texttt{IF/train-00000-of-00001.parquet}, contains
16,575 examples; strata cross \texttt{rm\_type} and six prompt-length
quantiles. All examples have type \texttt{ifevalg}, including all
48 selected prompts. The Agent file,
\texttt{Agent/train-00000-of-00001.parquet}, contains 10,229 examples;
strata cross ground-truth tool-call counts $0/1/2/3/4+$ and six
prompt-length quantiles. Both use the stratified construction above
with seed 42 and freeze $M=48$. Agent48 contains 6, 28, 6, 2, and 6
examples in the five tool-count bins, respectively.

\paragraph{Mixtures.}
Mix3 (also called Mix48) takes the first 16 examples in the stored
support-file order of each frozen DeepMath48, Code48, and Science48;
it does not redraw them. Nemotron and Mix-RL share this 16+16+16 support.
Mix5 instead selects 10 Math, 10 Science, 10 Code, 9 IF, and 9 Agent
prompts from the corresponding frozen M48 supports.
An example's identity hashes its \texttt{prompt}, \texttt{data\_source},
and \texttt{reward\_model} fields with SHA-256.
Within each domain, examples are re-ranked by the SHA-256 hash of
\texttt{42:\{domain\}:\{identity\}}, and the required number is retained.
The combined presentation order uses the hash of
\texttt{42:support:\{domain\}:\{identity\}}.
Thus, each domain's Mix5 subset belongs to its frozen M48 parent but
is not selected as a file-order prefix. Its Math, Code, and Science
subsets overlap Mix3's corresponding 16-prompt subsets in 3, 2, and 1
prompts, respectively. Moving from Mix3 to Mix5 therefore changes both
domain allocation and selected prompt membership.

\subsection{Prompting and evaluation}
\label{app:evaluation-protocols}

\crossref{Appendix~\ref{app:prompt-templates}} gives the training and evaluation
templates for math reasoning, 4B code OPD, and the direct-code
SFT/GRPO/OPD family, as well as evaluation-only templates for GPQA-Diamond,
IFEval/IFBench, and BFCL-v3. In the direct-code templates, HumanEval+/MBPP+ and
LiveCodeBench identify evaluation task styles, not training data sources.

\noindent\textbf{Generation format.}
Prompt instructions are distinct from generation mode: a non-thinking
template followed by the instruction to think first can still produce reasoning text.
The direct-code
templates contain no \texttt{<think>} prefix and request no reasoning trace.

The math suite comprises AIME 2024/2025 \citep{aopsAimeArchive}
and HMMT February/November 2025
\citep{hmmt2025february,hmmt2025november}.

Math-OOD evaluation uses GPQA-Diamond \citep{rein2024gpqa},
HumanEval+, and LiveCodeBench v6.
Both code settings use the EvalPlus benchmarks HumanEval+ and MBPP+
\citep{liu2023evalplus}, and LiveCodeBench v6
(\texttt{release\_v6}; \citealp{jain2025livecodebench}).
Multi-domain evaluations additionally include IFEval \citep{zhou2023ifeval},
IFBench \citep{pyatkin2025ifbench}, and BFCL-v3 \citep{patil2025bfcl}.
\crossref{Table~\ref{tab:evaluation-protocols}} reports the evaluation settings,
which are distinct from the training budgets.

\begingroup
\setlength{\intextsep}{6pt}
\begin{table}[!htbp]
    \setlength{\abovecaptionskip}{0pt}
    \setlength{\belowcaptionskip}{3pt}
    \caption{\textbf{Evaluation configurations.} Lengths are in tokens;
    all settings report mean@$k$ and pass@$k$, with $k=16$ for Math and
    $k=8$ for OOD.}
    \label{tab:evaluation-protocols}
    \centering
    \small
    \setlength{\tabcolsep}{4pt}
    \renewcommand{\arraystretch}{1.10}
    \setlength{\aboverulesep}{1.5pt}
    \setlength{\belowrulesep}{2pt}
    \begin{tabular*}{\linewidth}{@{}l@{\extracolsep{\fill}}ccc@{}}
        \toprule
        \textbf{Parameter} & \textbf{Math / OOD}
        & \textbf{Code (4B)} & \textbf{Direct-code} \\
        \midrule
        Samples per question ($k$) & 16 / 8 & 8 & 16 \\
        Temperature & 1.0 & 1.0 & 0.6 \\
        Top-$p$ & 1.0 & 1.0 & 0.95 \\
        Top-$k$ / min-$p$ & $-1$ / 0.0 & $-1$ / 0.0 & $-1$ / 0.0 \\
        Max.\ generation length & 16,384 & 16,384 & 4,096 \\
        \bottomrule
    \end{tabular*}
\end{table}
\endgroup

\noindent\textbf{Aggregation.}
Mean@$k$ averages the fraction of correct responses across questions;
pass@$k$ is the fraction of questions with at least one correct response
among $k$ samples. Initial-student, teacher, and full-pool OPD references
are included where available.
For each reported metric, we take an unweighted average over the benchmarks
included in the corresponding aggregate:
\begin{equation*}
    \mathrm{Overall} = \frac{1}{B}\sum_{b=1}^{B}\mathrm{score}_b,
\end{equation*}
where $B$ is the number of included benchmarks and $\mathrm{score}_b$ is
that metric's score on benchmark $b$. Each benchmark has equal weight,
irrespective of its question count.
For the multi-domain comparison in \crossref{Table~\ref{tab:mixrl-data-preferences}},
$\mathrm{Avg}_6$ averages six capability groups, with IFEval and IFBench
first averaged within the instruction-following group; HumanEval+ and LCB
each retain a separate weight. This is distinct from the four-task Avg
in \crossref{Figure~\ref{fig:data-preferences}a,c}.
The Eurus Code row uses the original support draw.

\subsection{Prompt templates}
\label{app:prompt-templates}

\begingroup
\raggedbottom
\definecolor{promptheader}{RGB}{225,239,236}
\definecolor{promptborder}{RGB}{73,112,107}
\definecolor{prompttitle}{RGB}{35,66,62}
\definecolor{promptdivider}{RGB}{201,220,216}
\definecolor{prompttext}{RGB}{30,35,40}
\renewcommand{\ttdefault}{cmtt}
\newsavebox{\promptTemplateBox}
\newenvironment{prompttemplate}[2]{%
    \def\promptTemplateTitle{#1}%
    \begin{lrbox}{\promptTemplateBox}%
    \begin{minipage}{\dimexpr\linewidth-17.4pt\relax}%
    \refstepcounter{subsubsection}\label{#2}%
    \color{prompttext}%
    \fontsize{8.3}{10}\selectfont
    \setlength{\topsep}{1pt}%
}{%
    \end{minipage}%
    \end{lrbox}%
    \begin{tcolorbox}[
        title=\promptTemplateTitle,
        colback=white, colbacktitle=promptheader,
        colframe=promptborder, coltitle=prompttitle,
        fonttitle=\normalfont\bfseries\fontsize{9}{10.5}\selectfont,
        boxrule=0.7pt, titlerule=0.5pt,
        arc=3pt, outer arc=3pt, boxsep=0pt,
        left=8pt, right=8pt, top=5pt, bottom=5pt,
        toptitle=3pt, bottomtitle=3pt,
        before skip=6pt, after skip=6pt,
    ]
        \usebox{\promptTemplateBox}
    \end{tcolorbox}%
}
\newcommand{\promptvariant}[2]{%
    \noindent{\normalfont\fontsize{8.5}{10}\selectfont\bfseries #1}%
    \hfill{\normalfont\fontsize{8}{10}\selectfont #2}\par\vspace{2pt}%
}

\begin{prompttemplate}{Training and Evaluation Prompt Template for Math Reasoning}{app:prompt-math}
\begin{verbatim}
<|im_start|>user
{question}

Please reason step by step, and put your final answer within \boxed{}.<|im_end|>
<|im_start|>assistant
\end{verbatim}
\end{prompttemplate}

\begin{prompttemplate}{Training and Evaluation Prompt Template for Code Generation}{app:prompt-code}
\begin{verbatim}
<|im_start|>user
{question}

Write Python code to solve the problem. Present the code in
```python
Your code
```
at the end.
You need to think first then write the Python code.<|im_end|>
<|im_start|>assistant
\end{verbatim}
\end{prompttemplate}

\begin{prompttemplate}{Training and Evaluation Prompt Template for Direct Code Generation}{app:prompt-direct-code}
\promptvariant{Function-style}{HumanEval+ / MBPP+}
\begin{verbatim}
<|im_start|>user
{question}

Complete the requested Python function. Preserve the required function or class
interface exactly. Return only the final implementation inside one
```python
code block
```.<|im_end|>
<|im_start|>assistant
\end{verbatim}
\par\vspace{4pt}
\noindent{\color{promptdivider}\rule{\linewidth}{0.3pt}}\par\vspace{4pt}
\promptvariant{Competitive-programming-style}{LiveCodeBench}
\begin{verbatim}
<|im_start|>user
{question}

[Optional starter code:
```python
{starter_code}
```]
Solve the programming problem in Python. Preserve any required starter-code
interface. Return only the complete final program inside one
```python
code block
```.<|im_end|>
<|im_start|>assistant
\end{verbatim}
\end{prompttemplate}

\begin{prompttemplate}{Evaluation Prompt Template for GPQA-Diamond}{app:prompt-gpqa}
\begin{verbatim}
<|im_start|>user
Answer the following multiple choice question. The last line of your response
should be of the following format: 'Answer: $LETTER' (without quotes) where
LETTER is one of ABCD. Think step by step before answering.

{question}

A) {choice_A}
B) {choice_B}
C) {choice_C}
D) {choice_D}<|im_end|>
<|im_start|>assistant
\end{verbatim}
\par\vspace{4pt}
{\normalfont\footnotesize
The final line must be \verb|Answer: $A|, with \texttt{A} replaced by the
selected letter (A--D); the dollar sign is literal.\par}
\end{prompttemplate}

\begin{prompttemplate}{Evaluation Prompt Template for Instruction Following}{app:prompt-instruction-following}
\promptvariant{Query-only}{IFEval / IFBench}
\begin{verbatim}
<|im_start|>user
{instruction_following_query}<|im_end|>
<|im_start|>assistant
\end{verbatim}
\par\vspace{4pt}
{\normalfont\footnotesize
No additional system prompt or shared instruction is appended; the dataset
query already contains the instruction-following constraints.\par}
\end{prompttemplate}

\begin{prompttemplate}{Evaluation Prompt Template for BFCL-v3: Initial Turn}{app:prompt-bfcl-initial}
\begin{verbatim}
<|im_start|>system
You are a helpful assistant with access to a set of tools. Use the tools to
accomplish the user's request. Call one tool at a time and wait for its result
before deciding the next step. When the task is complete, reply to the user
without calling a tool.

# Tools

You may call one or more functions to assist with the user query.

You are provided with function signatures within <tools></tools> XML tags:
<tools>
{tool_schema_1}
{tool_schema_2}
...
{tool_schema_n}
</tools>

For each function call, return a json object with function name and arguments
within <tool_call></tool_call> XML tags:
<tool_call>
{"name": <function-name>, "arguments": <args-json-object>}
</tool_call><|im_end|>
<|im_start|>user
{user_request}<|im_end|>
<|im_start|>assistant
\end{verbatim}
\par\vspace{4pt}
{\normalfont\footnotesize
Each example exposes only the tool subset specified by
\verb|extra_info.tool_selection|, not the full tool inventory.\par}
\end{prompttemplate}

\begin{prompttemplate}{Evaluation Prompt Template for BFCL-v3: Subsequent Turns}{app:prompt-bfcl-interaction}
\promptvariant{Model tool call}{Assistant}
\begin{verbatim}
<|im_start|>assistant
<tool_call>
{"name": "{function_name}", "arguments": {arguments}}
</tool_call><|im_end|>
\end{verbatim}
\par\vspace{4pt}
\noindent{\color{promptdivider}\rule{\linewidth}{0.3pt}}\par\vspace{4pt}
\promptvariant{Environment execution result}{User role}
\begin{verbatim}
<|im_start|>user
<tool_response>
{tool_execution_result}
</tool_response><|im_end|>
<|im_start|>assistant
\end{verbatim}
\par\vspace{4pt}
\noindent{\color{promptdivider}\rule{\linewidth}{0.3pt}}\par\vspace{4pt}
\promptvariant{Task completion}{Final response without a tool call}
\begin{verbatim}
<|im_start|>assistant
{final_response}<|im_end|>
\end{verbatim}
\par\vspace{4pt}
{\normalfont\footnotesize
BFCL-v3 uses a multi-turn tool agent: one tool is called at a time, and its
result is awaited before the next action.\par}
\end{prompttemplate}

\endgroup

\begingroup
\raggedbottom
\section{Complete Results on Data Requirements}
\label{app:prompt-count-results}

\subsection{Prompt-count sweeps}

\crossref{Tables~\ref{tab:prompt-count-math4b}--\ref{tab:prompt-count-cross17b}}
give benchmark-level prompt-count results for the six teacher--student
pairs in \mbox{\crossref{Figure~\ref{fig:prompt-scaling}a--f}}.
All scores are percentages, and $M$ counts distinct training prompts.
Math is the unweighted mean of AIME24, AIME25, HMMT-Feb, and HMMT-Nov;
Code macro averages HumanEval+, MBPP+, and LCB-v6.
A dash indicates that $M$ does not apply to a reference model.
Reported values and their precision are preserved.

\begingroup
\makeatletter
\newenvironment{promptcounttable}[2]{%
    \par\addvspace{9pt}\noindent
    \begin{minipage}{\linewidth}
    \def\@captype{table}%
    \centering\small
    \setlength{\tabcolsep}{3pt}%
    \renewcommand{\arraystretch}{1.08}%
    \setlength{\abovecaptionskip}{0pt}%
    \setlength{\belowcaptionskip}{4pt}%
    \caption{#1}\label{#2}%
}{%
    \end{minipage}\par
}
\makeatother
\newcommand{\promptmathheader}{%
    Source & $M$ & AIME24 & AIME25 & \shortstack{HMMT\\Feb}
    & \shortstack{HMMT\\Nov} & Math & GPQA-D & HE+ & LCB \\
}
\newcommand{\promptrefheader}{%
    Source & Step & AIME24 & AIME25 & \shortstack{HMMT\\Feb}
    & \shortstack{HMMT\\Nov} & Math & GPQA-D & HE+ & LCB \\
}

\begin{promptcounttable}{\textbf{Math-RL continuation.} Math GRPO-500 to Qwen3-4B
    (\crossref{Figure~\ref{fig:prompt-scaling}a}), with 3,840 rollouts per OPD run.
    Math reports mean@16; other tasks report mean@8.}{tab:prompt-count-math4b}
    \begin{tabular*}{\linewidth}{@{\extracolsep{\fill}}lrrrrrrrrr@{}}
        \toprule
        \promptmathheader
        \midrule
        Initial & -- & 22.29 & 19.79 & 11.04 & 8.75 & 15.47 & 42.74 & 77.06 & 17.71 \\
        Teacher & -- & 62.71 & 56.46 & 34.38 & 39.79 & 48.33 & 52.65 & 83.92 & 20.07 \\
        \midrule
        DeepMath & 1 & 55.21 & 54.17 & 32.29 & 37.29 & 44.74 & 51.70 & 82.39 & 29.93 \\
        DeepMath & 48 & 60.21 & 53.12 & 33.33 & 38.96 & 46.41 & 53.22 & 83.23 & 27.21 \\
        DeepMath & 3,840 & 61.25 & 55.83 & 32.92 & 37.08 & 46.77 & 52.15 & 85.14 & 29.79 \\
        \bottomrule
    \end{tabular*}
\end{promptcounttable}

\begin{promptcounttable}{\textbf{Code-RL continuation.} Code GRPO-300 to Qwen3-4B
    (\crossref{Figure~\ref{fig:prompt-scaling}b}), at OPD step 100; all scores are mean@8.
    The principal $M=1$ support is ID 5038 ($B$ in
    \crossref{Section~\ref{sec:transfer-failure-recovery}}); rows below the separator
    give additional single-prompt experiments.}{tab:prompt-count-code4b}
    \begin{tabular*}{\linewidth}{@{\extracolsep{\fill}}lrrrrr@{}}
        \toprule
        Source / support & $M$ & HE+ & MBPP+ & LCB & Code macro \\
        \midrule
        Initial & -- & 76.60 & 65.15 & 18.14 & 53.30 \\
        Teacher & -- & 86.36 & 71.03 & 28.00 & 61.80 \\
        \midrule
        Eurus Code (ID 5038) & 1 & 79.04 & 68.39 & 26.43 & 57.95 \\
        Eurus Code & 48 & 86.66 & 70.14 & 28.57 & 61.79 \\
        Eurus Code & 3,840 & 86.05 & 70.90 & 28.14 & 61.70 \\
        Eurus Code (full) & 22,618 & 86.74 & 71.06 & 27.86 & 61.89 \\
        \midrule
        $A'$ / ID 15240 (replication) & 1 & 69.74 & 61.71 & 18.64 & 50.03 \\
        Selected M1: ID 187 & 1 & 83.23 & 70.07 & 27.93 & 60.41 \\
        Selected M1: ID 228 & 1 & 83.23 & 69.21 & 29.21 & 60.55 \\
        Selected M1: ID 19303 & 1 & 86.20 & 68.92 & 29.00 & 61.37 \\
        \bottomrule
    \end{tabular*}
\end{promptcounttable}

\begin{promptcounttable}{\textbf{Direct-code RL continuation.} Code GRPO-400 to
    Qwen3-1.7B-Base after direct-code SFT at step 554
    (\crossref{Figure~\ref{fig:prompt-scaling}c}).
    OPD endpoints are at step 50; all scores are mean@16.}{tab:prompt-count-code17b}
    \begin{tabular*}{\linewidth}{@{\extracolsep{\fill}}lrrrrr@{}}
        \toprule
        Source / reference & $M$ & HE+ & MBPP+ & LCB & Code macro \\
        \midrule
        Initial: direct-code SFT-554 & -- & 44.32 & 52.88 & 11.25 & 36.15 \\
        Teacher: Code GRPO-400 & -- & 58.12 & 58.47 & 15.96 & 44.18 \\
        \midrule
        Open-R1 Code & 1 & 58.82 & 58.74 & 15.96 & 44.51 \\
        Open-R1 Code & 256 & 59.98 & 59.51 & 15.57 & 45.02 \\
        Open-R1 Code & 3,840 & 59.68 & 59.62 & 16.61 & 45.30 \\
        Open-R1 Code (full) & 6,519 & 58.73 & 58.80 & 15.50 & 44.34 \\
        \bottomrule
    \end{tabular*}
\end{promptcounttable}

\begin{promptcounttable}{\textbf{JustRL continuation.}
    JustRL-DeepSeek-1.5B to DeepSeek-R1-Distill-Qwen-1.5B
    (\crossref{Figure~\ref{fig:prompt-scaling}d}), at OPD step 100.
    Math reports mean@16; other tasks report mean@8.
    Full denotes the available source pool; DeepMath $M=8$ is the
    random support in the prompt-count sweep.}{tab:prompt-count-justrl}
    \begin{tabular*}{\linewidth}{@{\extracolsep{\fill}}lrrrrrrrrr@{}}
        \toprule
        \promptmathheader
        \midrule
        Initial & -- & 27.50 & 25.63 & 12.50 & 11.25 & 19.22 & 35.92 & 54.34 & 13.50 \\
        Teacher & -- & 53.96 & 39.38 & 21.67 & 23.75 & 34.69 & 40.47 & 64.56 & 19.36 \\
        \midrule
        DAPO & 1 & 42.71 & 30.42 & 16.88 & 17.08 & 26.77 & 38.13 & 60.98 & 16.29 \\
        DAPO & 2 & 49.58 & 36.25 & 20.83 & 20.83 & 31.87 & 38.64 & 62.27 & 16.71 \\
        DAPO & 4 & 47.71 & 36.04 & 22.08 & 24.38 & 32.55 & 41.10 & 63.87 & 17.50 \\
        DAPO & 8 & 51.04 & 35.00 & 21.25 & 22.71 & 32.50 & 40.91 & 64.10 & 18.64 \\
        DAPO & 384 & 49.38 & 39.17 & 21.04 & 22.29 & 32.97 & 39.14 & 63.57 & 19.21 \\
        DAPO (full) & 17,170 & 50.21 & 35.83 & 21.46 & 21.88 & 32.34 & 40.28 & 64.25 & 19.21 \\
        \midrule
        DeepMath & 1 & 44.38 & 30.83 & 20.62 & 16.67 & 28.12 & 38.45 & 61.13 & 16.57 \\
        DeepMath & 8 & 45.42 & 34.38 & 17.92 & 18.96 & 29.17 & 38.83 & 63.34 & 17.43 \\
        DeepMath & 48 & 46.67 & 33.96 & 20.42 & 21.04 & 30.52 & 38.51 & 64.02 & 17.21 \\
        DeepMath & 384 & 49.17 & 35.21 & 22.29 & 20.00 & 31.67 & 38.83 & 63.64 & 18.79 \\
        DeepMath & 3,840 & 49.79 & 37.92 & 20.83 & 21.67 & 32.55 & 40.97 & 63.80 & 18.64 \\
        DeepMath (full) & 57,046 & 50.21 & 38.54 & 20.83 & 19.79 & 32.34 & 39.14 & 63.03 & 18.57 \\
        \bottomrule
    \end{tabular*}
\end{promptcounttable}

\begin{promptcounttable}{\textbf{Cross-model distillation.} Qwen3-30B-A3B-Instruct-2507
    to Qwen3-4B (\crossref{Figure~\ref{fig:prompt-scaling}e}), at OPD step 15.
    Math reports mean@16; other tasks report mean@8.
    Full-pool endpoints at other training durations appear in
    \crossref{Table~\ref{tab:prompt-count-full-references}}.}{tab:prompt-count-cross4b}
    \begin{tabular*}{\linewidth}{@{\extracolsep{\fill}}lrrrrrrrrr@{}}
        \toprule
        \promptmathheader
        \midrule
        Initial & -- & 22.29 & 19.79 & 11.04 & 8.75 & 15.47 & 42.74 & 77.13 & 17.71 \\
        Teacher & -- & 75.21 & 61.67 & 42.50 & 56.25 & 58.91 & 54.86 & 82.39 & 33.14 \\
        \midrule
        DeepMath & 1 & 46.0 & 46.7 & 24.6 & 31.9 & 37.29 & 48.3 & 82.4 & 24.3 \\
        DeepMath & 48 & 56.46 & 45.63 & 29.38 & 39.79 & 42.81 & 52.15 & 82.32 & 26.79 \\
        DeepMath & 3,840 & 53.8 & 48.3 & 30.2 & 38.1 & 42.60 & 52.0 & 83.7 & 27.0 \\
        \midrule
        DAPO & 1 & 49.4 & 40.4 & 26.2 & 31.5 & 36.88 & 49.7 & 83.5 & 26.3 \\
        DAPO & 8 & 56.0 & 45.0 & 28.3 & 39.0 & 42.08 & 50.9 & 81.4 & 26.9 \\
        DAPO & 48 & 54.4 & 45.6 & 27.7 & 34.0 & 40.42 & 51.5 & 82.2 & 27.9 \\
        DAPO & 3,840 & 51.5 & 43.3 & 29.0 & 35.2 & 39.74 & 51.5 & 81.1 & 27.3 \\
        \bottomrule
    \end{tabular*}
\end{promptcounttable}

\begin{promptcounttable}{\textbf{Cross-model distillation.} Qwen3-30B-A3B-Instruct-2507
    to the official Qwen3-1.7B checkpoint
    (\crossref{Figure~\ref{fig:prompt-scaling}f}), not the SFT-554 initialization in
    \crossref{Table~\ref{tab:prompt-count-code17b}}.
    Math reports mean@16; other tasks report mean@8.}{tab:prompt-count-cross17b}
    \begin{tabular*}{\linewidth}{@{\extracolsep{\fill}}lrrrrrrrrr@{}}
        \toprule
        \promptmathheader
        \midrule
        Initial & -- & 13.33 & 10.00 & 5.00 & 5.63 & 8.49 & 29.10 & 59.83 & 12.36 \\
        Teacher & -- & 75.21 & 61.67 & 42.50 & 56.25 & 58.91 & 54.86 & 82.39 & 33.14 \\
        \midrule
        DeepMath & 1 & 27.3 & 24.0 & 16.0 & 14.2 & 20.4 & 30.1 & 65.5 & 16.1 \\
        DeepMath & 8 & 33.1 & 26.7 & 15.0 & 15.2 & 22.5 & 32.5 & 65.8 & 19.4 \\
        DeepMath & 48 & 35.0 & 26.9 & 14.2 & 17.1 & 23.3 & 36.2 & 66.7 & 18.1 \\
        DeepMath & 3,840 & 36.0 & 30.8 & 14.8 & 16.7 & 24.6 & 32.7 & 68.9 & 18.7 \\
        \midrule
        DAPO & 1 & 33.5 & 29.2 & 15.0 & 16.2 & 23.5 & 35.6 & 71.2 & 18.2 \\
        DAPO & 8 & 32.1 & 26.0 & 14.6 & 10.2 & 20.7 & 33.9 & 66.9 & 17.9 \\
        DAPO & 48 & 30.2 & 27.3 & 11.7 & 15.4 & 21.1 & 33.7 & 64.1 & 16.9 \\
        DAPO & 3,840 & 33.5 & 29.8 & 15.4 & 14.8 & 23.4 & 34.5 & 66.5 & 18.9 \\
        \bottomrule
    \end{tabular*}
\end{promptcounttable}

\subsection{Additional full-pool references}

The following endpoints use different training durations or budgets from
the principal sweeps above. Full denotes the available sampling pool,
not a guarantee that every prompt was consumed.

\begin{promptcounttable}{\textbf{Additional full-pool OPD endpoints.}
    Math reports mean@16; other tasks report mean@8.
    Math-RL runs use batch size 1,024, with 25,600 and 51,200 rollouts
    at OPD steps 25 and 50, respectively.}{tab:prompt-count-full-references}
    \begin{tabular*}{\linewidth}{@{\extracolsep{\fill}}lrrrrrrrrr@{}}
        \toprule
        \promptrefheader
        \midrule
        \multicolumn{10}{@{}l}{Math GRPO-500 $\rightarrow$ Qwen3-4B} \\
        \addlinespace[2pt]
        DeepMath & 25 & 62.92 & 54.58 & 32.29 & 38.75 & 47.14 & 52.78 & 84.91 & 29.86 \\
        DeepMath & 50 & 64.79 & 56.04 & 34.38 & 41.46 & 49.17 & 54.29 & 84.45 & 28.07 \\
        \midrule
        \multicolumn{10}{@{}l}{Qwen3-30B-A3B-Instruct-2507 $\rightarrow$ Qwen3-4B} \\
        \addlinespace[2pt]
        DeepMath & 25 & 55.0 & 48.8 & 27.3 & 37.9 & 42.2 & 48.4 & 82.8 & 25.8 \\
        DeepMath & 50 & 56.9 & 49.0 & 31.7 & 38.8 & 44.1 & 48.4 & 83.5 & 26.6 \\
        DAPO & 25 & 53.1 & 46.7 & 27.9 & 35.2 & 40.7 & 49.2 & 80.9 & 25.9 \\
        \midrule
        \multicolumn{10}{@{}l}{Qwen3-30B-A3B-Instruct-2507 $\rightarrow$ Qwen3-1.7B} \\
        \addlinespace[2pt]
        DeepMath & 25 & 34.0 & 27.3 & 17.3 & 16.7 & 23.8 & 34.3 & 66.8 & 17.9 \\
        DeepMath & 50 & 34.6 & 29.0 & 16.9 & 16.7 & 24.3 & 35.2 & 67.8 & 18.6 \\
        DAPO & 25 & 34.2 & 26.9 & 16.5 & 14.2 & 22.9 & 35.5 & 66.8 & 19.1 \\
        \bottomrule
    \end{tabular*}
\end{promptcounttable}
\endgroup

\endgroup
\subsection{Support size and the learning process}
\label{app:support-size-training-curves}

\begin{figure}[!ht]
    \centering
    \includegraphics[width=\linewidth]{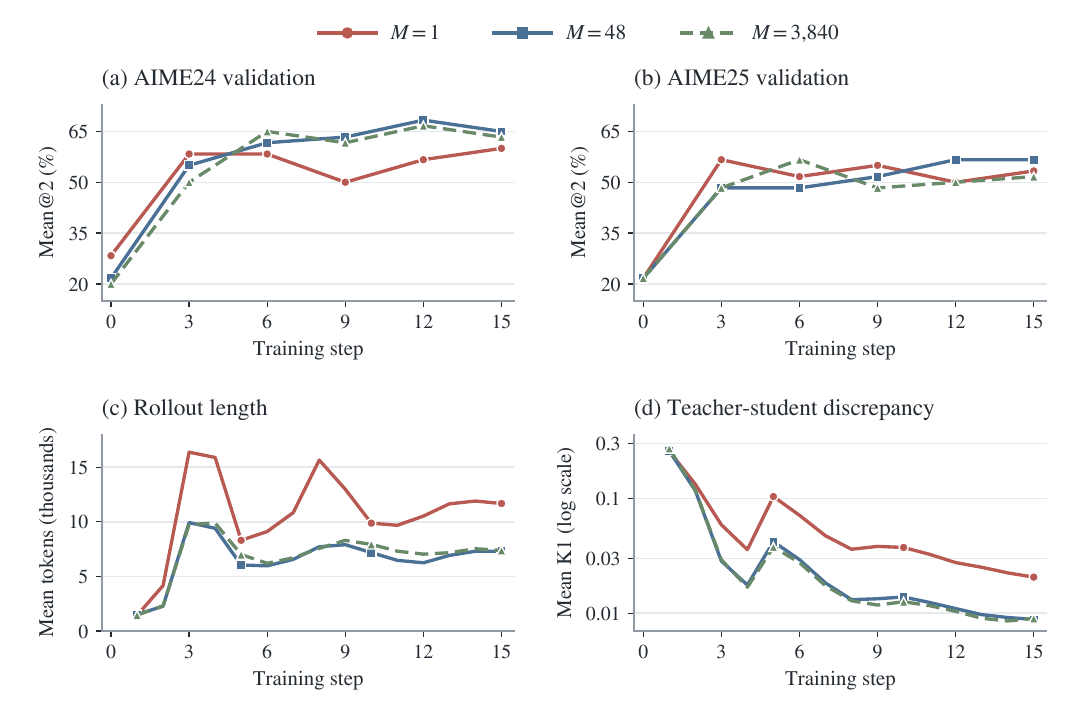}
    \caption{\textbf{Support size and learning dynamics.}
    All runs use the Qwen3-4B Math-RL continuation pair.
    \textbf{(a--b)} Training-time AIME24 and AIME25 validation
    accuracy (mean@2). \textbf{(c)} Mean rollout response length, in thousands of
    tokens. \textbf{(d)} Teacher--student discrepancy, reported as logged mean K1
    on a logarithmic scale.
    These validation traces are distinct from the final mean@16/mean@8
    evaluations in the complete result tables.}
    \label{fig:support-size-training-curves}
\end{figure}

\crossref{Figure~\ref{fig:support-size-training-curves}} complements the endpoint
tables with training trajectories for the Qwen3-4B Math-RL continuation
pair. The $M=48$ and $M=3{,}840$ runs closely track one another in rollout
length and logged K1. The $M=1$ run already improves validation
accuracy at the first post-training check (step 3), but generates longer
responses with larger step-to-step length fluctuations.
At step 15, mean rollout lengths are approximately 7,293, 7,377, and
11,676 tokens for $M=48$, $3{,}840$, and $1$, respectively.
Similar endpoint performance therefore need not imply the same generation
behavior during training.

\par\noindent\begin{minipage}{\linewidth}
\subsection{Single-prompt code supports}
\label{app:figure1-code-prompts}

The following cards summarize the single-prompt supports compared in
\crossref{Figure~\ref{fig:overview}b} for the Qwen3-4B Code-RL setting.

\begingroup
\definecolor{codesupportheader}{RGB}{225,239,236}
\definecolor{codesupportborder}{RGB}{73,112,107}
\definecolor{codesupporttitle}{RGB}{35,66,62}
\newenvironment{codesupport}[3]{%
    \begin{tcolorbox}[
        title={\textbf{#1\enspace{}#2}\hfill
            {\normalfont\small Index #3}},
        colback=white, colbacktitle=codesupportheader,
        colframe=codesupportborder, coltitle=codesupporttitle,
        coltext=black,
        fonttitle=\normalfont\small,
        fontupper=\normalfont\small,
        boxrule=0.5pt, titlerule=0.4pt,
        arc=2pt, outer arc=2pt, boxsep=0pt,
        left=8pt, right=8pt, top=6pt, bottom=6pt,
        toptitle=4pt, bottomtitle=4pt,
        before skip=6pt, after skip=2pt,
    ]
}{%
    \end{tcolorbox}%
}

\begin{codesupport}{P1}{Conditional arithmetic}{15240}
Given positive integers $A$ and $B$, output $A+B$ if $A\mid B$;
otherwise, output $B-A$.
\end{codesupport}

\begin{codesupport}{P2}{Largest perfect power}{187}
Given an integer $X$ with $1\le X\le1000$, find the largest perfect
power $b^p$ not exceeding $X$, where $b$ is a positive integer and $p$
is an integer satisfying $p\ge2$.
\end{codesupport}

\begin{codesupport}{P3}{Fixed-start Hamiltonian paths}{228}
Given an undirected graph with $N\le8$ vertices, count the paths that
start at vertex 1 and visit every vertex exactly once; that is, count
Hamiltonian paths with the starting vertex fixed to 1.
\end{codesupport}
\endgroup
\end{minipage}\par

\section{Additional Data-Preference Results}
\label{app:data-preferences}
\label{app:additional-results}

\crossref{Figure~\ref{fig:data-preferences}} and \crossref{Table~\ref{tab:mixrl-data-preferences}}
report the fixed-support source comparisons, teacher swap, and multi-domain
mixtures in \crossref{Section~\ref{sec:data-preferences}}. This appendix adds
teacher--source learning curves and seed robustness, the benchmark-level
supervision-token control, and the matched RL-instance overlap comparison.
Support membership and mixture construction are
specified in \crossref{Appendix~\ref{app:prompt-support-construction}}.

\subsection{Teacher and source effects}
\label{app:teacher-source-training-curves}

\begin{figure}[!ht]
    \centering
    \includegraphics[width=\linewidth]{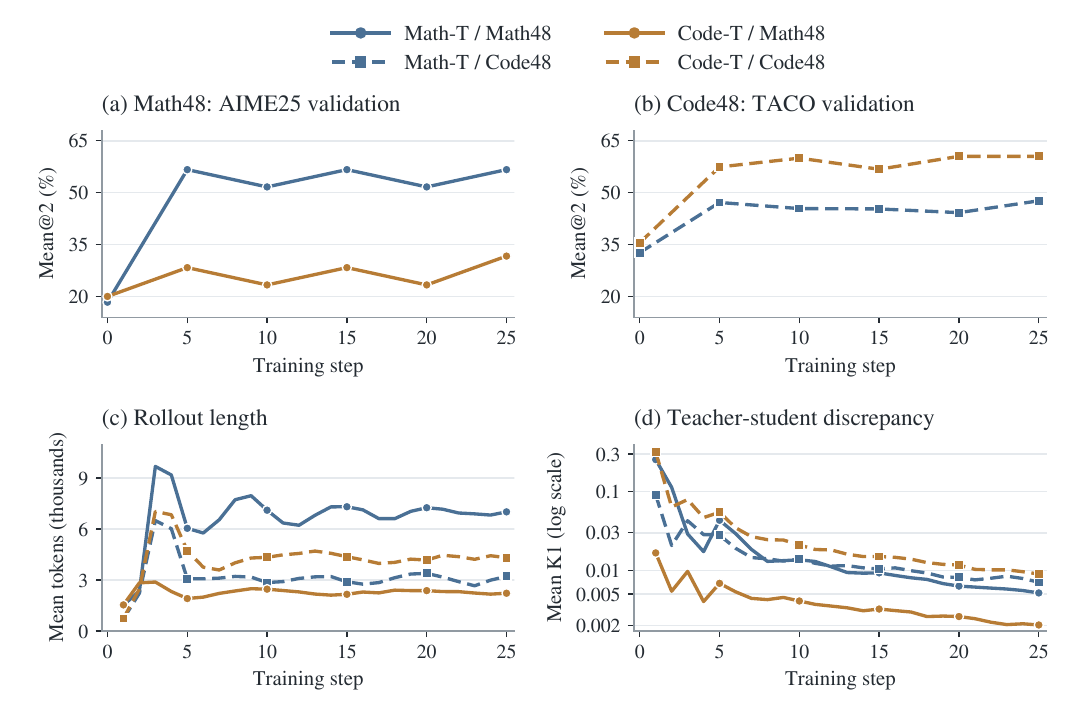}
    \caption{\textbf{Teacher and prompt-source effects during Qwen3-4B OPD.}
    Math-T and Code-T denote Math GRPO-500 and Code GRPO-300;
    Math48 and Code48 denote the fixed DeepMath and Eurus Code supports.
    \textbf{(a)} AIME25 validation with Math48. \textbf{(b)} TACO validation with Code48.
    Both report training-log mean@2. \textbf{(c)} Mean rollout response length.
    \textbf{(d)} Teacher--student discrepancy, reported as logged mean K1 (log scale).
    Panels (a--b) use different validation tasks; the common-benchmark
    source-preference reversal is shown separately in
    \crossref{Figure~\ref{fig:data-preferences}b}.}
    \label{fig:teacher-source-training-curves}
\end{figure}

\crossref{Figure~\ref{fig:teacher-source-training-curves}} shows when teacher-dependent
differences appear. With Math48 fixed, the Math-RL teacher yields higher
AIME25 validation accuracy than the Code-RL teacher from the first
post-training validation at step 5. With Code48 fixed, the Code-RL teacher
yields higher TACO validation accuracy at the same check.
The separation persists at subsequent validation checkpoints, even though
logged mean K1 generally decreases in all four runs, with transient
increases. A declining K1 trace thus coexists with distinct validation
outcomes. The two validation tasks are kept separate rather than pooled:
these curves describe the onset and persistence of teacher-dependent
differences, while \crossref{Figure~\ref{fig:data-preferences}b} compares source
preferences on the same final-evaluation benchmarks.

\paragraph{Seed robustness.}
We vary support-selection seeds over $\{42,43,44\}$ with training seed 42,
and training seeds over $\{42,43,44\}$ with support-selection seed 42,
keeping the student initialization fixed. Within each source and seed
configuration, the two teacher conditions use identical supports
($M=48$) and presentation schedules.

\par\addvspace{6pt}\noindent
\begin{minipage}{\linewidth}
\makeatletter\def\@captype{table}\makeatother
\centering
\small
\setlength{\tabcolsep}{5pt}
\renewcommand{\arraystretch}{1.12}
\setlength{\abovecaptionskip}{0pt}
\setlength{\belowcaptionskip}{5pt}
\caption{\textbf{Seed robustness of teacher-dependent source preferences.}
Avg$_4$ (\%) across support-selection and training seeds.}
\label{tab:teacher-source-seed-robustness}
\begin{tabular*}{\linewidth}{@{\extracolsep{\fill}}lcccc@{}}
\toprule
 & \multicolumn{2}{c}{Math-RL teacher}
 & \multicolumn{2}{c}{Code-RL teacher} \\
\cmidrule(lr){2-3}\cmidrule(l){4-5}
Support / training seed & DeepMath48 & Code48 & DeepMath48 & Code48 \\
\midrule
42 / 42 & 54.30 & 46.84 & 44.94 & 50.75 \\
43 / 42 & 54.20 & 46.62 & 44.79 & 51.27 \\
44 / 42 & 54.23 & 46.17 & 45.22 & 50.51 \\
42 / 43 & 53.87 & 46.37 & 44.69 & 50.31 \\
42 / 44 & 54.26 & 46.66 & 44.67 & 50.42 \\
\bottomrule
\end{tabular*}
\end{minipage}\par\addvspace{4pt}

Across all five configurations, Math-RL favors DeepMath48 and Code-RL
favors Code48 on each of the four benchmarks.

\subsection{Supervision-token control}
\label{app:token-match-breakdown}
\crossref{Table~\ref{tab:source-token-match-breakdown}} gives the benchmark-level
results underlying \crossref{Table~\ref{tab:source-token-match}}. All supports contain
48 prompts. Token counts are cumulative supervised tokens, in millions;
the matched GSM8K run is close to, but not exactly at, the DeepMath budget.
The main-text Avg is the unweighted mean of Math, GPQA, HumanEval+, and LCB\@.

\begin{table}[!htbp]
    \centering
    \small
    \setlength{\tabcolsep}{6pt}
    \renewcommand{\arraystretch}{1.08}
    \setlength{\belowcaptionskip}{4pt}
    \caption{\textbf{Benchmark-level supervision-token control.}
    $M=48$; tokens are cumulative supervised tokens in millions,
    and GSM8K token matching to DeepMath is approximate.}
    \label{tab:source-token-match-breakdown}
    \begin{tabular}{lrrrrr}
        \toprule
        Data & Tokens (M) & Math & GPQA & HE+ & LCB \\
        \midrule
        GSM8K & 1.19 & 31.15 & 46.84 & 81.25 & 16.79 \\
        \rowcolor[HTML]{F1F2F4}
        \hspace{1em}+ token match & 16.14 & 31.56 & 48.04 & 82.32 & 17.43 \\
        DeepMath & 16.63 & 42.81 & 52.15 & 82.32 & 26.79 \\
        \bottomrule
    \end{tabular}
\end{table}

\subsection{RL-instance overlap control}
\label{app:rl-instance-overlap}
DAPO is also the JustRL teacher's RL pool, so its comparison with DeepMath
cannot isolate exact training-instance reuse. The matched direct-code 1.7B
control fixes source mixture, prompt template, and prompt length:
RL-seen and RL-unseen sets yield close transfer point estimates
($45.26$ versus $45.28$). This provides evidence that exact RL-instance
reuse is not required for the observed transfer in this controlled comparison.
The $30$B teachers' RL instances
are unknown, so their outcomes cannot identify an overlap effect.

\section{Functional Changes, Failure, and Recovery}
\label{app:delta-exposure}

This appendix specifies the measurements, probe scopes, and supporting
evidence used in \crossref{Section~\ref{sec:transfer-failure-recovery}}: finite endpoint
prediction changes, conditional KL closure, generation behavior, and recovery
interventions.

\subsection{Finite endpoint prediction changes}
\label{app:functional-measurement}

Within each comparison, each model is loaded and fully forward-evaluated on the same frozen probe
prefixes. The measured effects are finite changes in endpoint conditional
predictions relative to the initial student. Freezing prefixes fixes the
input states; it does not linearize the model with respect to its parameters.
We do not compute $J_{\theta_0}\Delta\theta$ or use gradients at
initialization to predict an endpoint. Fisher geometry is used to compare
the measured prediction changes, not to replace the endpoint forward passes.

For the 30B-to-4B source comparison, the frozen probe comprises 128
initial-Qwen3-4B trajectories: 32 each from Math, GPQA, HumanEval+, and
LCB, totaling 73,823 prediction positions. Action support is fixed to
the initial student's top 128 tokens at each position. The initial-student
anchor and all seven evaluated models use FP32 forward passes with SDPA\@.
Student and teacher tokenizer mappings agree, and retained support mass
is at least $0.996$ across the evaluated models and domains. The Code
trajectory study likewise fixes 128 trajectories from its own initial
student, the initial student's top-128 action support, and an FP32 anchor;
each checkpoint is actually forward-evaluated. These protocol details do
not identify an all-domain aggregate with a Math-only comparison.

\paragraph{Finite endpoint vectors.}
At frozen prediction position $j$, let $S_j$ be the initial student's
top-128 token set, and let $p_{a,j}(i)$ be model $a$'s full-vocabulary
probability of token $i$. Here $a\in\{0,T,D\}$ denotes the initial student,
teacher, or distilled endpoint, respectively. Each model's conditional
distribution on the same set is
\begin{equation}
    m_{a,j}=\sum_{k\in S_j}p_{a,j}(k),
    \qquad
    q_{a,j}(i)=\frac{p_{a,j}(i)}{m_{a,j}},\quad i\in S_j.
    \label{eq:frozen-support-distribution}
\end{equation}
We center the actual endpoint-minus-initial log-probability changes
under $q_{0,j}$ and embed them using the initial probabilities:
\begin{align}
    d_{a,j}(i)&=\log p_{a,j}(i)-\log p_{0,j}(i),
    &\bar d_{a,j}&=\sum_{i\in S_j}q_{0,j}(i)d_{a,j}(i),
    \label{eq:endpoint-logprob-change}\\
    v_{a,j}(i)&=\sqrt{p_{0,j}(i)}\,
        [d_{a,j}(i)-\bar d_{a,j}],
    &&a\in\{T,D\}.
    \label{eq:endpoint-functional-vector}
\end{align}
Thus centering uses the renormalized $q_{0,j}$, while the Fisher inner-product
weight is the unnormalized $p_{0,j}(i)$, not teacher or endpoint probabilities.
The square-root embedding absorbs this weight: inner products of $v$ below
are ordinary Euclidean inner products, with no second Fisher weighting.
These are finite endpoint log-probability changes, not probability differences
or a parameter-local approximation.

\paragraph{Geometry and aggregation.}
For the evaluated position set $\mathcal J$, concatenate
$v_a=\bigoplus_{j\in\mathcal J}v_{a,j}$. Functional cosine and projection are
\begin{equation}
    \cos_{F}
    =\frac{\sum_{j\in\mathcal J}\langle v_{D,j},v_{T,j}\rangle}
           {\sqrt{\sum_{j\in\mathcal J}\lVert v_{D,j}\rVert^2}
            \sqrt{\sum_{j\in\mathcal J}\lVert v_{T,j}\rVert^2}},
    \qquad
    \alpha=\frac{\langle v_D,v_T\rangle}{\lVert v_T\rVert^2}.
    \label{eq:functional-decomposition}
\end{equation}
Both norms must be nonzero for cosine, and projection requires a nonzero
teacher direction. The orthogonal component $e_\perp=v_D-\alpha v_T$
is defined in the same embedded space; it is not a token-level training
log-ratio.

For all-domain cosine, $\mathcal J$ contains all evaluated positions across
domains. Each position enters once; longer trajectories and domains with
more positions contribute more entries. There is no trajectory-equal or
domain-equal averaging, nor an average of separately computed domain cosines.
Confidence intervals use whole-trajectory or whole-question resampling, as
specified for each comparison; this does not alter the point-estimate
definition above. Task-specific comparisons restrict $\mathcal J$ to
their stated probe scope.

\paragraph{Conditional forward KL and closure.}
The diagnostic uses teacher-first forward KL, with each model separately
renormalized on $S_j$:
\begin{equation}
    K_{T\Vert a,j}
    =D_{\mathrm{KL}}(q_{T,j}\Vert q_{a,j})
    =\sum_{i\in S_j}q_{T,j}(i)
       \log\frac{q_{T,j}(i)}{q_{a,j}(i)},\qquad a\in\{0,D\}.
    \label{eq:conditional-forward-kl}
\end{equation}
This is conditional KL on the initial student's frozen top-128 support,
not full-vocabulary KL\@. Probability outside $S_j$ is neither included in
this KL nor merged into an ``other'' token; retained support mass $m_{a,j}$
is reported separately. In particular, the conditional KL does not directly
penalize changes in the omitted mass.

Let $D_{T\Vert0}$ and $D_{T\Vert D}$ denote the baseline and remaining
conditional KL, respectively, using the same prefix set and aggregation
protocol. Diagnostic closure is
\begin{equation}
    \operatorname{KLClosure}
    =1-\frac{D_{T\Vert D}}{D_{T\Vert0}},\qquad D_{T\Vert0}>0.
    \label{eq:conditional-kl-closure}
\end{equation}
Reported percentages multiply this quantity by 100. A negative value denotes
a larger remaining conditional KL gap, not a negative cosine or a percentage
loss in accuracy. This teacher-first evaluation diagnostic is separate from
the sampled reverse-KL OPD training objective. Neither closure nor directional
cosine establishes equality of full prediction distributions or downstream
behavior.

\subsection{Probe scope}
\label{app:measurement-versions}

The 43 checkpoints across seven teacher families in
\crossref{Figure~\ref{fig:functional-changes}(a)} include intermediate states,
not 43 independent training runs. Their functional cosines use
all-domain frozen-prefix probes, whereas the source comparisons in
\crossref{Figure~\ref{fig:functional-changes}(b)} and the task-specific diagnostics
in \crossref{Figure~\ref{fig:extended-source-diagnostics}} restrict the probe scope
to the stated task. These scopes should not be treated as interchangeable.

\subsection{Source-resolved functional diagnostics}
\label{app:source-functional-diagnostics}

\begin{figure}[!htbp]
    \centering
    \includegraphics[width=\linewidth,trim=0 0 0 168bp,clip]{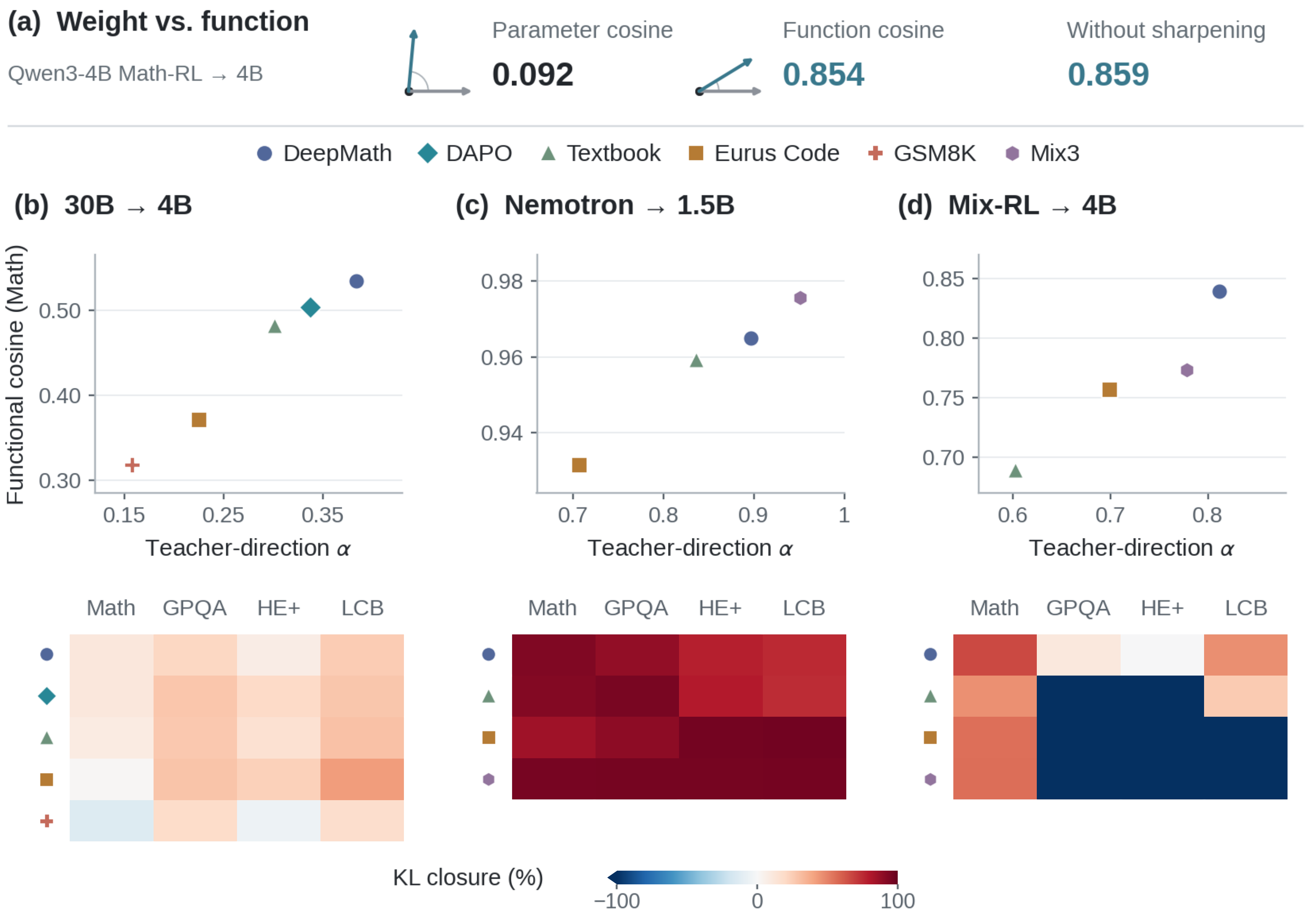}
    \caption{\textbf{Additional source-dependent functional diagnostics.}
    \textbf{(b)} 30B-to-4B. \textbf{(c)} Nemotron. \textbf{(d)} Mix-RL.
    Upper plots report Math functional cosine and teacher-direction
    coefficient $\alpha$; lower matrices report task-specific KL closure
    (\%). Closure uses teacher-first conditional KL on the initial student's
    frozen top-128 support, not a task-score recovery rate.
    Blue and red indicate negative and positive
    closure, respectively; colors saturate at $-100\%$ and $100\%$.}
    \label{fig:extended-source-diagnostics}
\end{figure}

\crossref{Figure~\ref{fig:extended-source-diagnostics}} supplements the source comparisons
in \crossref{Section~\ref{sec:functional-transfer}} with task-specific diagnostic
breakdowns. Marker shapes and colors identify the prompt sources in both
the scatter plots and the matrices. The three scatter plots use different
axis ranges. KL closure and directional alignment remain distinct
diagnostics; the large negative closure values are not percentage decreases
in benchmark accuracy.

\subsection{Additional generation-behavior evidence}
\label{app:generation-behavior}

\crossref{Table~\ref{tab:accuracy-length}} gives the complete accuracy and length
results. Model identities and evaluation protocols are
given in \crossref{Table~\ref{tab:study-pairs}} and
\crossref{Appendix~\ref{app:evaluation-protocols}}, respectively.

\begingroup
\definecolor{lengthStudentBg}{HTML}{F6EFE5}
\definecolor{lengthTeacherBg}{HTML}{EAEFF8}
\definecolor{lengthStudentText}{HTML}{8B6C48}
\definecolor{lengthTeacherText}{HTML}{5E6F9E}
\definecolor{lengthPairText}{RGB}{105,82,119}
\newcommand{\accLength}[2]{%
    \makebox[26pt][r]{#1}\,\textcolor{black!55}{/}\,%
    \makebox[31pt][r]{#2}}
\newcommand{\lengthPair}[1]{%
    \multicolumn{4}{l}{\textcolor{lengthPairText}{\textbf{#1}}}}
\small
\setlength{\tabcolsep}{5pt}
\renewcommand{\arraystretch}{1.07}
\setlength{\LTleft}{0pt}
\setlength{\LTright}{0pt}
\setlength{\LTpre}{5pt}
\setlength{\LTpost}{6pt}
\setlength{\LTcapwidth}{\linewidth}
\begin{longtable}{%
    >{\raggedright\arraybackslash}p{\dimexpr.34\linewidth-2\tabcolsep\relax}%
    *{3}{>{\centering\arraybackslash}p{\dimexpr.22\linewidth-2\tabcolsep\relax}}}
    \caption{\textbf{Accuracy and output length for seven teacher--student pairs.}
    Each cell gives mean@8 accuracy (\%) / mean output tokens;
    $s$ denotes the OPD update step.}
    \label{tab:accuracy-length}\\
    \toprule
    \textbf{Model / OPD support} & \textbf{GPQA-D}
    & \textbf{HumanEval+} & \textbf{LCB v6} \\
    \midrule
    \endfirsthead
    \caption[]{\textbf{Accuracy and output length.} Seven teacher--student pairs (continued).}\\
    \toprule
    \textbf{Model / OPD support} & \textbf{GPQA-D}
    & \textbf{HumanEval+} & \textbf{LCB v6} \\
    \midrule
    \endhead
    \midrule
    \multicolumn{4}{r}{\footnotesize Continued on the next page.}\\
    \endfoot
    \bottomrule
    \endlastfoot

    \lengthPair{(1) Math-RL: Qwen3-4B $\to$ Qwen3-4B}\\*
    \rowcolor{lengthStudentBg}
    \textcolor{lengthStudentText}{Initial (Compact)}
    & \accLength{42.74}{870} & \accLength{77.06}{325} & \accLength{17.71}{557}\\*
    \rowcolor{lengthTeacherBg}
    \textcolor{lengthTeacherText}{Teacher (GRPO-500)}
    & \accLength{52.65}{2,619} & \accLength{83.92}{700} & \accLength{20.07}{2,670}\\*
    DeepMath M48 (Compact)
    & \accLength{53.22}{5,590} & \accLength{83.23}{2,328} & \accLength{27.21}{9,567}\\
    \addlinespace[4pt]

    \lengthPair{(2) Qwen3-30B-A3B-Instruct-2507 $\to$ Qwen3-4B}\\*
    \rowcolor{lengthStudentBg}
    \textcolor{lengthStudentText}{Initial}
    & \accLength{42.74}{870} & \accLength{77.13}{325} & \accLength{17.71}{557}\\*
    \rowcolor{lengthTeacherBg}
    \textcolor{lengthTeacherText}{Teacher}
    & \accLength{54.86}{669} & \accLength{82.39}{609} & \accLength{33.14}{3,784}\\*
    DeepMath M48
    & \accLength{52.15}{5,350} & \accLength{82.32}{1,532} & \accLength{26.79}{7,599}\\*
    DAPO M48
    & \accLength{51.52}{5,378} & \accLength{82.20}{1,698} & \accLength{27.90}{8,718}\\*
    Textbook M48
    & \accLength{51.01}{6,000} & \accLength{81.90}{1,585} & \accLength{25.30}{7,814}\\*
    Eurus Code M48
    & \accLength{50.38}{5,002} & \accLength{78.40}{1,618} & \accLength{27.10}{9,474}\\*
    GSM8K M48
    & \accLength{46.84}{3,458} & \accLength{81.25}{883} & \accLength{16.79}{4,997}\\
    \addlinespace[4pt]

    \lengthPair{(3) Qwen3-30B-A3B-Instruct-2507 $\to$ Qwen3-1.7B}\\*
    \rowcolor{lengthStudentBg}
    \textcolor{lengthStudentText}{Initial}
    & \accLength{29.10}{725} & \accLength{59.83}{198} & \accLength{12.36}{422}\\*
    \rowcolor{lengthTeacherBg}
    \textcolor{lengthTeacherText}{Teacher}
    & \accLength{54.86}{669} & \accLength{82.39}{609} & \accLength{33.14}{3,784}\\*
    DeepMath M48
    & \accLength{36.24}{6,716} & \accLength{66.70}{1,731} & \accLength{18.10}{8,053}\\*
    DAPO M48
    & \accLength{33.71}{7,218} & \accLength{64.10}{2,077} & \accLength{16.90}{9,015}\\
    \addlinespace[4pt]

    \lengthPair{(4) JustRL $\to$ DeepSeek-R1-Distill-Qwen-1.5B}\\*
    \rowcolor{lengthStudentBg}
    \textcolor{lengthStudentText}{Initial}
    & \accLength{35.92}{10,290} & \accLength{54.34}{5,475} & \accLength{13.50}{14,946}\\*
    \rowcolor{lengthTeacherBg}
    \textcolor{lengthTeacherText}{Teacher}
    & \accLength{40.47}{7,217} & \accLength{64.56}{5,252} & \accLength{19.36}{11,783}\\*
    DeepMath M8, s100
    & \accLength{38.83}{7,583} & \accLength{63.34}{4,546} & \accLength{17.43}{11,439}\\*
    DAPO M8, s100
    & \accLength{40.91}{7,937} & \accLength{64.10}{4,777} & \accLength{18.64}{11,514}\\
    \addlinespace[4pt]

    \lengthPair{(5) Nemotron $\to$ DeepSeek-R1-Distill-Qwen-1.5B}\\*
    \rowcolor{lengthStudentBg}
    \textcolor{lengthStudentText}{Initial}
    & \accLength{35.92}{10,290} & \accLength{54.34}{5,475} & \accLength{13.50}{14,946}\\*
    \rowcolor{lengthTeacherBg}
    \textcolor{lengthTeacherText}{Teacher}
    & \accLength{40.66}{5,185} & \accLength{66.69}{4,760} & \accLength{22.00}{6,113}\\*
    DeepMath M48, s75
    & \accLength{40.09}{6,060} & \accLength{64.86}{3,882} & \accLength{19.93}{7,521}\\*
    Textbook M48, s75
    & \accLength{39.84}{6,308} & \accLength{64.41}{3,914} & \accLength{19.00}{7,447}\\*
    Eurus Code M48, s75
    & \accLength{40.09}{4,910} & \accLength{65.55}{5,308} & \accLength{22.00}{7,002}\\*
    Mix3 M48, s75
    & \accLength{42.17}{5,933} & \accLength{66.16}{4,752} & \accLength{21.79}{7,008}\\
    \addlinespace[4pt]

    \lengthPair{(6) Mix-RL $\to$ Qwen3-4B-Instruct-2507}\\*
    \rowcolor{lengthStudentBg}
    \textcolor{lengthStudentText}{Initial}
    & \accLength{45.52}{458} & \accLength{81.48}{976} & \accLength{27.86}{2,944}\\*
    \rowcolor{lengthTeacherBg}
    \textcolor{lengthTeacherText}{Teacher}
    & \accLength{45.27}{505} & \accLength{79.57}{1,103} & \accLength{30.21}{4,808}\\*
    DeepMath M48, s25
    & \accLength{47.35}{555} & \accLength{81.86}{1,207} & \accLength{29.71}{3,371}\\*
    Textbook M48, s25
    & \accLength{51.14}{952} & \accLength{83.23}{1,332} & \accLength{28.93}{4,051}\\*
    Eurus Code M48, s25
    & \accLength{61.81}{3,432} & \accLength{81.17}{1,452} & \accLength{31.93}{7,940}\\*
    IF M48, s25
    & \accLength{46.46}{473} & \accLength{68.29}{133} & \accLength{22.07}{1,769}\\*
    Agent M48, s25
    & \accLength{50.88}{992} & \accLength{78.35}{768} & \accLength{28.21}{3,958}\\*
    Mix3 M48, s25
    & \accLength{59.34}{2,804} & \accLength{82.01}{1,535} & \accLength{32.36}{7,770}\\*
    Mix5 M48, s25
    & \accLength{55.18}{1,698} & \accLength{80.18}{1,593} & \accLength{31.14}{7,956}\\
    \addlinespace[4pt]

    \lengthPair{(7) Light-R1 $\to$ DeepSeek-R1-Distill-Qwen-7B}\\*
    \rowcolor{lengthStudentBg}
    \textcolor{lengthStudentText}{Initial}
    & \accLength{49.05}{8,414} & \accLength{81.78}{3,782} & \accLength{23.36}{11,998}\\*
    \rowcolor{lengthTeacherBg}
    \textcolor{lengthTeacherText}{Teacher}
    & \accLength{46.97}{9,078} & \accLength{82.93}{3,865} & \accLength{24.14}{12,515}\\*
    DAPO M8, s50
    & \accLength{49.24}{9,366} & \accLength{82.55}{4,108} & \accLength{25.14}{12,740}\\*
    DeepMath M8, s50
    & \accLength{49.49}{9,283} & \accLength{82.62}{3,814} & \accLength{24.21}{12,180}\\*
    Light3533, s50
    & \accLength{50.63}{9,630} & \accLength{81.78}{4,249} & \accLength{24.00}{12,617}\\
\end{longtable}
\endgroup

Across three Mix-RL Code48 support draws, GPQA scores range from
$61.24$ to $61.81$ and mean output length from 3,308 to 3,432 tokens.
These are support-draw replications, not independent training-seed estimates.

Code48 and Mix3 use 3,428 and 2,800 pre-answer tokens and have repeated
8-gram rates of $2.13\%$ and $1.83\%$, respectively. An early explicit
answer candidate is identifiable in 244 and 261 responses, approximately
$15$--$16\%$ of the evaluated outputs. Within these subsets,
wrong-to-correct revisions number 5 and 8, compared with 10 and 11
correct-to-wrong revisions. The negative net counts refer only to these
observable explicit changes; they do not measure implicit reasoning or
establish a causal decomposition of the overall gain.

Similar output lengths can also accompany different accuracy:
30B-to-4B GSM8K48 and DeepMath48 produce 9,277 and 9,536 mean Math
tokens but score $31.15$ and $42.81$, respectively.
These observations concern free generation at evaluation, not the
training-token consumption controlled in \crossref{Section~\ref{sec:source-preferences}};
they do not establish that longer test-time generation alone causes gains.

\subsection{Failure trajectories and recovery protocols}
\label{app:recovery-protocols}
\label{app:code-recovery-pca}

\begin{figure}[!htb]
    \centering
    \includegraphics[width=\linewidth]{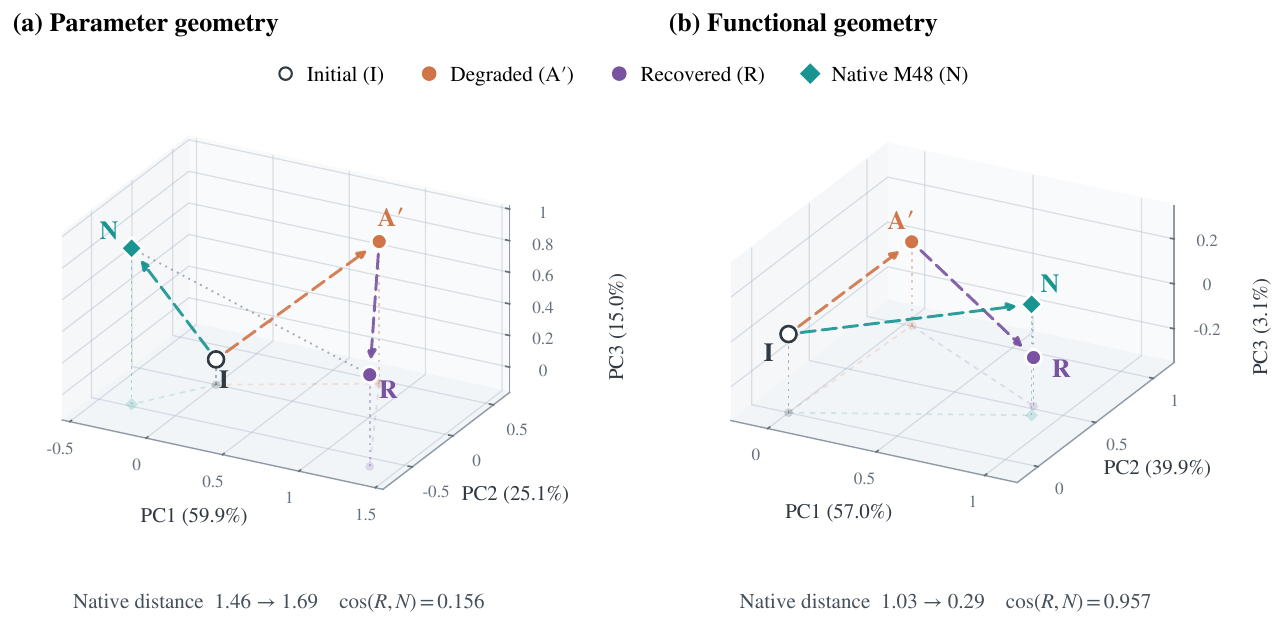}
    \caption{\textbf{Code recovery in parameter and function space.}
    PCA views of the initial student ($I$), degraded single-prompt endpoint
    ($A'$), recovered endpoint ($R$), and native Code-M48 endpoint ($N$).
    Native Code-M48 is trained directly from $I$; recovery continues from
    $A'$ on an effective 48-prompt support.
    From $A'$ to $R$, the reported distance to $N$ increases from $1.46$
    to $1.69$ in parameter space but decreases from $1.03$ to $0.29$
    in function space. Axis percentages report explained variance;
    arrows connect endpoints rather than tracing intermediate updates.}
    \label{fig:code-recovery-endpoint-pca}
\end{figure}

\crossref{Figure~\ref{fig:code-recovery-endpoint-pca}} highlights the geometry of Code
recovery relative to native Code-M48. Continued distillation brings the
recovered endpoint closer to this reference in function space even as
its parameter-space distance increases. The distance to native Code-M48
decreases by approximately $71\%$ in function space but increases by
approximately $15\%$ in parameter space.

\begin{figure}[!htbp]
    \centering
    \includegraphics[width=\linewidth]{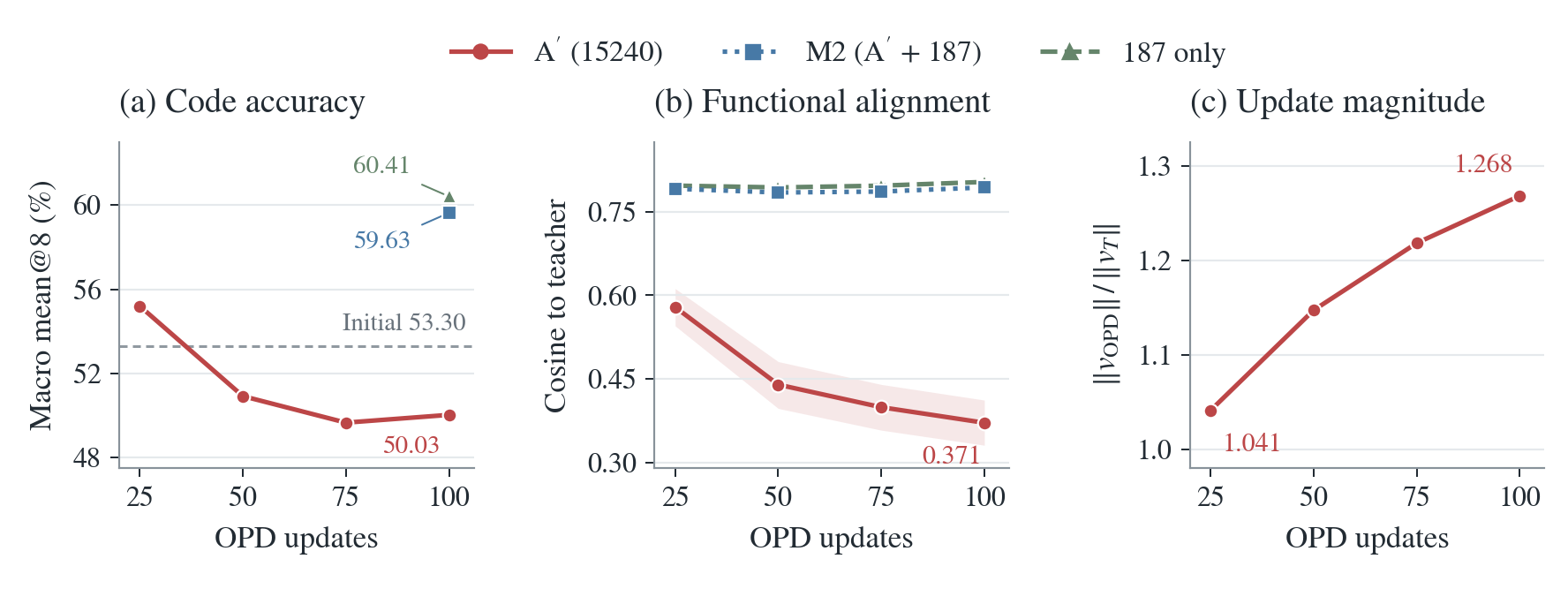}
    \caption{\textbf{Training trajectories under single-prompt and two-prompt supports.}
    All runs use the same Qwen3-4B initialization and Code GRPO-300 teacher.
    $A'$ uses prompt 15240; M2 jointly uses prompts 15240 and 187 from
    initialization.
    \textbf{(a)} Code accuracy: the unweighted mean of mean@8 over
    HumanEval+, MBPP+, and LCB-v6. Markers for M2 and prompt 187 alone
    show only their step-100 endpoints.
    \textbf{(b)} Functional cosine to the teacher's change.
    \textbf{(c)} Normalized functional-update magnitude of $A'$,
    $\lVert v_{\mathrm{OPD}}\rVert/\lVert v_T\rVert$.}
    \label{fig:failure-trajectories}
\end{figure}

\paragraph{Failure trajectories.}
\crossref{Figure~\ref{fig:failure-trajectories}} traces how $A'$ diverges during OPD
and contrasts its functional alignment with the two-prompt support and
prompt 187 alone. The increasing update magnitude is measured in
function space, not parameter space.

For $A'$ (prompt ID $15240$), checkpoints 25, 50, 75, and 100 have
Code Macro scores $55.20$, $50.92$, $49.66$, and $50.03$, compared
with the initial student's $53.30$. At the same checkpoints, functional
cosine to the teacher is $0.5787$, $0.4397$, $0.3992$, and $0.3713$,
and normalized functional-update magnitude is $1.041$, $1.147$,
$1.218$, and $1.268$. Accuracy is not strictly monotonic: it rises
slightly between steps 75 and 100 despite the overall degradation.
For the joint $A'+187$ support, the corresponding cosines are
$0.7910$, $0.7843$, $0.7860$, and $0.7934$; for prompt $187$ alone,
they are $0.7967$, $0.7934$, $0.7966$, and $0.8033$.
The joint support is trained from initialization, not used as a
continuation of $A'$.
The joint support achieves a Code Macro score of $59.63$, compared with
$60.41$ for prompt 187 alone and $50.03$ for prompt 15240 alone.
Thus, a prompt that degrades performance in isolation need not prevent
strong transfer when combined with an effective prompt.

\paragraph{Bidirectional representation intervention.}
The effective endpoint $B$ is trained on prompt 5038; $A'$ is the
degraded endpoint above. The intervention replaces the output hidden
state of zero-indexed block 30 (the 31st block), at the current prediction
position on every autoregressive step. It copies the complete hidden
vector, not weights or selected coordinates. Donor and recipient maintain
separate, synchronously updated KV caches.
The probe contains 20 questions with eight samples per question.
$B\to A'$ changes the probe pass rate from $18.750\%$ to $24.375\%$;
$A'\to B$ changes it from $28.125\%$ to $21.250\%$;
$A'\to A'$ leaves it unchanged. These are probe rates, not the full
Code Macro evaluations used for continuation. The intervention requires
a donor at evaluation and is not a standalone student or an efficiency claim.

\crossref{Figure~\ref{fig:transfer-failure-recovery}(b)} also reports replacement at
zero-based block 10, with aggregate changes of $-1.3$ and $-1.9$
percentage points for $B\to A'$ and $A'\to B$, respectively.
At block 30, $B\to A'$ improves HumanEval+ and MBPP+ but lowers LCB;
the reverse intervention has the opposite benchmark-level signs.
The tested interfaces show location- and task-dependent effects, not a
uniquely responsible layer or uniform benefit across benchmarks.

\paragraph{Code continuation: protocol and endpoints.}
Code recovery starts from the step-100 $A'$ weights and trains for a
further 100 updates on an effective 48-prompt support with the same
Code GRPO-300 teacher. It uses batch size 256, learning rate $10^{-5}$,
sampled-token K1 policy-gradient feedback, and a fresh Adam optimizer.
Training and evaluation use seed 42; evaluation reports mean@8 on
HumanEval+, MBPP+, and LCB-v6 with temperature 1 and top-p 1.
Only the prespecified continuation step-100 endpoint is evaluated.
The initial student, teacher, and native M48 reference have Code Macro
scores $53.30$, $61.80$, and $61.79$, respectively; the recovered
endpoint scores $63.16$. Its 200 total updates are not matched to
the native reference's 100. This combined intervention does not show
that starting from an unsuccessful support is preferable.

\paragraph{Math continuation: protocol, endpoints, and controls.}
\label{app:support-switch-controls}
For Qwen3-30B-A3B-Instruct-2507 $\to$ Qwen3-4B, we branch each
stage-1 checkpoint, trained for 15 updates on GSM8K ($G$) or DeepMath
($D$), into two continuations: a further 15 updates on $G$ or on $D$.
Thus, $G\to G$ and $G\to D$ share the same starting weights, as do
$D\to G$ and $D\to D$; all four paths have 30 total updates.
All four stage-2 runs use a fresh Adam optimizer and retain the learning
rate, batch size, and training seed of the original 30B-to-4B configuration
in \crossref{Appendix~\ref{app:experimental-details}}.
Optimizer resets and additional update counts are therefore common
to the paired continuation arms; token budgets need not be equal.

For Math mean@16, $G\to D$ reaches $43.65$ from $31.15$; its
matched-update performance reference $D\to D$ scores $43.75$.
LCB remains lower after the GSM8K history ($25.36$ versus $26.93$),
so the similar Math aggregate does not erase every task-specific difference.
\crossref{Figure~\ref{fig:support-switch-main}(b)} compares formal Avg$_4$
endpoints: continuing on DeepMath rather than GSM8K yields $51.75$
versus $43.77$ from $G$, and $51.98$ versus $46.38$ from $D$.
Thus, the continuation support matters at matched update counts, while
the two DeepMath continuations reach close scores.

\begin{figure}[!htb]
    \centering
    \setlength{\abovecaptionskip}{4pt}
    \includegraphics[width=0.50\linewidth]{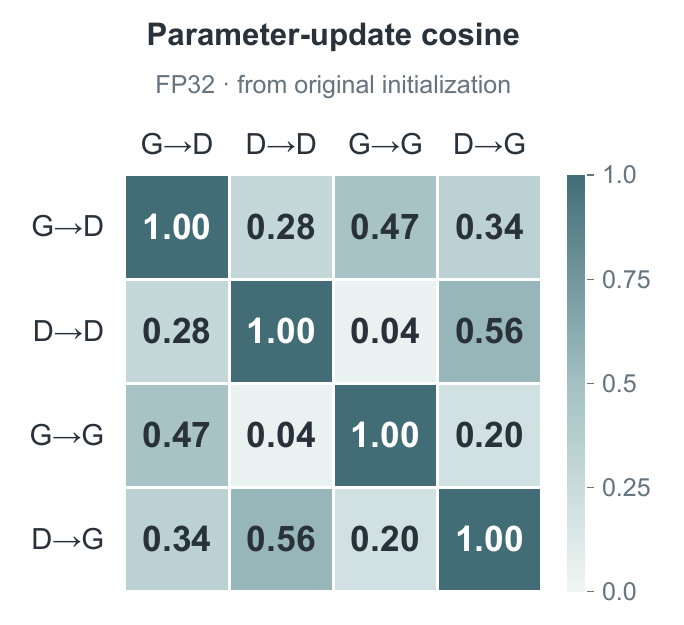}
    \caption{\textbf{Parameter displacements after bidirectional support switching.}
    Cosines compare FP32 endpoint-minus-original-initialization vectors
    for the four paths in \crossref{Figure~\ref{fig:support-switch-main}}.
    Every endpoint has 30 total updates; $G$ denotes GSM8K and $D$ denotes
    DeepMath. The $G\to D$ versus $D\to D$ comparison uses the step-30
    reference, unlike the step-15 Math reference in
    \crossref{Figure~\ref{fig:transfer-failure-recovery}(d)}.}
    \label{fig:support-switch-controls}
\end{figure}

\crossref{Figure~\ref{fig:support-switch-main}(a)} uses AIME 2024/2025 validation mean@2;
at stage-2 step 0, each pair of branches reuses its shared stage-1
validation result, not a measurement before OPD. Panel (b)'s
Avg$_4$ averages Math mean@16 and GPQA-D, HumanEval+, and LCB-v6 mean@8.
The supplementary parameter matrix is retained in
\crossref{Figure~\ref{fig:support-switch-controls}}, with its references specified below.

\paragraph{Geometry references and parameter paths.}
\crossref{Figure~\ref{fig:transfer-failure-recovery}(d)} uses native Code-M48 at
step 100 as the Code reference and native DeepMath at step 15 as the
Math reference, not the step-30 Math performance reference.
Relative to these fixed references, functional cosine rises
from $0.474$ to $0.957$ for Code and from $0.687$ to $0.938$ for Math.
Parameter-displacement cosine changes from $0.123$ to $0.156$ for Code
and from $0.086$ to $0.374$ for Math.
These functional comparisons use a native OPD endpoint, whereas the
failure-trajectory diagnostic compares with the teacher's change.

Endpoint parameter displacements in these cosine comparisons are measured
from the original initialization $\theta_0$, not the stage-2 starting
checkpoint. \crossref{Figure~\ref{fig:support-switch-controls}} compares FP32 endpoint
displacements and gives $0.28$ for $G\to D$ versus $D\to D$ at step 30.
The Math value $0.374$ in \crossref{Figure~\ref{fig:transfer-failure-recovery}(d)}
instead uses DeepMath step 15 as the comparator; the two values are not
interchangeable.

Let $\Delta_{\mathrm{bad}}=\theta_{\mathrm{bad}}-\theta_0$ and
$U=\theta_{\mathrm{recovered}}-\theta_{\mathrm{bad}}$.
Then $\cos(U,-\Delta_{\mathrm{bad}})$ is $0.208$ for Code and $0.178$
for Math: continuation does not simply reverse the earlier parameter
displacement. These comparisons do not establish exact removal of earlier
functional changes. Likewise, high functional cosine indicates directional
agreement, not equal amplitudes, conditional distributions, or predictions.

\paragraph{Robustness to answer scoring.}
We rescore all 9,600 outputs from five checkpoints across four mathematics
benchmarks using math-verify 0.9.0 \citep{kydlicek2026mathverify} and
conservative final-answer anchors.
Under this alternative scoring pipeline, $G\to D$ improves over $G$ by
12.24 percentage points (95\% CI $[8.70, 16.04]$) and over $G\to G$ by
15.16 points (95\% CI $[11.41, 19.17]$). Its difference from $D\to D$ is
$-0.16$ points (95\% CI $[-2.34, +2.08]$).
A blinded audit of 152 stratified cases identifies 46 explicit final
answers missed by the original parser, of which only six are correct.
These counts describe the audited sample, not the prevalence of missed
answers across all outputs.
These results support recovery in evaluated performance that is robust
to the tested scoring procedure; missed final-answer extraction is
insufficient to explain the main gain. They do not establish recovery
of an underlying reasoning mechanism.

\section{Prompt Selection and Mixed-Pool Sampling}
\label{app:selection}

This appendix details the within-pool selectors and paired mixed-pool
comparison in \crossref{Section~\ref{sec:prompt-selection}}, together with additional
selection experiments on code and 30B-to-4B distillation.
It also specifies the initial-gradient representativeness diagnostic.
\crossref{Table~\ref{tab:prompt-selection}} reports the selector results and
selection costs; \crossref{Table~\ref{tab:mixed-pool-selection}} reports the
source-filtering comparison.

\subsection{Prompt-selection methods}
\label{app:prompt-selection-methods}

\crossref{Table~\ref{tab:prompt-selection}} compares eight selectors for distilling
\modelname{JustRL-DeepSeek-1.5B} into
\modelname{DeepSeek-R1-Distill-Qwen-1.5B}, using DeepMath candidates.
Let $\mathcal C$ be a candidate pool of size $C$ and $\mathcal S$ the
selected support of size $M$. The two reported settings use
$(M,C)=(8,384)$ and $(48,2{,}304)$. A nested ranking refers to prefixes
within the same fixed candidate pool, not nesting between these two
different candidate-pool settings. Selection is performed before OPD,
using the initial student wherever student-dependent features are needed.

\paragraph{Uniform random.}
We sample $M$ prompts uniformly without replacement from $\mathcal C$.
This baseline tests whether a small unscored sample is sufficient.
The random selectors in \crossref{Table~\ref{tab:prompt-selection}} report the mean
and sample standard deviation across three independently sampled supports;
these are not standard errors or training-seed variations.
For uniform random at $M=8$, $C=384$, the support-selection seeds are
42, 43, and 44, with the training seed held fixed.

\paragraph{Stratified random.}
We jointly stratify candidates by data source, prompt character-length
quartile, and initial-student difficulty. Difficulty is measured by the
mean official reward of the first eight discovery rollouts:
\begin{equation}
    \widehat p_i = \frac{1}{8}\sum_{r=1}^{8} R(y_{ir}),
    \label{eq:selection-student-pass-rate}
\end{equation}
where $R$ is the official correctness reward. The four difficulty bins
are zero ($\widehat p_i=0$), low ($0<\widehat p_i<0.5$), medium
($0.5\leq\widehat p_i<1$), and perfect ($\widehat p_i=1$).
All candidates are from DeepMath-103K,
so the source dimension introduces no further subdivision; the joint
partition has up to $4\times4$ length--difficulty strata.

Stratum quotas are proportional to their candidate-pool frequencies,
not equal across difficulty bins. A cumulative quota-deficit rule selects
one stratum at a time, so successive prefixes approximate the original
joint distribution. Prompts within each stratum follow a seeded SHA-256
permutation; the reference ranking uses seed 42.
This selector tests whether preserving length and difficulty composition
helps avoid omissions in a small support.

\paragraph{Semantic diversity.}
We embed the full prompt messages, serialized as
\texttt{<role>: <content>}, with
\modelname{Qwen/Qwen3-Embedding-0.6B} \citep{zhang2025qwen3embedding}.
Inputs are truncated to 2,048 tokens.
The final non-padding token's hidden state is L2-normalized to obtain
$e_i$. Selection uses deterministic farthest-first traversal, not
$k$-means: the first prompt has the highest cosine similarity to the
global embedding center, and each subsequent prompt maximizes
\begin{equation}
    i^* = \arg\max_{i\in\mathcal C\setminus\mathcal S}
        \min_{j\in\mathcal S}\bigl(1-\cos(e_i,e_j)\bigr).
    \label{eq:selection-farthest-first}
\end{equation}
The resulting order supplies nested prefixes within a fixed pool.
There are no semantic-cluster quotas, and answers, rewards, and teacher
feedback are not used. The method tests broad semantic coverage with
reduced redundancy.

\paragraph{Hard selection.}
We rank prompts by increasing initial-student pass rate
$\widehat p_i$ from \crossref{Equation~\ref{eq:selection-student-pass-rate}} and retain
the first $M$. Ties follow a seed-42 SHA-256 order.
The teacher does not contribute to this difficulty ranking; it supplies
token-level feedback during subsequent distillation.
This tests whether prompts that the initial student finds harder provide
more useful training supervision.

\paragraph{Shortest.}
We rank candidates by the initial student's mean generated response
length and retain the $M$ shortest. This tests selection for cheaper
generated trajectories rather than shorter prompt text.

\paragraph{Longest.}
We use the same initial-student response-length statistic but retain the
$M$ longest candidates. This tests whether longer trajectories offer
more opportunities for distillation.

\paragraph{Teacher--student disagreement.}
The initial student generates $K=16$ discovery responses per candidate,
and the teacher scores the same sampled tokens. We rank prompts by a
length-normalized, clipped Monte Carlo reverse-KL score. For sampled
token $y_{irt}$ at prefix $h_{irt}$, define
\begin{align}
    q_{irt}
    &=\operatorname{clip}\!\left(
        \log p_0(y_{irt}\mid h_{irt})-
        \log p_T(y_{irt}\mid h_{irt}),-10,10\right),
    \label{eq:selection-disagreement-token}\\
    D_i
    &=\frac{1}{K}\sum_{r=1}^{K}
        \left[\frac{1}{T_{ir}}\sum_{t=1}^{T_{ir}}q_{irt}\right].
    \label{eq:selection-disagreement-score}
\end{align}
Here $p_0$ denotes the initial student, $p_T$ the teacher, and $T_{ir}$
the response length in tokens. We select the $M$ largest $D_i$ values.
Tokens are averaged within each response before the 16 response means
are averaged, so longer responses do not receive greater weight solely
because of their length. This uses signed student-minus-teacher
log-ratios, not absolute differences or a full-vocabulary KL computation.

\paragraph{Cost-D-opt.}
This selector favors complementary gradient directions while accounting
for response-length cost. Its base construction is D-optimal selection;
the cost-aware greedy score, rather than the gradient features, distinguishes
Cost-D-opt from pure D-opt.

For each candidate $i$, the initial student generates 16 discovery
responses. The frozen teacher scores the same sampled tokens. At prefix
$h_{irt}$ in response $r$, the detached token feedback is
\begin{equation}
    a_{irt}=\operatorname{sg}\!\left[
        \operatorname{clip}\!\left(
            \log p_T(y_{irt}\mid h_{irt})-
            \log p_0(y_{irt}\mid h_{irt}),-10,10
        \right)\right],
    \label{eq:selection-gradient-feedback}
\end{equation}
where $p_0$ is the initial student and $\operatorname{sg}$ denotes
stop-gradient. The response-level token-feedback loss is
\begin{equation}
    \ell_{ir}(\theta)=-\frac{1}{T_{ir}}
        \sum_{t=1}^{T_{ir}}a_{irt}\log p_\theta(y_{irt}\mid h_{irt}),
    \label{eq:selection-gradient-loss}
\end{equation}
with gradients evaluated at the initial student. The implementation also
applies rollout importance correction before gradient sketching; this
factor is omitted from \crossref{Equation~\ref{eq:selection-gradient-loss}} to display
the token-feedback component. Three CountSketch projections
\citep{charikar2002finding} avoid
retaining full parameter gradients. Let $\widetilde g_{ir}$ denote the
resulting correction-applied, sketched response gradient. Prompt-level
aggregation is token-weighted:
\begin{equation}
    \widetilde g_i=
        \frac{\sum_{r=1}^{16}T_{ir}\widetilde g_{ir}}
             {\sum_{r=1}^{16}T_{ir}},
    \qquad L_i=\sum_{r=1}^{16}T_{ir}.
    \label{eq:selection-gradient-aggregation}
\end{equation}

The 16 responses are split into two fixed eight-response subsets.
Their cross-covariance provides a noise correction, retaining positive
directions reproducible across the two subsets. The retained stable
space has dimension $d\leq8$, and $z_i$ denotes the stable coordinates
of prompt $i$. D-opt uses $x_i=\sqrt{L_i}\,z_i$ and the objective
\begin{equation}
    F(\mathcal S)=\log\det\!\left(
        \lambda I+\sum_{i\in\mathcal S}x_i x_i^\top\right),
    \qquad \lambda=10^{-3}\frac{\operatorname{tr}(X^\top X)}{d},
    \label{eq:selection-d-opt}
\end{equation}
where row $i$ of $X$ is $x_i^\top$. Starting from $A=\lambda I$, the
marginal gain of candidate $i$ is
\begin{equation}
    \Delta_i=\log\!\left(1+x_i^\top A^{-1}x_i\right).
    \label{eq:selection-d-opt-gain}
\end{equation}
Pure D-opt selects the largest $\Delta_i$; Cost-D-opt selects the largest
$\Delta_i/c_i$, where $c_i>0$ is the relative response-length cost.
After selecting $i$, both update $A\leftarrow A+x_i x_i^\top$ and repeat
until $M$ prompts have been selected.

Pure D-opt uses neither correctness nor semantic embeddings, and it
does not penalize response length. Indeed, $L_i$ gives token-rich prompts
greater information weight, potentially favoring long responses.
Cost-D-opt additionally normalizes the greedy gain by length cost.
\crossref{Table~\ref{tab:prompt-selection}} reports the cost-aware variant; pure
D-opt is described here only to specify its underlying objective.
Both share the gradient-statistics computation and its GPU cost.

\begingroup
\raggedbottom
\subsection{Mixed-pool and source-filtered sampling}
\label{app:mixed-pool-selection}

For the three-source filtering comparison, the student is
\modelname{Qwen3-4B}, trained with
\texttt{enable\_thinking=False}, and the teacher is
\modelname{Qwen3-30B-A3B-Instruct-2507}.
Each run in this comparison uses $M=48$ prompts and 25 optimization steps. We vary the
support-selection seed across 42/43/44 while holding the training seed fixed.
Evaluation uses thinking off, temperature $1.0$, top-$p=1.0$, and a
16,384-token response limit, with 16 samples per Math question and eight
per OOD question. Other training and evaluation settings follow the
30B-Instruct-to-4B protocols in
\crossref{Appendices~\ref{app:experimental-details} and~\ref{app:evaluation-protocols}}.

\paragraph{Support construction.}
Candidates are formed by first drawing 3,840 prompts uniformly from each
deduplicated source (DeepMath, DAPO and GSM8K; 11,520 in total, shared
across selection seeds). Each selection seed then draws a single uniform
per-prompt ordering over this pool; Mixed takes its first 48 prompts, and
Filtered removes GSM8K from the same ordering before taking the first 48.
\crossref{Table~\ref{tab:mixed-pool-composition}} reports the source counts and overlap.

\par\addvspace{6pt}\noindent
\begin{minipage}{\linewidth}
\makeatletter\def\@captype{table}\makeatother
\centering
\setlength{\abovecaptionskip}{0pt}
\setlength{\belowcaptionskip}{4pt}
\caption{\textbf{Mixed-pool support composition and overlap.}
Shared counts prompt instances selected by both arms under the same
random ordering, not shared sampled responses.}
\label{tab:mixed-pool-composition}
\small
\setlength{\tabcolsep}{9pt}
\renewcommand{\arraystretch}{1.12}
\begin{tabular}{@{}lccc@{}}
\toprule
Selection seed & Mixed & Filtered & Shared prompts \\
 & DeepMath / DAPO / GSM8K & DeepMath / DAPO / GSM8K & \\
\midrule
42 & 17 / 15 / 16 & 21 / 27 / 0 & 32 \\
43 & 13 / 15 / 20 & 22 / 26 / 0 & 28 \\
44 & 20 / 13 / 15 & 31 / 17 / 0 & 33 \\
\bottomrule
\end{tabular}
\end{minipage}\par\addvspace{6pt}

\paragraph{Per-seed and per-benchmark results.}
\crossref{Tables~\ref{tab:mixed-pool-math-results} and~\ref{tab:mixed-pool-ood-results}}
report the complete results; OOD averages GPQA-Diamond, HumanEval+,
and LiveCodeBench v6 equally. Paired summaries use unrounded scores,
so displayed differences need not equal calculations from rounded entries.

\par\addvspace{8pt}\noindent
\begin{minipage}{\linewidth}
\makeatletter\def\@captype{table}\makeatother
\centering
\small
\setlength{\tabcolsep}{4pt}
\renewcommand{\arraystretch}{1.15}
\setlength{\abovecaptionskip}{0pt}
\setlength{\belowcaptionskip}{4pt}
\caption{\textbf{Math results for mixed-pool and source-filtered sampling.}
30B Instruct $\rightarrow$ Qwen3-4B, $M=48$, 25 optimization steps;
selection seeds 42/43/44 share a fixed training seed.
Scores are mean@16 (\%); Math average is the unweighted mean of the four
benchmarks. $\Delta$ denotes paired Filtered $-$ Mixed differences;
means, SDs, and differences use unrounded scores.}
\label{tab:mixed-pool-math-results}
\begin{tabular*}{\linewidth}{@{\extracolsep{\fill}}llccc@{}}
\toprule
Benchmark & Selection seed & Mixed & Filtered & $\Delta$ \\
\midrule
Math average & 42 / 43 / 44
 & 43.2 / 40.3 / 38.8 & 41.4 / 39.3 / 40.7 & $-1.8$ / $-0.9$ / $+1.9$ \\
 & Mean $\pm$ SD & $40.7\pm2.3$ & $40.5\pm1.0$ & $-0.3\pm2.0$ \\
\midrule
AIME24 & 42 / 43 / 44
 & 55.8 / 54.2 / 49.2 & 56.5 / 51.3 / 49.8 & $+0.6$ / $-2.9$ / $+0.6$ \\
 & Mean $\pm$ SD & $53.1\pm3.5$ & $52.5\pm3.5$ & $-0.6\pm2.0$ \\
\addlinespace[4pt]
AIME25 & 42 / 43 / 44
 & 50.4 / 48.1 / 43.8 & 44.4 / 46.7 / 46.9 & $-6.0$ / $-1.5$ / $+3.1$ \\
 & Mean $\pm$ SD & $47.4\pm3.4$ & $46.0\pm1.4$ & $-1.5\pm4.6$ \\
\addlinespace[4pt]
HMMT-Feb & 42 / 43 / 44
 & 28.8 / 24.6 / 29.0 & 27.7 / 26.5 / 30.0 & $-1.0$ / $+1.9$ / $+1.0$ \\
 & Mean $\pm$ SD & $27.4\pm2.5$ & $28.1\pm1.8$ & $+0.6\pm1.5$ \\
\addlinespace[4pt]
HMMT-Nov & 42 / 43 / 44
 & 37.7 / 34.2 / 33.1 & 36.9 / 32.9 / 36.0 & $-0.8$ / $-1.3$ / $+2.9$ \\
 & Mean $\pm$ SD & $35.0\pm2.4$ & $35.3\pm2.1$ & $+0.3\pm2.3$ \\
\bottomrule
\end{tabular*}
\end{minipage}\par\addvspace{8pt}

\begin{table}[!t]
\centering
\small
\setlength{\tabcolsep}{4pt}
\renewcommand{\arraystretch}{1.15}
\setlength{\abovecaptionskip}{0pt}
\setlength{\belowcaptionskip}{4pt}
\caption{\textbf{OOD results for mixed-pool and source-filtered sampling.}
30B Instruct $\rightarrow$ Qwen3-4B, $M=48$, 25 optimization steps;
selection seeds 42/43/44 share a fixed training seed.
Scores are mean@8 (\%); OOD average is the unweighted mean of GPQA-D,
HE+, and LCB. $\Delta$ denotes paired Filtered $-$ Mixed differences;
means, SDs, and differences use unrounded scores.}
\label{tab:mixed-pool-ood-results}
\begin{tabular*}{\linewidth}{@{\extracolsep{\fill}}llccc@{}}
\toprule
Benchmark & Selection seed & Mixed & Filtered & $\Delta$ \\
\midrule
OOD average & 42 / 43 / 44
 & 55.2 / 54.9 / 53.9 & 55.2 / 53.7 / 54.6 & $0.0$ / $-1.2$ / $+0.7$ \\
 & Mean $\pm$ SD & $54.7\pm0.7$ & $54.5\pm0.8$ & $-0.2\pm0.9$ \\
\midrule
GPQA-D & 42 / 43 / 44
 & 51.5 / 50.7 / 50.0 & 51.0 / 50.4 / 51.2 & $-0.4$ / $-0.3$ / $+1.2$ \\
 & Mean $\pm$ SD & $50.7\pm0.7$ & $50.9\pm0.4$ & $+0.2\pm0.9$ \\
\addlinespace[4pt]
HE+ & 42 / 43 / 44
 & 87.0 / 87.0 / 86.1 & 88.1 / 85.3 / 86.1 & $+1.1$ / $-1.8$ / $+0.1$ \\
 & Mean $\pm$ SD & $86.7\pm0.6$ & $86.5\pm1.4$ & $-0.2\pm1.4$ \\
\addlinespace[4pt]
LCB & 42 / 43 / 44
 & 27.1 / 26.9 / 25.7 & 26.4 / 25.4 / 26.4 & $-0.6$ / $-1.5$ / $+0.7$ \\
 & Mean $\pm$ SD & $26.5\pm0.7$ & $26.1\pm0.6$ & $-0.5\pm1.1$ \\
\bottomrule
\end{tabular*}
\end{table}

\paragraph{Source composition and a four-prompt control.}
\crossref{Figure~\ref{fig:mixture-composition}} in the main text reports a
two-source composition sweep with 0, 4, 16, or 48 DeepMath prompts and
GSM8K filling the remaining positions in each 48-prompt support.
Within each selection seed, the $M=4$ control reuses exactly the four
DeepMath prompts in the corresponding 4/44 mixture. Error bars and bands
report sample SD across selection seeds 42/43/44.
The sweep and control use the same 30B-to-4B pair and 25-update configuration
as above, with the training seed, batch size, learning rate, and thinking-off
training and evaluation settings unchanged.
Changing support size may also change prompt repetition; this is not a
matched-exposure test of complementary information from the two sources.

\endgroup

\subsection{Prompt selection within Eurus Code}
\label{app:code-prompt-selection}

We additionally compare five selectors for the Qwen3-4B code-RL
continuation pair, distilling the Code-RL teacher (GRPO-300) into
Qwen3-4B. Each method selects $M=48$ prompts from $C=2{,}304$
Eurus Code candidates and trains for 100 optimization steps.
Uniform random reports the mean and standard deviation across three
independently sampled supports. The method names follow
\crossref{Appendix~\ref{app:prompt-selection-methods}}.

\begin{table}[!htbp]
    \centering
    \caption{\textbf{Prompt selection within Eurus Code.}
    Code-RL 4B $\to$ Qwen3-4B; $M=48$, $C=2{,}304$, 100 updates.
    Scores (\%) use mean@8; Code Macro averages the three benchmarks.
    Uniform random reports mean $\pm$ sample SD across three support draws.
    Bold marks column-best scores. Costs cover selection only;
    $0$ denotes no GPU screening.}
    \label{tab:code-prompt-selection}
    \begingroup
    \small
    \setlength{\tabcolsep}{2.8pt}
    \renewcommand{\arraystretch}{1.10}
    \newcommand{\codeselcell}[2]{%
        \makebox[22.5pt][r]{#1}%
        \makebox[25pt][l]{%
            \if\relax\detokenize{#2}\relax\else
                {\fontsize{8}{9}\selectfont\textcolor[HTML]{596579}{\,$\pm$\,#2}}%
            \fi}%
    }
    \begin{tabular*}{\linewidth}{@{}p{86pt}@{\hspace{5pt}}c@{\extracolsep{\fill}}cccr@{}}
        \toprule
        Method & HE+ & MBPP+ & LCB & Code Macro & GPU$\cdot$h \\
        \midrule
        Uniform random
            & \codeselcell{88.26}{0.32}
            & \codeselcell{71.33}{0.34}
            & \codeselcell{27.50}{0.43}
            & \codeselcell{62.36}{0.22} & 0 \\
        Semantic diversity
            & \codeselcell{88.03}{}
            & \codeselcell{\textbf{71.83}}{}
            & \codeselcell{\textbf{28.14}}{}
            & \codeselcell{\textbf{62.67}}{} & 0.1 \\
        Hard selection
            & \codeselcell{88.03}{}
            & \codeselcell{71.56}{}
            & \codeselcell{\textbf{28.14}}{}
            & \codeselcell{62.58}{} & 6.9 \\
        T--S disagreement
            & \codeselcell{87.58}{}
            & \codeselcell{70.92}{}
            & \codeselcell{27.86}{}
            & \codeselcell{62.12}{} & 8.6 \\
        Cost-D-opt
            & \codeselcell{\textbf{88.42}}{}
            & \codeselcell{71.17}{}
            & \codeselcell{27.64}{}
            & \codeselcell{62.41}{} & 48.6 \\
        \bottomrule
    \end{tabular*}
    \endgroup
\end{table}

Semantic diversity leads Code Macro at low screening cost; greater
screening cost does not reliably produce a higher score in this comparison.

\subsection{Prompt selection in cross-model distillation}
\label{app:30b-prompt-selection}
\label{app:crossmodel-prompt-selection}

We compare five selection methods for
\modelname{Qwen3-30B-A3B-Instruct-2507} $\to$ \modelname{Qwen3-4B}
distillation. Each method selects $M=48$ prompts from the same DeepMath L6
candidate pool of $C=2{,}304$ prompts, followed by 25 OPD updates.
Hard selection uses the lowest initial-student pass@8, and Shortest
uses the lowest mean response length over eight initial-student samples.
Random selectors use selection seeds 42/43/44; selection costs are approximate.

\begin{table}[!htbp]
    \centering
    \caption{\textbf{Prompt selection within DeepMath L6.}
    30B Instruct $\to$ Qwen3-4B; $M=48$, $C=2{,}304$, 25 updates.
    Scores (\%) use mean@16 for Math and mean@8 otherwise;
    Avg$_4$ averages the four benchmarks.
    Random selectors report mean $\pm$ sample SD across three support draws.
    Bold marks column-best scores. Costs cover selection only;
    $0$ denotes no GPU screening.}
    \label{tab:30b-prompt-selection}
    \label{tab:crossmodel-prompt-selection}
    \begingroup
    \small
    \setlength{\tabcolsep}{2.8pt}
    \renewcommand{\arraystretch}{1.10}
    \newcommand{\crossselcell}[2]{%
        \makebox[22.5pt][r]{#1}%
        \makebox[25pt][l]{%
            \if\relax\detokenize{#2}\relax\else
                {\fontsize{8}{9}\selectfont\textcolor[HTML]{596579}{\,$\pm$\,#2}}%
            \fi}%
    }
    \begin{tabular*}{\linewidth}{@{}p{86pt}@{\hspace{5pt}}c@{\extracolsep{\fill}}ccccr@{}}
        \toprule
        Method & Math & GPQA-D & HE+ & LCB & Avg$_4$ & GPU$\cdot$h \\
        \midrule
        Uniform random
            & \crossselcell{40.56}{1.07}
            & \crossselcell{49.87}{0.55}
            & \crossselcell{86.16}{0.52}
            & \crossselcell{25.74}{0.52}
            & \crossselcell{50.58}{0.61} & 0 \\
        Stratified random
            & \crossselcell{\textbf{41.79}}{1.13}
            & \crossselcell{\textbf{50.65}}{1.03}
            & \crossselcell{86.81}{1.47}
            & \crossselcell{26.81}{0.36}
            & \crossselcell{\textbf{51.52}}{0.83} & 5.7 \\
        Semantic diversity
            & \crossselcell{40.21}{}
            & \crossselcell{50.44}{}
            & \crossselcell{86.20}{}
            & \crossselcell{\textbf{27.36}}{}
            & \crossselcell{51.05}{} & $<0.1$ \\
        Hard selection
            & \crossselcell{40.00}{}
            & \crossselcell{49.87}{}
            & \crossselcell{\textbf{87.65}}{}
            & \crossselcell{26.64}{}
            & \crossselcell{51.04}{} & 5.7 \\
        Shortest
            & \crossselcell{35.05}{}
            & \crossselcell{46.53}{}
            & \crossselcell{83.61}{}
            & \crossselcell{20.21}{}
            & \crossselcell{46.35}{} & 5.7 \\
        \bottomrule
    \end{tabular*}
    \endgroup
\end{table}

Stratified random sampling achieves the highest reported four-task
average, $51.52\pm0.83$, compared with $50.58\pm0.61$ for uniform
random sampling; selecting the shortest responses yields $46.35$
(\crossref{Table~\ref{tab:30b-prompt-selection}}).

\subsection{Initial-gradient representativeness}
\label{app:initial-gradient-representativeness}
\begingroup
\setlength{\abovedisplayskip}{5pt}
\setlength{\belowdisplayskip}{5pt}
\setlength{\abovedisplayshortskip}{3pt}
\setlength{\belowdisplayshortskip}{3pt}
\setlength{\jot}{2pt}

We diagnose the JustRL--DeepMath selection study with $C=384$ and $M=8$,
using the initial-student gradients from \crossref{Appendix~\ref{app:prompt-selection-methods}}.
We use the discovery half and concatenate three independent, 32,768-dimensional
CountSketch projections before SVD. No SVD truncation, PCA, or top-$r$
projection is applied; these are not the stable coordinates used by Cost-D-opt.

Let $\widetilde g_{ir}$ concatenate the three token-mean gradient sketches
for discovery rollout $r$ of prompt $i$, and let $T_{ir}$ count its valid
response tokens. Both within-prompt and across-prompt aggregation are
token-weighted:
\begin{align}
    L_i &= \sum_{r\in\mathcal R_i^{\mathrm{disc}}} T_{ir},
    &\widetilde g_i
    &= \frac{\sum_{r\in\mathcal R_i^{\mathrm{disc}}}T_{ir}\widetilde g_{ir}}{L_i},
    \label{eq:representativeness-prompt-gradient}\\
    \bar g_{\mathcal A}
    &= \frac{\sum_{i\in\mathcal A}L_i\widetilde g_i}
             {\sum_{i\in\mathcal A}L_i},
    &\mathcal A &\in\{\mathcal S,\mathcal C\}.
    \label{eq:representativeness-support-gradient}
\end{align}
The relative approximation error and directional alignment are
\begin{align}
    E(\mathcal S)
    &=\frac{\|\bar g_{\mathcal S}-\bar g_{\mathcal C}\|_2}
            {\|\bar g_{\mathcal C}\|_2}
      =\sqrt{\frac{\sum_{q=1}^{3}
          \|\bar g_{\mathcal S}^{(q)}-\bar g_{\mathcal C}^{(q)}\|_2^2}
          {\sum_{q=1}^{3}\|\bar g_{\mathcal C}^{(q)}\|_2^2}},
    \label{eq:initial-gradient-error}\\
    \cos(\bar g_{\mathcal S},\bar g_{\mathcal C})
    &=\frac{\langle\bar g_{\mathcal S},\bar g_{\mathcal C}\rangle}
            {\|\bar g_{\mathcal S}\|_2\|\bar g_{\mathcal C}\|_2}.
    \label{eq:initial-gradient-cosine}
\end{align}
The error is reported as $100E(\mathcal S)$ in percent. The three projection
contributions are summed before taking norms or computing cosine; their
individual error ratios and cosines are not averaged.
In \crossref{Figure~\ref{fig:selection-gradient}}, each line connects the same support's
initial-gradient approximation error and final four-task average; it does
not trace a training trajectory.

We evaluate 2,000 random $M=8$ supports in this same pre-SVD,
token-weighted geometry. Their cosine mean is $0.97053344$; the 5th,
50th, and 95th percentiles are $0.95150783$, $0.97278932$, and
$0.98164110$. These are diagnostic draws, distinct from the three
trained uniform-random supports (selection seeds 42/43/44).

\begin{table}[!htbp]
    \centering
    \caption{\textbf{Per-support gradient representativeness and transfer.}
    JustRL--DeepMath, $C=384$, $M=8$.
    Each row is one support in \crossref{Figure~\ref{fig:selection-gradient}}, not a
    mean across selection seeds. $E(\mathcal S)$ and $\mathrm{Avg}_4$
    are percentages; cosine is dimensionless.}
    \label{tab:initial-gradient-representativeness}
    \begingroup
    \small
    \setlength{\tabcolsep}{10pt}
    \renewcommand{\arraystretch}{1.10}
    \begin{tabular}{@{}lcrrr@{}}
        \toprule
        Method & Selection seed & $E(\mathcal S)$ (\%) & Cosine & $\mathrm{Avg}_4$ (\%) \\
        \midrule
        Uniform random & 42 & 29.7109 & 0.9670 & 37.5227 \\
        Uniform random & 43 & 29.7557 & 0.9565 & 38.0760 \\
        Uniform random & 44 & 23.4674 & 0.9722 & 37.3466 \\
        Stratified random & 42 & 21.7985 & 0.9833 & 37.0561 \\
        Semantic diversity & 42 & 23.5230 & 0.9721 & 37.4771 \\
        Hard selection & 42 & 39.0314 & 0.9606 & 37.2021 \\
        Cost-D-opt & 42 & 55.5506 & 0.9515 & 37.1266 \\
        Shortest & 42 & 64.9447 & 0.8412 & 36.4925 \\
        \bottomrule
    \end{tabular}
    \endgroup
\end{table}

Across the eight evaluated supports, Pearson and Spearman correlations
between $E(\mathcal S)$ and $\mathrm{Avg}_4$ are $-0.638$ and $-0.262$,
respectively, computed from the unrounded values with each support
counted once. Excluding Shortest changes them to $-0.255$ and $0.107$.
These descriptive comparisons do not provide a consistent ranking of final
accuracy: stratified seed 42 has the lowest error, whereas uniform seed 43
has the highest accuracy.
\endgroup

\section{Supplementary Training Efficiency}
\label{app:training-efficiency}

Low distinct-prompt requirements do not automatically reduce training cost.
This appendix presents trajectory-reuse results and defines cost accounting.
These address the resources consumed during training, separately from the
main-text questions of how
many distinct prompts are needed and how to select them.

\subsection{Trajectory Reuse}
\label{app:trajectory-reuse}

We compare fresh-only OPD, which updates once per fresh rollout
batch, with two reuse strategies. Full-batch reuse optimizes each
batch for four passes. Uniform token replay performs one full-batch
update followed by three updates, each using a uniformly sampled
25\% subset of response tokens.

\begin{table}[!htbp]
\centering
\small
\setlength{\tabcolsep}{3.5pt}
\renewcommand{\arraystretch}{1.20}
\caption{\textbf{Performance and training cost with trajectory reuse.}
$F/U$ denotes fresh rollout batches / optimizer updates;
generated and loss tokens are in millions (including repeated exposures
for loss), and time is in minutes. Math reports mean@16 and the other
benchmarks mean@8; Avg$_4$ is their unweighted mean.}
\label{tab:trajectory-reuse}
\begin{tabular*}{\linewidth}{@{\extracolsep{\fill}}lrrrrrrrrr@{}}
\toprule
& \multicolumn{4}{c}{Training cost}
& \multicolumn{5}{c@{}}{Performance (\%)} \\
\cmidrule(lr){2-5}\cmidrule(l){6-10}
Method & $F/U$ & Generated & Loss & Time
& Math & GPQA-D & HE+ & LCB & Avg$_4$ \\
\midrule
\multicolumn{10}{@{}l}{\textbf{DeepSeek 1.5B}} \\
\addlinespace[3pt]
Fresh-only
& 67/67 & 128.19 & 128.19 & 590.5
& 32.55 & 39.02 & 63.64 & 18.57 & 38.45 \\
Fresh-only
& 20/20 & 39.13 & 39.13 & 178.8
& 31.09 & 38.83 & 63.80 & 18.36 & 38.02 \\
\addlinespace[2pt]
Uniform token replay
& 17/68 & 32.78 & 57.64 & 205.1
& 32.92 & 39.65 & 62.73 & 19.21 & 38.63 \\
Full-batch reuse
& 17/68 & 32.60 & 130.40 & 226.0
& 32.14 & 40.15 & 64.94 & 18.50 & 38.93 \\
\addlinespace[3pt]
\midrule
\multicolumn{10}{@{}l}{\textbf{30B Instruct $\rightarrow$ Qwen3-4B}} \\
\addlinespace[3pt]
Fresh-only
& 15/15 & 24.59 & 24.59 & 148.7
& 41.41 & 50.82 & 88.34 & 26.79 & 51.84 \\
\addlinespace[2pt]
Uniform token replay
& 15/60 & 22.82 & 40.00 & 142.4
& 39.64 & 49.81 & 85.52 & 26.50 & 50.37 \\
Full-batch reuse
& 15/60 & 23.94 & 95.76 & 190.2
& 40.05 & 48.42 & 78.35 & 27.00 & 48.46 \\
\bottomrule
\end{tabular*}
\end{table}

\crossref{Table~\ref{tab:trajectory-reuse}} shows setting-dependent
quality--cost trade-offs. For DeepSeek 1.5B, reuse achieves
38.63--38.93 Avg$_4$ in 205--226 minutes, compared with 38.45
in 590.5 minutes for 67-batch fresh-only OPD. Reducing fresh-only
training to 20 batches is faster still, but yields a lower score
of 38.02. For 30B$\rightarrow$4B, increasing updates from 15 to 60
at a fixed 15 fresh batches lowers Avg$_4$ from 51.84 to 50.37
with uniform replay and 48.46 with full-batch reuse.
Thus, low distinct-prompt requirements do not by themselves
imply that repeated optimization of existing trajectories
preserves transfer quality.

\subsection{Cost--quality accounting}
\label{app:cost-quality}
\label{app:budget-accounting}

Quality is considered alongside fresh generation, cumulative training
tokens, and wall-clock time.
Costs at the same quality target and quality at the same budget
answer different questions and must be distinguished.

Generated tokens count newly sampled response tokens; training tokens
count valid response tokens processed across all optimization passes.
Rollout collection and optimizer updates are recorded separately.
Teacher-scored tokens and prompt-prefill work are tracked separately where
relevant; wall-clock measurements include teacher scoring.
Evaluation samples and generation limits are excluded from the training budget.
Prompt and response lengths are reported separately. Matched-token and
matched-time controls are identified explicitly, and historical runs retain
their run-specific configurations.

\section{Future Work}
\label{app:future-work}

Our findings motivate three directions for future work. First, prompt
selection could adapt to the teacher, evolving student, and target
capability, with gains evaluated against random sampling after accounting
for screening cost. Second, controlled interventions on source composition,
prompt exposure, and switching time could clarify the mechanisms and limits
of recovery and tolerance to individually ineffective supports. Finally,
our preliminary experiments in \crossref{Appendix~\ref{app:trajectory-reuse}}
demonstrate the potential of trajectory replay to improve training
efficiency. Future work could explore adaptive replay
strategies that balance fresh trajectory generation and reuse while
maintaining transfer performance.

\end{document}